\let\mypdfximage\pdfximage
\def\pdfximage{\immediate\mypdfximage}

\documentclass[journal,onecolumn,10pt]{IEEEtran1}

\makeatletter
\def\markboth#1#2{\def\leftmark{\@IEEEcompsoconly{\sffamily}\MakeUppercase{\protect#1}}%
\def\rightmark{\@IEEEcompsoconly{\sffamily}\MakeUppercase{\protect#2}}}
\makeatother

\let\labelindent\relax

\usepackage{times}
\usepackage{marvosym}
\usepackage{longtable}
\usepackage{graphics}
\usepackage{graphicx}
\usepackage{caption}
\usepackage[english]{babel}
\usepackage{subfig}
\usepackage{epsfig}
\usepackage{color}
\usepackage{multirow}
\usepackage{mathrsfs}
\usepackage{textcomp}
\usepackage{amsfonts}
\usepackage{amsmath}
\usepackage{amssymb}
\usepackage{nccmath}
\usepackage[numbers,square,sort&compress,comma]{natbib}
\usepackage{chapterbib}
\usepackage[ruled,vlined,linesnumbered]{algorithm2e}
\usepackage{setspace}
\usepackage{verbatim}
\usepackage{wrapfig}
\usepackage{dblfloatfix}
\usepackage{comment}
\usepackage[strict]{changepage}
\usepackage{ragged2e}
\usepackage{microtype}
\usepackage{enumitem}
\usepackage{nccmath}
\usepackage{hyperref}
\usepackage{doi}
\usepackage{afterpage}
\usepackage{multirow,bigdelim}
\usepackage{booktabs}
\usepackage{color}
\usepackage[outline]{contour}
\usepackage{xcolor}
\usepackage{colortbl}%
  
\usepackage{adjustbox}
\usepackage[nameinlink]{cleveref}
\usepackage{pdfpages}
\usepackage{bibunits}
\usepackage{multicol}
\usepackage{lscape}
\usepackage{comment}
\usepackage{array}

\newcolumntype{L}[1]{>{\raggedright\let\newline\\\arraybackslash\hspace{0pt}}m{#1}}
\newcolumntype{C}[1]{>{\centering\let\newline\\\arraybackslash\hspace{0pt}}m{#1}}

\setlist{parsep=0pt,listparindent=\parindent}

\definecolor{mygreen}{RGB}{81, 203, 0}
\definecolor{myred}{RGB}{203, 29, 0}

{\fontsize{9.75}{10}\selectfont \defaultbibliographystyle{Sledge-TNNLS-2020-1col-1} \defaultbibliographystyle{IEEEtran}}

\DeclareCaptionFont{singlespacing}{\setstretch{1}}
\renewcommand \thesection{\arabic{section}}
\numberwithin{equation}{section}

\hypersetup{bookmarks=false,bookmarksopen=false,pdfpagemode=empty,pdfstartview=}

\let\oldnl\nl
\newcommand{\nonl}{\renewcommand{\nl}{\let\nl\oldnl}}

\title{\singlespacing\sf\huge Faster Convergence in Deep-Predictive-Coding Networks to Learn Deeper Representations} 
\author{Isaac J. Sledge, \emph{Member, IEEE}, and Jos\'{e} C. Pr\'{i}ncipe, \emph{Life Fellow, IEEE}%
\thanks{\fontdimen2\font=1.55pt Isaac J. Sledge is the Senior Machine Learning Research Scientist and Dr. Delores M. Etter Assistant Secretary of the Navy Emergent Engineer with the Advanced Signal Processing and Automated Target Recognition Branch, Naval Surface Warfare Center, Panama City, FL, USA (email: isaac.j.sledge@navy.mil).  He is the director of the Machine Intelligence Defense (MIND) lab at the US Naval Sea Systems Command.}  
\thanks{\fontdimen2\font=1.55pt Jos\'{e} C. Pr\'{i}ncipe is the Don D. and Ruth S. Eckis Chair and the Distinguished Professor with both the Department of Electrical and Computer Engineering and the Department of Biomedical Engineering, University of Florida, Gainesville, FL, USA (email: principe@ufl.edu).  He is the director of the Computational NeuroEngineering Laboratory (CNEL) at the University of Florida.\vspace{0.1cm}}
\thanks{The work of the authors was funded by grants N00014-14-1-0542 (Marc Steinberg, ONR 35), N00014-19-WX-00636 (Marc Steinberg, ONR 35), N00014-21-WX-01657 (Thomas McKenna, ONR 34), and N00014-21-WX-00476 (J. Tory Cobb, ONR 32) from the US Office of Naval Research.  The first author was also supported by in-house laboratory independent research (ILIR) grant N00014-19-WX-00687 (Frank Crosby) from the US Office of Naval Research and a Naval Innovation in Science and Engineering (NISE) grant from NAVSEA.}%
}
\begin{document}
\begin{bibunit}
\bstctlcite{IEEEexample:BSTcontrol}






\maketitle
\RaggedRight\parindent=1.5em
\fontdimen2\font=2.1pt
\vspace{-1.55cm}\begin{abstract}\normalsize\singlespacing
\vspace{-0.25cm}{\small{\sf{\textbf{Abstract}}}}---Deep-predictive-coding networks (DPCNs) are hierarchical, generative models.  They rely on feed-forward and feed-back connections to modulate latent feature representations of stimuli in a dynamic and context-sensitive manner.  A crucial element of DPCNs is a forward-backward inference procedure to uncover sparse, invariant features.  However, this inference is a major computational bottleneck.  It severely limits the network depth due to learning stagnation.  Here, we prove why this bottleneck occurs.  We then propose a new forward-inference strategy based on accelerated proximal gradients.  This strategy has faster theoretical convergence guarantees than the one used for DPCNs.  It overcomes learning stagnation.  We also demonstrate that it permits constructing deep and wide predictive-coding networks.  Such convolutional networks implement receptive fields that capture well the entire classes of objects on which the networks are trained.  This improves the feature representations compared with our lab's previous non-convolutional and convolutional DPCNs.  It yields unsupervised object recognition that surpass convolutional autoencoders and are on par with convolutional networks trained in a supervised manner.



\end{abstract}%
\begin{IEEEkeywords}\normalsize\singlespacing
\vspace{-1.35cm}{{\small{\sf{\textbf{Index Terms}}}}---Bio-inspired vision, predictive coding, unsupervised learning}
\end{IEEEkeywords}
\IEEEpeerreviewmaketitle
\allowdisplaybreaks
\singlespacing

\vspace{-0.4cm}\subsection*{\small{\sf{\textbf{1.$\;\;\;$Introduction}}}}\addtocounter{section}{1}

Predictive coding is a promising theory for unsupervised sensory information processing.  Under this theory, a hierarchical, generative model \cite{RaoRPN-jour1999a,FristonK-jour2009a} of a dynamic environment is formed.  This model is consistently updated to infer possible states of the environment and physical causes of the environmental stimuli.  These causes, in turn, permit reproducing the stimuli \cite{SpratlingMW-jour2011a} (see Section 2).

Several predictive coders have been created and their biological plausibility investigated \cite{SrinivasanMV-jour1982a,RaoRPN-jour1999a,JeheeJFM-jour2006a,FristonK-jour2008a}.  None of these early contributions has been known to form causes that are highly discriminative for complex stimuli, however.  Our lab thus developed multi-stage, deep-predictive-coding networks (DPCNs) \cite{ChalasaniR-conf2013a,PrincipeJC-jour2014a,ChalasaniR-jour2015a}.  We showed that DPCNs could form discriminative causes of temporal, spatial, and spatio-temporal stimuli in certain cases.  Alternate networks later followed \cite{LotterW-conf2017a,HanK-coll2018a}.

DPCNs learn about an environment in an unsupervised manner.  They can thus be thought of as parameter-light, non-traditional autoencoders.  DPCNs differ substantially from autoencoders, though.  The former contain feed-forward and recurrent, feed-back connections, allowing information to be propagated between stages to stabilize the internal representation.  Another distinction is that DPCNs do not have a corresponding decoder.  They hence require a self-organizing principle to be effective.  That is, they must learn to extract meaningful features, in the form of causes, that cluster the stimuli effectively.  This clustering objective is aided by imposing cause sparsity and transformation invariance.  Including these cause constraints make DPCNs generalize well to novel stimuli.  Lastly, DPCNs are composed of stages, not layers.  Each stage implements a recurrent state model.  This permits DPCNs to characterize the dynamics of temporal and spatio-temporal stimuli.  




DPCNs exhibit promise for unsupervised object recognition in images and video.  Often, convolutional DPCNs like \cite{ChalasaniR-jour2015a} learn sparse, invariant features that are better for classification than those from convolutional autoencoders.  Such behavior arises from the interaction of feed-forward and feed-back connections in the DPCNs.  It also arises due to the implicit supervision imposed from leveraging temporal information \cite{BakerR-jour2014a}.  Multiple presentations of the stimuli additionally facilitate the extraction of spatial and temporal regularities \cite{PerruchetP-jour2006a,AslinRN-jour2012a}, which we hypothesize permits high-level object analysis \cite{BradyTF-jour2007a}, like object recognition.  Sparsity contributes too \cite{ZylberbergJ-jour2011a,CarlsonNL-jour2012a}, as it aids in generalization and can preempt overfitting to specific stimuli.  


Learning sufficiently robust features in a DPCN is quite computationally intensive.  A multi-stage optimization strategy, based on proximal gradients \cite{GulerO-jour1992b,BeckA-jour2009a}, is typically used \cite{ChalasaniR-conf2013a,PrincipeJC-jour2014a,ChalasaniR-jour2015a} to conduct feed-forward and feed-back inference of the causes.  Both forms of inference are needed for cause self-organization.  Sub-quadratic function-value convergence rates are theoretically guaranteed for this strategy \cite{ChambolleA-jour2015a,AttouchH-jour2016a}.  Only sub-linear rates are often obtainable, however, due to severe oscillations in the cost.  That is, the search is not a pure descent strategy in some cases.  This has the dual effect of returning poor causes and doing so slowly.  

Due to these optimization difficulties, DPCNs are practically limited to two stages.  Two stages may be sufficient for characterizing certain stimuli.  However, it typically will not yield representations that handle objects in complex environments.  The networks exhibit poor stimuli-reconstruction performance when extended beyond two stages due to being stymied by poor convergence \cite{SantanaE-jour2018a}.  The deeper stages do not reach a stable cause representation, which impacts the causes in preceding stages.  Learning essentially stagnates regardless of how many stimuli are presented.  This, in turn, prevents object recognition for many challenging environments.  The causes simply do not organize the stimuli well.  Moreover, the causes are not semantically rich enough to handle distractors.  Even simple textural backgrounds in visual stimuli can be sufficiently distracting enough to confound the DPCNs.  The causes are also sometimes unable to handle object variability effectively.  They cannot always resolve that objects from distinct viewpoints are same, for instance.  

Here, we develop a novel inference process that enables investigators to go beyond the two-stage network limitation that is observed in practice for DPCNs.  This leads to what we refer to as accelerated DPCNs (ADPCNs).  Both the DPCNs and ADPCNs are unsupervised networks.  They share the same underlying architecture, with one exception: the ADPCNs extract convolutional features (see Section 3).  Both properties mean that ADPCNs are better suited for discrimination than our lab's original, non-convolutional DPCNs \cite{ChalasaniR-conf2013a,PrincipeJC-jour2014a}.  The improved inference also makes ADPCNs better than our lab's convolutional-recurrent-predictive networks (CRPNs) \cite{ChalasaniR-jour2015a}.  CRPNs similarly rely on slow proximal-gradient inference.  

More specifically, we replace the proximal gradient search in the original DPCN with an accelerated version (see Section 3).  The accelerated approach in the ADPCN relies on a polynomial inertial sequence for updating the internal feature representations.  The inertial sequence has the effect of sufficiently delaying the occurrence of cost oscillation.  We prove this (see Appendix A).  Monotonically decreasing costs occur throughout almost all of the learning process.  We perform in-place restarts of the inference when it does not.  The ADPCNs therefore extract meaningful error signals that stabilize the sparse causes early during training.  ADPCNs also possesses a faster, sub-polynomial rate of function-value convergence compared to the inference scheme used by DPCNs (See Appendix A).  ADPCNs hence exhibit improved empirical convergence too over DPCNs.  We are thus able to efficiently train both deep and wide predictive-coding networks whose learning does not stall.  The deep convolutional layers of the ADPCN extract progressively richer, transformation-invariant features for sparsely describing complex stimuli.  They aid in stimuli generalization.  Wide convolutional layers better approximate interactions between stimuli and causes than narrower ones.  They promote some measure of stimuli memorization without overfitting, which permits recognizing perceptually similar objects from the same class.  While such behaviors can manifest in convolutional DPCNs, they do not due to the aforementioned slow inference.


We show that the deeper and wider feature representations from the ADPCN are far more robust than those obtainable for DPCNs. Even a single stage difference between convolutional DPCNs and ADPCNs leads to a huge improvements in discriminability.  In particular, the later-stage causes in the ADPCNs have receptive fields that embody the entirety of the objects being presented, despite the lack of training labels (see Section 3).  We observe this phenomenon across a variety of benchmark datasets.  It enables multi-class object recognition under different scene conditions.  The ADPCNs can handle perspective changes, shape changes, illumination changes, and more.  ADPCNs thus form an approximate identity mapping that preserves perceptual difference.  Whole-object sensitivity also yields unsupervised classifiers that are on par with supervised-trained convolutional and convolutional-recurrent networks (see Appendix B).  The ADPCNs have orders of magnitude fewer parameters, though, than these other deep networks.  ADPCNs, just like DPCNs, are parameter-light, non-traditional autoencoders.  

\subsection*{\small{\sf{\textbf{2.$\;\;\;$Predictive Coding}}}}\addtocounter{section}{1}

The objective of predictive coding is to approximate external sensory stimuli using generative, latent-variable models.  Such models hierarchically encode residual prediction errors.  The prediction errors are the differences between either the actual stimuli or a transformed version of it and the predicted stimuli produced from the underlying latent variables.  We refer to these latent variables as causes.  We also interchangeably refer to causes as features.

By learning in this way, the internal representations of a predictive-coding model are modified only for unexpected changes in the stimuli.  This enables a model to recall stimuli that it has encountered before.  It also enables the model to adapt to new stimuli without disrupting its internal representation of the environment for previously observed stimuli.

We can characterize predictive coding in the following way for temporal and spatio-temporal stimuli.  Examples include audio and video, respectively.  Both DPCNs and ADPCNs are instances of this general framework.

\begin{itemize}
\item[] \-\hspace{0.5cm}{\small{\sf{\textbf{Definition 1: Predictive Coding Model.}}}} Let $y_t \!\in\! \mathbb{R}^p$ represent a time-varying sensory stimulus at time $t$.  The stimuli can be described by an underlying cause, $\kappa_{1,t} \!\in\! \mathbb{R}^{d_1}$, and a time-varying intermediate state,\\ \noindent $\gamma_{1,t} \!\in\! \mathbb{R}^{k_1}$, through a pair of $\theta$-parameterized mapping functions, $f_1 : \mathbb{R}^{k_1} \!\to\! \mathbb{R}^p$, the cause-update function, and\\ \noindent $g_1 : \mathbb{R}^{k_1} \!\times\! \mathbb{R}^{d_1} \!\to\! \mathbb{R}^{k_1}$, the state-transition function.  These functions define a latent-variable model,
\begin{equation*}
y_t \!=\! f_1(\gamma_{1,t};\theta) \!+\! \epsilon_{1,t},\; \gamma_{1,t} \!=\! g_1(\gamma_{1,t-1},\kappa_{1,t};\theta) \!+\! \epsilon_{0,t}'.
\end{equation*}
Here, $\epsilon_{1,t} \!\in\! \mathbb{R}^p$ and $\epsilon_{1,t}' \!\in\! \mathbb{R}^{k_1}$ are noise terms that represent the stochastic and model uncertainty, respectively, in\\ \noindent the predictions.  This model can be extended to a multi-stage hierarchy by cascading additional $\theta$-parameterized mapping functions, $f_i : \mathbb{R}^{k_{i}} \!\to\! \mathbb{R}^{d_{i-1}}$ and $g_i : \mathbb{R}^{k_{i}} \!\times\! \mathbb{R}^{d_i} \!\to\! \mathbb{R}^{k_i}$, at each stage $i$ beyond the first,
\begin{equation*}
\kappa_{i-1,t} \!=\! f_{i}(\gamma_{i,t};\theta) \!+\! \epsilon_{i,t},\; \gamma_{i,t} \!=\! g_i(\gamma_{i,t-1},\kappa_{i,t};\theta) \!+\! \epsilon_{i,t}'.
\end{equation*}
Here, $\kappa_{i,t} \!\in\! \mathbb{R}^{d_i}$, $\gamma_{i,t} \!\in\! \mathbb{R}^{k_i}$, $\epsilon_{i,t} \!\in\! \mathbb{R}^{d_{i-1}}$ and $\epsilon_{i,t}' \!\in\! \mathbb{R}^{k_i}$.
\end{itemize}
Spatial stimuli, which include images, are a degenerate case of this framework.  They have no temporal component, so there is no modification of the states as they are fed back.  States are still extracted, though.  Predictive coding thus behaves similarly to sparse coding \cite{HyvarinenA-jour2001a} in this case.  The main difference is that, for predictive coding, the states are pooled and transformed to form causes.  Sparse coding only has notions of states.  For DPCNs and ADPCNs, the causes are made invariant, which aids in discrimination.  Feature invariance does not naturally occur in many sparse coding and hierarchical sparse coding models \cite{BoutinV-jour2020a,BoutinV-jour2021a}.  


For the above framework, both feed-forward, bottom-up and feed-back, top-down processes are used to characterize observed stimuli.  For the feed-forward case, the observed stimuli are propagated through the model to extract progressively abstract details.  The stimuli are first converted to a series of states that encode either spatial, temporal, or spatio-temporal relationships.  The type of relationship depends on the stimuli being considered.  These states are then made invariant to various transformations, thereby forming the hidden causes.  The causes are latent variables that describe the environment, as we noted above.

The causes at lower stages of the model form the observations to the stages above.  Hidden causes therefore provide a link between the stages.  The states, in contrast, both connect the dynamics over time, to ignore temporal discontinuities \cite{StoneJV-jour2008a}, and mediate the effects of the causes on the stimuli \cite{FristonK-jour2008a}.  In the feed-back case, the model generates top-down predictions such that the neural activity at one stage predicts the activity at a lower stage.  The predictions from a higher level are sent through feed-back connections to be compared to the actual activity.  This yields a model uncertainty error that is forwarded to subsequent stages to update the population activity and improve prediction.  Such a top-down process repeats until the bottom-up stimuli transformation process no longer imparts any new information.  That is, there are no unexpected changes in the stimuli that the model cannot predict.  Once this occurs, if the model is able to synthesize the input stimuli accurately using the uncovered features, then it means that it has previously seen a similar observation \cite{SchendanHE-jour2008a}.

In short, a predictive-coding model has two processing pathways.  Recurrent, top-down connections carry predictions about activity to the lower model levels.  These predictions reflect past experience \cite{FristonK-jour2008a}.  They form priors to disambiguate the incoming sensory inputs.  Bottom-up connections relay prediction errors to higher levels to update the physical causes.  The interaction of the feed-forward and feed-back connections \cite{HosoyaT-jour2005a} on the causes enables robust object analysis \cite{AuksztulewiczR-jour2016a} from the observed stimuli.  A well-trained predictive-coding model should thus distinguish between objects, despite learning about them in an unsupervised way.  This only occurs if the models preserve some notion of perceptual difference.

\begin{figure*}
   \hspace{1.25cm}\includegraphics[height=1.75in]{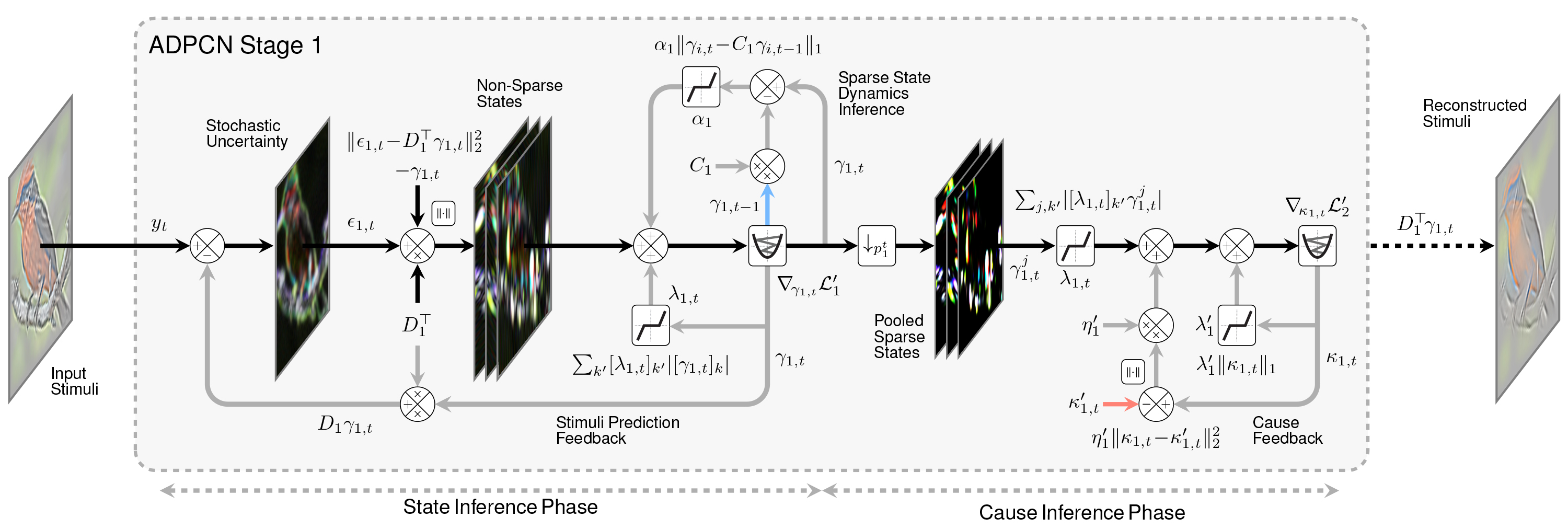}
   \caption[]{\fontdimen2\font=1.55pt\selectfont An overview of the ADPCN architecture.  For simplicity, the ADPCN is presented for only one stage, but more stages are implicitly assumed for top-down feed-back purposes.  Note that the final stage of the DPCN has no output, unlike in a standard autoencoder network.  The goal of the ADPCN is to learn a series of causes that explain the input stimuli, in this case, frames from a video of a bird, and hence recreate them.  Each stage in an ADPCN can be roughly decomposed into two inference phases, one for updating the states and the other for updating the causes.  State and causal inference relies on intra-stage feed-forward (black lines) and feed-back processes (gray lines), along with intra-stage recurrent feed-back (blue lines).  Much of the network is devoted to feed-back processes to provide self supervision.  Inter-stage feed-back (red line) of higher-level states is used to update the lower-level causes.  For the inference process, we denote multiplication of a quantity on a given path with a rounded square.  Addition, subtraction, and multiplication of quantities along multiple paths are denoted using circular gated symbols.  Pooling is denoted using a down arrow inside a rounded square.  The function blocks apply a sparsity operation to the quantities on the given path; corresponding parameter values for these operations are offset from these blocks.  Function blocks offset from circular gates denote rectified, linear-unit operations.  Norm blocks offset from the circular gates denote squared-norm operations.  We omit showing the actual optimization process but note that the feed-forward connections are largely devoted to computing gradients of the two cost functions, $\nabla_{\gamma_{i,t}}\mathcal{L}_1$ and $\nabla_{\kappa_{i,t}}\mathcal{L}_2$, that are used to update the states $\gamma_{i,t}$ and causes $\kappa_{i,t}$.  We also omit showing the update of $\lambda_{i,t}$ along with the updates for the matrices $C_i$, $D_{i}$, and $G_{i}$.\vspace{-0.4cm}} 
   \label{fig:dpcn-network}
\end{figure*}

\subsection*{\small{\sf{\textbf{3.$\;\;\;$Deep Predictive Coding}}}}\addtocounter{section}{1}

We propose an efficient architecture for the above hierarchical, latent-variable model.  This model is suitable for uncovering discriminative details from the stimuli.  

More specifically, we consider a faster, convolutional DPCN, the ADPCN.  ADPCNs can extract highly sparse, invariant features for either dynamic or static stimuli (see Section 3.1).  We then show how to effectively infer the ADPCN's latent variables using a fast proximal gradient scheme (see Section 3.2).  This optimization process permits effectively forming deep feature hierarchies that preserve perceptual differences.  It typically overcomes the learning stagnation that we observed in DPCNs.  Convergence properties are presented in the online appendix (see Appendix A).  In this appendix, we prove why stagnation occurs in the DPCNs.  We also quantify the convergence rate of the DPCNs and ADPCNs to show that the latter, theoretically, converge more quickly than the former.  This justifies our new inference process, as do our empirical results.  

\subsection*{\small{\sf{\textbf{3.1$\;\;\;$ADPCN Cost Functions}}}}


ADPCNs consist of two phases at each stage, which are outlined in \cref{fig:dpcn-network}.  The first phase entails inferring the hidden states, which are a feature-based representation used to describe the stimuli.  States are formed at the first phase via sparse coding in conjunction with a temporal-state-space model.  Stimuli are mapped to an over-complete dictionary of convolutional filters.  Subsequent ADPCN stages follow the same process, with the only change being that the hidden causes assume the role of the observed stimuli.

We define state inference via a least-absolute-shrinkage-and-selection-operator (LASSO) cost.  We present this for the case of single-channel stimuli, for the ease of readability.  The extension to multiple channels is straightforward.

 
\begin{itemize}
\item[] \-\hspace{0.5cm}{\small{\sf{\textbf{Definition 2: State LASSO Cost.}}}} Let $\gamma_{i,t} \!\in\! \mathbb{R}^{k_i}$ be the hidden states at time $t$ and at model stage $i$.  Let $C_i \!\in\! \mathbb{R}^{k_i \times k_i}$ be a hidden-state-transition matrix.  Let $D_{i}^\top \!\in\! \mathbb{R}^{k_i \times k_{i-1}}$ be a Toeplitz-form matrix with $q_i \!\in\! \mathbb{N}_+$\\ \noindent filters structured as in \cite{SulamJ-jour2018a}. The state-inference cost function to be minimized, with respect to $\gamma_{i,t}$, $C_i$, and $D_{i,q}^\top$, is given by
\begin{equation*}
\mathcal{L}_1(\gamma_{i,t},\kappa_{i,t},C_i,D_{i}^\top;\alpha_i,\lambda_{i,t}) = \frac{1}{2}\Bigg(\|\kappa_{i-1,t} \!-\! D_{i}^\top\gamma_{i,t}\|_2^2 + \alpha_i\|\gamma_{i,t} \!-\! C_i\gamma_{i,t-1}\|_1 + \sum_{k'=1}^{k_i}[\lambda_{i,t}]_{k'} |[\gamma_{i,t}]_{k'}|\Bigg),
\end{equation*}
\noindent where $\kappa_{0,t} \!=\! y_t$.  The first term in this cost quantifies the $L_2$ prediction error, $\epsilon'_{i,t} \!=\! \kappa_{i-1,t} \!-\! D_i^\top \gamma_{i,t}$, at stage $i$.  The aim is to ensure that the local reconstruction error between stages is minimized.  The second term constrains the next-state dynamics to be described by the state-transition matrix.  For static stimuli, indexed by $t$, the state feed-back is replaced by $\kappa_{i,t} \!-\! D_{i+1}^\top \gamma_{i+1,t}$.  The strength of the recurrent feed-back connection is driven by $\alpha_i \!\in\! \mathbb{R}_+$.  The transitions are $L_1$-sparse to make the state-space representation consistent.  Without such a norm penalty, the innovations would not be sparse due to the feed-back.  The final term enforces $L_1$-sparsity of the states, with the amount controlled by $\lambda_{i,t} \!\in\! \mathbb{R}_+^{k_i}$.
\end{itemize}
\begin{itemize}
\item[] \-\hspace{0.5cm}{\small{\sf{\textbf{Proposition 1.}}}} Let $\gamma_{i,t} \!\in\! \mathbb{R}^{k_i}$ be the hidden states at time $t$ and at model stage $i$.  Let $D_{i}^\top \!\in\! \mathbb{R}^{k_i \times k_{i-1}}$ be\\ \noindent a Toeplitz-form matrix of $q_i \!\in\! \mathbb{N}_+$ filters.  The matrix-vector multiplication $D_{i}^\top\gamma_{i,t}$ is functionally equivalent to\\ \noindent convolution for all stages $i$.
\end{itemize}
When projected back into the original visual space of the input, the dictionaries define a series of receptive fields.  The hidden states, at least for the initial stages of the hierarchy thus act as basic feature detectors.  They often resemble the simple cells in the visual cortex \cite{OlshausenBA-jour1996a}.  




Ideally, we would prefer $L_0$-sparsity to the $L_1$ variant used in the state-inference cost, as it does not impose shrinkage on the hidden-state values.  However, by sacrificing this property, we gain cost convexity, which aids in efficient numerical optimization with provable convergence.
\begin{itemize}
\item[] \-\hspace{0.5cm}{\small{\sf{\textbf{Proposition 2.}}}} Let $\gamma_{i,t} \!\in\! \mathbb{R}^{k_i}$ be the hidden states at time $t$ and at model stage $i$.  Let $C_i \!\in\! \mathbb{R}^{k_i \times k_i}$ be a hidden-state-transition matrix.  Let $D_i^\top \!\in\! \mathbb{R}^{k_i \times k_{i-1}}$ be a Toeplitz-form matrix with $q_i \!\in\! \mathbb{N}_+$ filters.  Let $\alpha_i,\lambda_{i,t} \!\in\! \mathbb{R}_+$.\\ \noindent The hidden-state cost function $\mathcal{L}_1(\gamma_{i,t},\kappa_{i,t},C_i,{D_i^\top};\alpha_i,\lambda_{i,t})$ is convex for appropriate parameter values.  
\end{itemize}
Note that the state LASSO may switch from convex to non-convex as the variables are updated during inference.  We may thus only recover local minimizers, not global ones.


The state-based feature representations constructed by the first phase are not guaranteed to be invariant to various transforms.  Discrimination can be impeded, as a result.  The second ADPCN processing phase thus entails explicitly imposing this behavior.  Local translation invariance is attained by leveraging the spatial relationships of the states in neighborhoods via the max-pooling of states.  Invariance to more complex transformations, like rotation and spatial frequency, is made possible through the inference of subsequent hidden causes.  

Sparse cause inference is driven by a LASSO-based cost that captures non-linear dependencies between components in the pooled states.  We present this cost for the case of a single convolutional layer, for the ease of readability.  We utilize multiple layers in our simulations.

\begin{itemize}
\item[] \-\hspace{0.5cm}{\small{\sf{\textbf{Definition 3: Cause LASSO Cost.}}}} Let $\gamma_{i,t} \!\in\! \mathbb{R}^{k_i}$ be the hidden states and $\kappa_{i,t} \!\in\! \mathbb{R}^{d_i}$ be the bottom-up hidden causes at time $t$ and model stage $i$.  Let $\kappa'_{i,t} \!\in\! \mathbb{R}^{d_i}$ be the top-down-inferred causes.  Let $C_i \!\in\! \mathbb{R}^{k \times k}$ be a hidden-\\ \noindent state-transition matrix.  Let $D_i^\top \!\in\! \mathbb{R}^{k_i \times k_{i-1}}$ be a Toeplitz-form matrix of $q_i \!\in\! \mathbb{N}_+$ filters.  Let $G_i \!\in\! \mathbb{R}^{d_i \times k_i}$ be an\\ \noindent invariant Toeplitz matrix.  The hidden-cause cost to be minimized, with respect to $\kappa_{i,t}$ and $G_i$, is given by
\begin{equation*}
\mathcal{L}_2(\gamma_{i,t},\kappa_{i,t},G_i;\alpha_i',\lambda_i',\eta_i',\lambda_{i,t}) = \frac{1}{2}\Bigg(\!\Bigg(\sum_{j=1}^n\sum_{k'=1}^{k_i} |[\lambda_{i,t}]_{k'}\gamma_{i,t}^{j}| \Bigg) + \eta_i'\|\kappa_{i,t} \!-\! \kappa'_{i,t}\|_2^2 + \lambda_i'\|\kappa_{i,t}\|_1\!\Bigg), 
\end{equation*}
where $\lambda_{i,t,k'} \!=\! \alpha_i'(1 \!+\! \textnormal{exp}(-[G_i^\top \kappa_{i,t}]_k))$, $\alpha_i' \!\in\! \mathbb{R}$.  The first term in this cost models the multiplicative inter-\\ \noindent action of the causes $\kappa_{i,t}$ with the max-pooled states $\gamma_{i,t}^{j}$ through an invariant Toeplitz matrix $G_i$.  This characterizes the shape of the sparse prior on the states.  That is, the invariant matrix is adapted such that each component of the causes are connected to element groups in the accumulated states that co-occur frequently.  Co-occurring components typically share common statistical regularities, thereby yielding locally invariant representations \cite{KarklinY-jour2006a}.  The second term specifies that the difference between the bottom-up $\kappa_{i,t}$ and top-down inferred causes $\kappa'_{i+1,t}$ should be small, with the term weight specified by $\eta_i' \!\in\! \mathbb{R}_+$.  The final term imposes $L_1$ sparsity, with\\ \noindent the amount controlled by $\lambda_i' \!\in\! \mathbb{R}_+$, to prevent the intermediate representations from being dense.
\end{itemize}
The causes obtained by solving the above LASSO cost will behave somewhat like complex cells in the visual cortex \cite{ItoM-jour2004a}.  Similar results are found in temporally coherent networks \cite{HurriJ-coll2003a}, albeit without guaranteed feature invariance.

As with the state-inference cost, we employ $L_1$ sparsity in the hidden-cause cost for practical reasons, even though we would prefer $L_0$ sparsity for its theoretical appeal.  
\begin{itemize}
\item[] \-\hspace{0.5cm}{\small{\sf{\textbf{Proposition 3.}}}} Let $\gamma_{i,t} \!\in\! \mathbb{R}^{k_i}$ be the hidden states and $\kappa_{i,t} \!\in\! \mathbb{R}^{d_i}$ be the hidden causes at time $t$ and model stage $i$.  Let $C_i \!\in\! \mathbb{R}^{k_i \times k_i}$ be a hidden-state transition-matrix, $D_i^\top \!\in\! \mathbb{R}^{k_i \times k_{i-1}}_+$ be a Toeplitz-form matrix of filters,\vspace{-0.025cm}\\ \noindent and $G_i \!\in\! \mathbb{R}^{d_i \times k_i}$ be an invariant Toeplitz matrix.  Let $\alpha_i',\lambda_i',\eta_i' \!\in\! \mathbb{R}_+$ and $\lambda_{i,t} \!\in\! \mathbb{R}_+^{k_i}$. The hidden-cause cost-\\ \noindent function $\mathcal{L}_2(\gamma_{i,t},\kappa_{i,t},G_i;\alpha_i',\lambda_i',\eta_i',\lambda_{i,t})$ is convex for appropriate parameter values.  
\end{itemize}
Practically, we have found that traditional $L_0$ sparsity in the ADPCN preempts learning.  The causes are often too sparse to act as priors.  Large errors continuously accumulate, so the APDCNs are unable to reduce the residual prediction errors.  Approximating the $L_0$ term is often a better option.  We will explore it more in our future endeavors.


ADPCNs and the DPCNs our lab proposed \cite{ChalasaniR-conf2013a,PrincipeJC-jour2014a} have almost the same cost functions that we outline above.  They both build unsupervised representations of input stimuli \cite{FoldiakP-jour1991a} via a free-energy principle \cite{FristonK-jour2006a}.  The difference is that ADPCNs implement convolution by way of the Toeplitz-form matrix of filters.  The original DPCNs relies on non-convolutional filters.  The expressive power of these DPCNs is thus quite poor for complex stimuli compared to the ADPCNs.  ADPCNs can take advantage of local spatial coherence in addition to temporal coherence.  They hence require fewer filters than DPCNs to extract meaningful stimuli representations.  

Defining states and causes as we have specified above has significant advantages.  ADPCNs are, for instance, incredibly parameter efficient compared to standard recurrent-convolutional autoencoders.  Few filters often are needed to adequately synthesize an observed stimuli under varying conditions.  This is a byproduct of the explicit feature invariance imposed by the non-linear, sparse cause inference.  ADPCNs are also far more efficient and effective than alternate DPCNs that have been proposed.  For instance, the DPCN proposed by Lotter et al. \cite{LotterW-conf2017a} relies on convolutional, long shot-term memories (LSTMs) to model spatio-temporal stimuli.  LSTM cells can only store and recall a single event through time, limiting their characterization of complex temporal patterns.  Many LSTM cells are needed for associative event recall.  They also require lengthy inference times.  Behaviors like memorization, associative recall, and others are possible with ADPCNs.  ADPCNs can be kernelized to realize non-linear, non-parametric state-space updates, further increasing the ADPCNs' expressive power without impacting inference times.  The DPCN proposed by Han et al. \cite{HanK-coll2018a} relies on convolutional layers with linear recurrent feedback, similar to our lab's recurrent-convolutional, winner-take-all networks \cite{SantanaE-jour2018a}.  However, the network in \cite{HanK-coll2018a} lacks an unsupervised organization mechanism.

\subsection*{\small{\sf{\textbf{3.2$\;\;\;$ADPCN Inference}}}}

The propagation and transformation of observed stimuli in a ADPCN is more involved than for standard network architectures.  At any stage in the ADPCN, the hidden, sparse states and unknown, sparse causes that minimize the two-part LASSO cost must be inferred to create the feed-forward observations for the next ADPCN stage.  

Joint inference of the states and causes can be done in a manner similar to block coordinate descent.  That is, for a given mini-batch of stimuli, the states can be updated by solving the corresponding LASSO cost while holding the causes fixed.  The causes can then be updated while holding the states fixed.  Altering either of these representations amounts to solving a convolutional, $L_1$-sparse-coding problem.  The presence of discontinuous, $L_1$-based terms in the LASSO costs complicates the application of standard optimization techniques, though.  




Here, we consider a fast proximal-gradient-based approach for separating and accounting for the smooth and non-smooth components of the LASSO costs.  This approach is motivated and analyzed in the appendix.
\begin{itemize}
\item[] \-\hspace{0.5cm}{\small{\sf{\textbf{Definition 4: Feed-Forward, Bottom-Up State Inference.}}}} Let $\gamma_{i,t} \!\in\! \mathbb{R}^{k_i}$ be the hidden states and $\pi_{i,t} \!\in\! \mathbb{R}^{k_i}$ be the auxiliary hidden states at time $t$ and model stage $i$.  These auxiliary hidden states will be combinations of\\ \noindent hidden states across different times.  Let $C_i \!\in\! \mathbb{R}^{k_i \times k_i}$ be a hidden-state-transition matrix.  As well, let\\ \noindent $D_i^\top \!\in\! \mathbb{R}^{k_i \times k_{i-1}}$ be a Toeplitz-form matrix of $q_i \!\in\! \mathbb{N}_+$ filters.  For an inertial sequence $\beta_{m} \!\in\! \mathbb{R}_+$ and an adjust-\vspace{-0.025cm}\\ \noindent able step size $\tau_{i,t}^m \!\in\! \mathbb{R}_+$, the hidden-state inference process, indexed by iteration $m$, is given by the following expressions
\begin{equation*}
\gamma_{i,t+1}^{m} \!=\! \textnormal{\sc prox}_{\!\lambda_{i,t}}\!\Bigg(\!\pi_{i,t}^{m} \!-\! \lambda_{i,t}\tau^m_{i,t}\nabla_{\pi_{i,t}^m}\mathcal{L}_1(\pi_{i,t}^m,\kappa_{i,t},C_i,D_i^\top;\alpha_i,\lambda_{i,t})\!\Bigg),\;\; \pi_{i,t}^{m+1} \!=\! \gamma_{i,t+1}^{m} \!+\! \beta_m (\gamma_{i,t+1}^{m} \!-\! \gamma_{i,t+1}^{m-1})
\end{equation*}
with $\mathcal{L}_1(\pi_{i,t}^m,\kappa_{i,t},C_i,D_i^\top;\alpha_i,\lambda_{i,t}) \!=\! D_i^\top(\kappa_{i-1,t} \!-\! D_i \pi_{i,t}^m) \!+\! \alpha_i\Omega_i(\pi_{i,t}^m)$.  Here, use a Nesterov smoothing,\\ \noindent $\Omega_i(\pi_{i,t}^m) \!=\! \textnormal{arg max}_{\|\Omega_{i,t}\|_\infty \leq 1} \Omega_{i,t}^\top (\pi_{i,t}^m \!-\! C_i \pi_{i,t}^{m-1})$, $\Omega_{i,t} \!\in\! \mathbb{R}^{k_i}$, to approximate the non-smooth state transition.  Small values for the hidden states are clamped via a soft thresholding function implicit to the proximal operator, which leads to a sparse solution.  The states are then spatially max pooled over local neighborhoods, using non-overlapping windows, to reduce their resolution, $\gamma_{i,t+1} \!=\! \textnormal{\sc pool}(\gamma_{i,t+1})$. 
\end{itemize}

\begin{itemize}
\item[] \-\hspace{0.5cm}{\small{\sf{\textbf{Definition 5: Feed-Forward, Bottom-Up Cause Inference.}}}} Let $\gamma_{i,t} \!\in\! \mathbb{R}^{k_i}$ be the hidden states and\\ \noindent $\kappa_{i,t} \!\in\! \mathbb{R}^{d_i}$ be the hidden causes at time $t$ and model stage $i$.  Let $C_i \!\in\! \mathbb{R}^{k_i \times k_i}$ be a hidden-state-transition matrix,\\ \noindent $D_i^\top \!\!\in\! \mathbb{R}^{k_i \times k_{i-1}}$ be a Toeplitz-form matrix of $q_i \!\in\! \mathbb{N}_+$ filters, and $G_i \!\in\! \mathbb{R}^{d_i \times k_i}$ be an invariant Toeplitz matrix.  For an adjustable step size $\tau'_{i,t}{}^{\!\hspace{-0.125cm}m} \!\in\! \mathbb{R}_+$ and inertial sequence $\beta_{m}' \!\in\! \mathbb{R}_+$, the hidden-cause inference process, indexed by $m$, is given by the following expressions
\begin{equation*}
\kappa_{i,t+1}^m \!=\! \textnormal{\sc prox}_{\!\lambda_i'}\!\Bigg(\!\pi'_{i,t}{}^{\!\!\!\!m} \!-\! \lambda_i'\tau'_{i,t}{}^{\!\hspace{-0.1cm}m}\nabla_{\pi'_{i,t}{}^{\!\hspace{-0.15cm}m}}\mathcal{L}_2(\gamma_{i,t+1},\pi'_{i,t}{}^{\!\hspace{-0.1475cm}m},G_i;\alpha_i',\lambda_i',\eta_i',\lambda_{i,t})\!\Bigg),\;\; \pi'_{i,t}{}^{\!\hspace{-0.1475cm}m+1} \!=\! \kappa_{i,t+1}^m \!+\! \beta'_m (\kappa_{i,t+1}^m \!-\! \kappa_{i,t+1}^{m-1})
\end{equation*}
with $\nabla_{\pi'_{i,t}{}^{\!\hspace{-0.15cm}m}}\mathcal{L}_2(\gamma_{i,t+1},\pi'_{i,t}{}^{\!\hspace{-0.1475cm}m},G_i;\alpha_i',\lambda_i',\eta_i',\lambda_{i,t}) \!=\! -\alpha_i' G_i^\top\textnormal{exp}(-G_i\pi'_{i,t}{}^{\!\hspace{-0.1475cm}m})|\gamma_{i,t+1}^j| \!+\! 2\eta_i'(\kappa_{i,t+1}^m \!-\! \kappa'_{i+1,t})$.  Small\\ \noindent values for the hidden causes are clamped via an implicit soft thresholding function, leading to a sparse solution.\\ \noindent The inferred causes are used to update the sparsity parameter $\lambda_{i,t+1} \!=\! \alpha_i'(1 \!+\! \textnormal{exp}(-\textnormal{\sc unpool}(G_i \kappa_{i,t+1})))$ via\\ \noindent spatial max unpooling. 
\end{itemize}

\noindent In both cases, the step size is bounded by the Lipschitz constant of the LASSO cost to be solved.  The choice of the inertial sequence greatly affects the convergence properties of the optimization.
\begin{itemize}
\item[] \-\hspace{0.5cm}{\small{\sf{\textbf{Proposition 4.}}}} Let $\gamma_{i,t} \!\in\! \mathbb{R}^{k_i}$ be the hidden states and $\kappa_{i,t} \!\in\! \mathbb{R}^{d_i}$ be the hidden causes.  The state iterates\\ \noindent $\{\gamma_{i,t+1}^m\}_{m=1}^\infty$ strongly converge to the global solution of $\mathcal{L}_1(\gamma_{i,t},\kappa_{i,t},C_i,D_i^\top;\alpha_i,\lambda_{i,t})$ for the accelerated proximal gradient scheme.  Likewise, the cause iterates $\{\kappa_{i,t+1}^m\}_{m=1}^\infty$ for the accelerated proximal gradient scheme strongly converge to the global solution of $\mathcal{L}_2(\gamma_{i,t+1},\kappa_{i,t},G_i;\alpha_i',\lambda_i',\eta_i',\lambda_{i,t})$ at a sub-polynomial rate.  This occurs when using the inertial sequences $\beta_m,\beta'_m \!=\! (k_m \!-\! 1)/k_{m+1}$, where $k_m$ depends polynomially on $m$. 
\end{itemize}



In this bottom-up inference process, there is an implicit assumption that the top-down predictions of the causes are available.  This, however, is not the case for each iteration of a mini batch being propagated through the ADPCN.  We therefore consider an approximate, top-down prediction using the states from the previous time instance and, starting from the first stage, perform bottom-up inference using this prediction.
\begin{itemize}
\item[] \-\hspace{0.5cm}{\small{\sf{\textbf{Definition 6: Feed-Back, Top-Down Cause Inference.}}}} At the beginning of every time step $t$, using the state-space model at each stage, the likely top-down causes, $\kappa'_{i-1,t} \!\in\! \mathbb{R}^{d_{i-1}}$, are predicted using the previous\\ \noindent states $\gamma_{i,t} \!\in\! \mathbb{R}^{k_{i+1}}$ and the causes $\kappa_{i,t} \!\in\! \mathbb{R}^{d_{i}}$.  That is, for the filter dictionary matrix, the following top-down\\ \noindent update is performed
\begin{equation*}
\kappa_{i-1,t}' \!=\! D_i^\top \gamma_{i,t}',\;\; \gamma_{i,t}' \!=\! \textnormal{arg min}_{\gamma_{i,t}} \Bigg(\!\lambda_i'\|\gamma_{i,t+1} \!-\! C_i\gamma_{i,t}\|_1 \!+\! \alpha_i'\|\gamma_{i,t}\|_1\Bigg(\!1 \!+\! \textnormal{exp}(-\textnormal{unpool}(G_i \kappa_{i,t}))\!\Bigg)\!\Bigg),
\end{equation*}
except for the last stage, wherein $\kappa'_{i,t+1} \!=\! \kappa_{i,t}$.  This minimization problem has an algebraic expression for the\\ \noindent global solution: $[\gamma_{i,t}']_k \!=\! [C_i\gamma_{i,t-1}]_k$, whenever $\alpha_i'\lambda_{i,t,k} \!<\! \alpha_i$, and zero otherwise.
\end{itemize}

These top-down predictions serve an important role during inference, as they transfer abstract knowledge from higher stages into lower ones. The overall representation quality is thereby improved.  The predictions also modulate the representations due to state zeroing by the sparsity hyperparameter. 


Alongside the state and cause inference is a learning process for fitting the ADPCN parameters to the stimuli.  Here, we consider gradient-descent training without top-down information, which is performed once inference has stabilized for a given mini batch.  An overview of this procedure is presented in the online appendix (see appendix A).

The inference procedure that we outline in the online appendix is different than the one proposed in \cite{ChalasaniR-conf2013a,PrincipeJC-jour2014a}.  Our lab's original DPCN relies on proximal gradients with an extra-gradient rule that is an almost-linear combination of previous state and cause iterates.  The ADPCNs leverage potentially non-linear iterate combinations.  This has the effect of taking larger steps along the error surface without diverging.  In essence, the DPCN inference process is overly conservative in its updates.  It also has issues that we outline in the online appendix (see Appendix A).  

Another change is that multiple iterations of top-down feed-back are executed in the ADPCNs.  As the recurrent processes unfold in time, the APDCN is used over and over to apply an increasing number of non-linear transformations to the stimuli.  This has the effect of simulating the propagation of the stimuli through an increasingly deeper, feed-forward network but without the overhead of adding and learning more network parameters \cite{ChenY-coll2017a}.  This promotes the formation of more expressive states and causes that quickly converge as the recurrent processes proceeds over time.  It suggests that the ADPCNs have a stable, self-organizing mechanism that minimizes surprise well \cite{FristonK-jour2010a}.  Our lab's DPCNs \cite{ChalasaniR-conf2013a,PrincipeJC-jour2014a} only consider a single feed-back iteration.  A great many hierarchical stages are needed to replicate the above behavior.  However, a lack of reasonable convergence prevents this from occurring.  They hence do not learn to preserve perceptual differences.

\subsection*{\small{\sf{\textbf{4.$\;\;\;$Simulation Results}}}}\addtocounter{section}{1}


We now assess the capability of our inference strategy for unsupervised ADPCNs.  We focus on static visual stimuli (see section 3.1).  We demonstrate that ADPCNs uncover meaningful feature representations.  They do so more quickly than convolutional DPCNs that rely on a non-accelerated, proximal-gradient update (see section 3.2).  We show that the improved inference offered by ADPCNs permits them to exceed the performance of deep unsupervised networks and behave similarly to deep supervised networks.  

\subsection*{\small{\sf{\textbf{4.1.$\;\;\;$Simulation Preliminaries}}}}

{\small{\sf{\textbf{Data and Pre-processing.}}}} We rely on five datasets for our simulations.  Two of these, MNIST and FMNIST, contain single-channel, static visual stimuli.  The remaining three, CIFAR-10, CIFAR-100 and STL-10, contain multi-channel, static visual stimuli.  We whiten each dataset and zero their means.  We use the default training and test set definitions for each dataset.  

{\small{\sf{\textbf{Training and Inference Protocols.}}}} For learning the DPCN and ADPCN parameters, we rely on ADAM-based gradient descent with mini batches \cite{KingmaDP-conf2015a}.  We set the initial learning rate to $\eta_0 \!=\! \textnormal{0.001}$, which helps prevent over-\\ \noindent shooting the global optimum.  The learning rate is decreased by half every epoch.  We use exponential decay rates of 0.9 and 0.99 for the first- and second-order gradient moments in ADAM, respectively, which are employed to perform bias correction and adjust the per-parameter learning rates.  An epsilon additive factor of $\textnormal{10}^{-\textnormal{8}}$ is used to preempt division by zero.  We use an initial forgetting factor value of $\theta_0 \!=\! \textnormal{0.7}$.  This factor is increased by a tenth every\\ \noindent thousand mini batches to stabilize convergence.  We consider a mini-batch size of 32 randomly selected samples to ensure good solution quality \cite{HardtM-conf2016a}.  

For DPCN and ADPCN inference, we primarily set the sparsity parameters to $\lambda_1,\lambda_1' \!=\! \textnormal{0.2}$, $\lambda_2,\lambda_2' \!=\! \textnormal{0.25}$, and\\ \noindent $\lambda_3,\lambda_3' \!=\! \textnormal{0.35}$.  Such values permit retaining much of the visual content in the first two stages while compressing\\ \noindent it more pronouncedly in the third.  Some stimuli datasets have slightly altered parameter values.  Due to the static nature of the stimuli, we do not have temporal state feed-back.  We do, however, propagate the cause-state difference between stages.  This is a slight modification of the network diagram shown in the previous section.  We set the feed-back strengths, for most simulations, to $\alpha_1,\alpha_2 \!=\! \textnormal{1}$ and $\alpha_3 \!=\! \textnormal{3}$.  In \cref{fig:dpcn-network}, these variables correspond to the strength of the temporal, intra-layer feed-back, but here they are used for non-temporal, inter-layer state feed-back.  The stronger feed-back amount in the third stage aids in suppressing noise without adversely impacting the in earlier stages' priors.  We fix the causal sparsity constants to $\alpha_1',\alpha_2',\alpha_3' \!=\! \textnormal{1}$.  We terminate the accelerated and regular inference processes after 500 and 1000 iterations, respectively, per mini batch.  A significantly lower number of iterations is used in the former case, since the new inertial sequence facilitates quick convergence.

\begin{figure*}
   \hspace{-0.175cm}\includegraphics[height=4.7in]{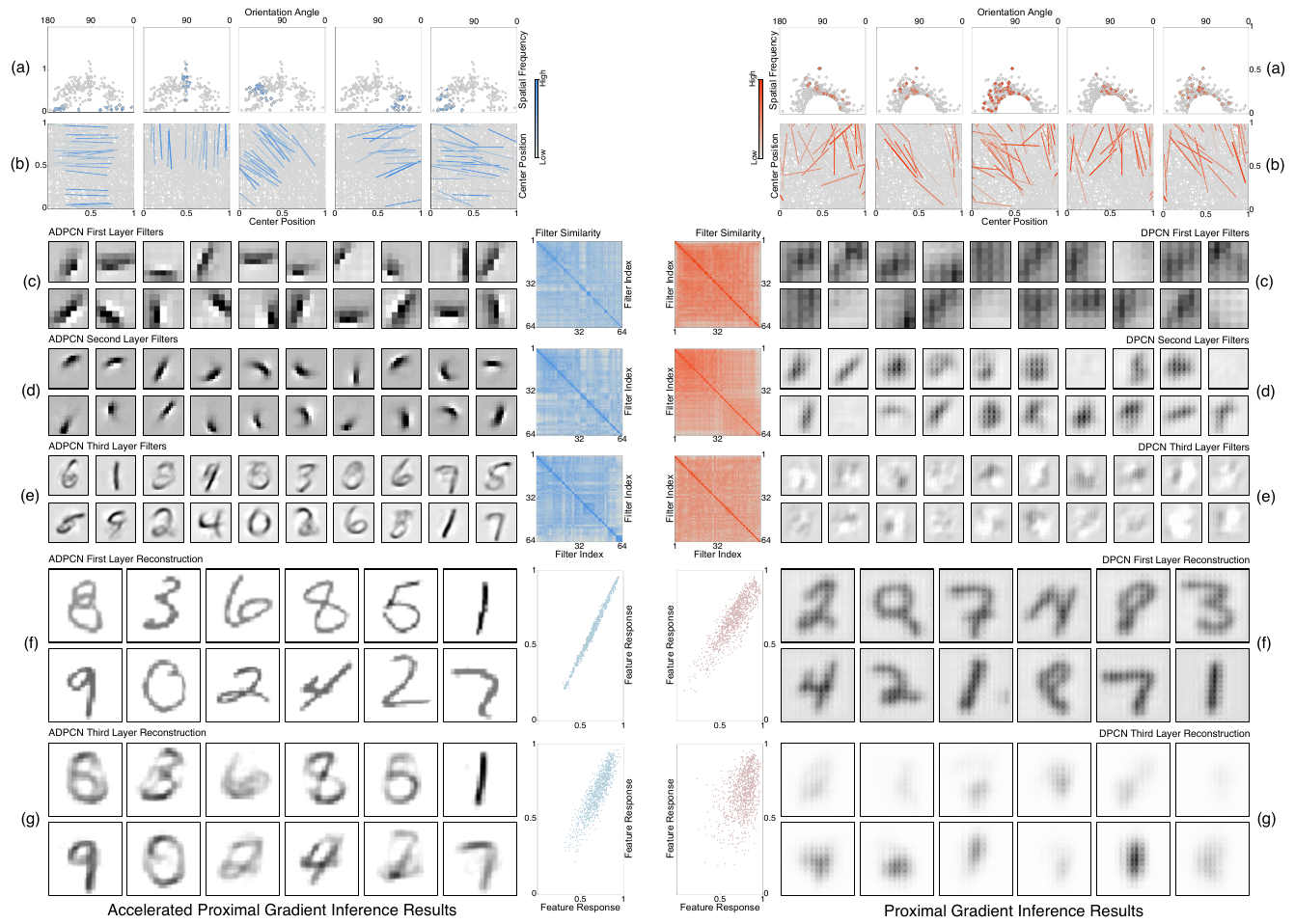}




   \caption[]{\fontdimen2\font=1.55pt\selectfont A comparison of accelerated proximal gradient inference and learning (left, blue) and proximal gradient inference and learning (right, red) the MNIST dataset.  The presented results are shown after training with mini batches for two epochs.  (a)  Polar scatter plots of the orientation angles versus spatial frequency for the first-stage causes.  (b) Line plots of the normalized center positions with included orientations for the first-stage causes.  For both (a) and (b), we fit Gabor filters to the first-stage causes; locally optimal filter parameters were selected via a gradient-descent scheme.  The plots are color-coded according to the connection strength between the invariant matrix and the observation matrix in the first network stage.  Higher connection strengths indicate subsets of dictionary elements from the observation matrix that are most likely active when a column of the invariance matrix is also active.  If a DPCN has been trained well, then the filters should have a small orientation-angle spread.  Each plot represents a randomly chosen column of the first-stage invariance matrix. (c)--(e) Back-projected causes from the first, second, and third stages of the networks, respectively.  Each plot represents a randomly chosen cause.  The back-projected causes can be interpreted as receptive fields, with darker colors indicating a higher degree of activation.  For each stage, we assess the filter similarity and provide VAT similarity plots.  In these plots, low similarities are denoted using gray while higher similarities are denoted using progressively more vivid shades of either blue (accelerated proximal gradients) or red (proximal gradients).  If a DPCN has been trained well, then there should be few to no duplicate filters.  There, hence, should not be any conspicuous blocky structures along the main diagonal of the VAT similarity plots.  (f)--(g) Reconstructed instances from a random batch at the first and third stage, respectively.  For each stage, we also assess the feature similarity between the original training samples and the reconstructed versions and provide corresponding scatter plots.  If a DPCN reconstructs the input samples well, then there should be a strong linear relationship between the features.  Higher distributional spreads and shifts away from the main diagonal indicate larger reconstruction errors.\vspace{-0.4cm}}
   \label{fig:mnist-results}
\end{figure*}

\begin{figure*}
   \hspace{-0.175cm}\includegraphics[height=4.7in]{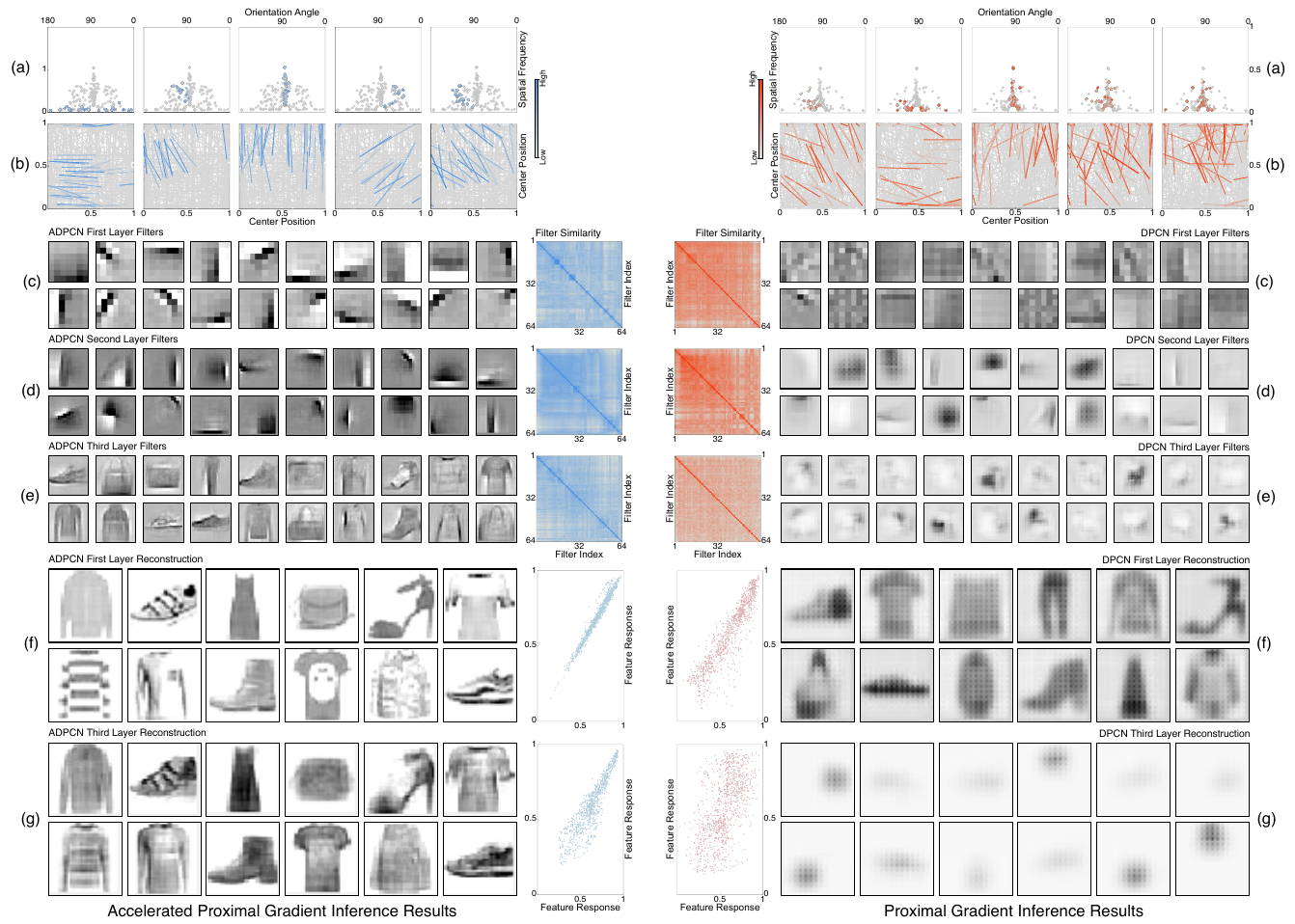}
   \caption[]{\fontdimen2\font=1.55pt\selectfont A comparison of accelerated proximal gradient inference and learning (left, blue) and proximal gradient inference and learning (right, red) the FMNIST dataset.  The presented results are shown after training with mini batches for two epochs.  See \cref{fig:mnist-results} for descriptions of the plots.\vspace{-0.4cm}}
   \label{fig:fmnist-results}
\end{figure*}

\subsection*{\small{\sf{\textbf{4.2.$\;\;\;$Simulation Results and Discussions}}}}



{\small{\sf{\textbf{Network Architecture.}}}} As we note in the previous section, our lab's DPCNs are non-convolutional.  For our simulations, we consider the same Toeplitz-based convolution as the ADPCNs.  This is done to highlight that the improvement in feature quality occurs due to the faster inference strategy.

We consider the same architecture for the convolutional DPCNs and ADPCNs.  At the first stage, we use 128 states with $\textnormal{5} \!\times\! \textnormal{5}$ filters and 256 causes with $\textnormal{5} \!\times\! \textnormal{5}$ filters.  This yields an over-complete reconstruction basis.  At the second and third stages, we use 128 states and 256 states with $\textnormal{5} \!\times\! \textnormal{5}$ and $\textnormal{5} \!\times\! \textnormal{5}$ filters, respectively.  For these two stages, we use 512 causes and 1024 causes with $\textnormal{5} \!\times\! \textnormal{5}$ and $\textnormal{5} \!\times\! \textnormal{5}$ filters, respectively.  We perform $\textnormal{2} \!\times\! \textnormal{2}$ max pooling between the states and the causes at each stage.  

Due to the choice of filter sizes, the ADPCNs will typically have a worse reconstruction error but better recognition rate in the later stages.  Using larger filters in the early stages and smaller ones in the later stages permits implementing traditional predictive-coding behaviors.  That is, the reconstruction error decreases deeper in the hierarchy, as parts of the stimuli are explained away in an iterative manner.

\begin{figure*}
   \hspace{-0.175cm}\includegraphics[height=4.7in]{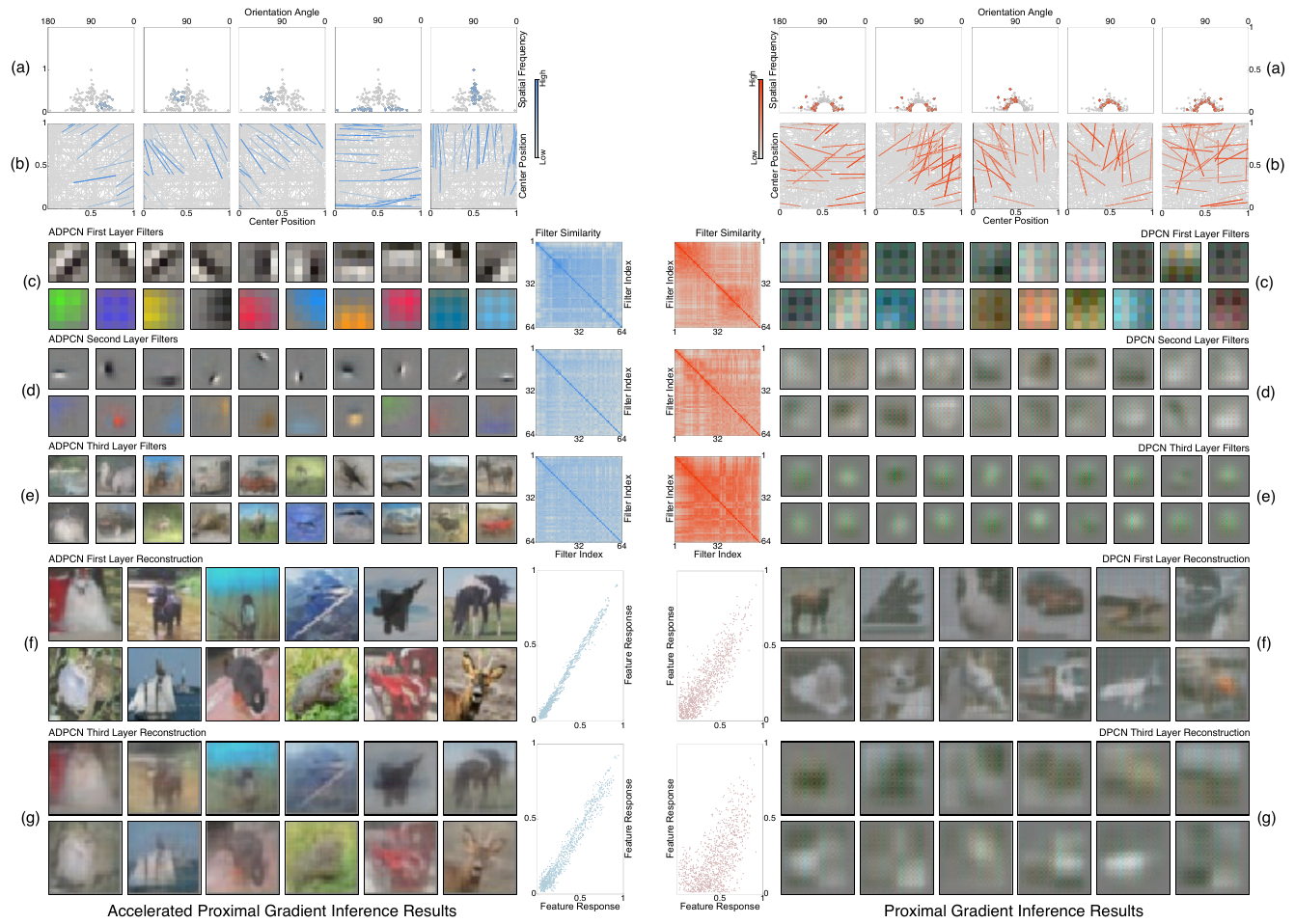}
   \caption[]{\fontdimen2\font=1.55pt\selectfont A comparison of accelerated proximal gradient inference and learning (left, blue) and proximal gradient inference and learning (right, red) the CIFAR-10/100 datasets.  The presented results are shown after training with mini batches for two epochs.  See \cref{fig:mnist-results} for descriptions of the plots.\vspace{-0.4cm}}
   \label{fig:cifar10-results}
\end{figure*}

\begin{figure*}
   \hspace{-0.175cm}\includegraphics[height=4.7in]{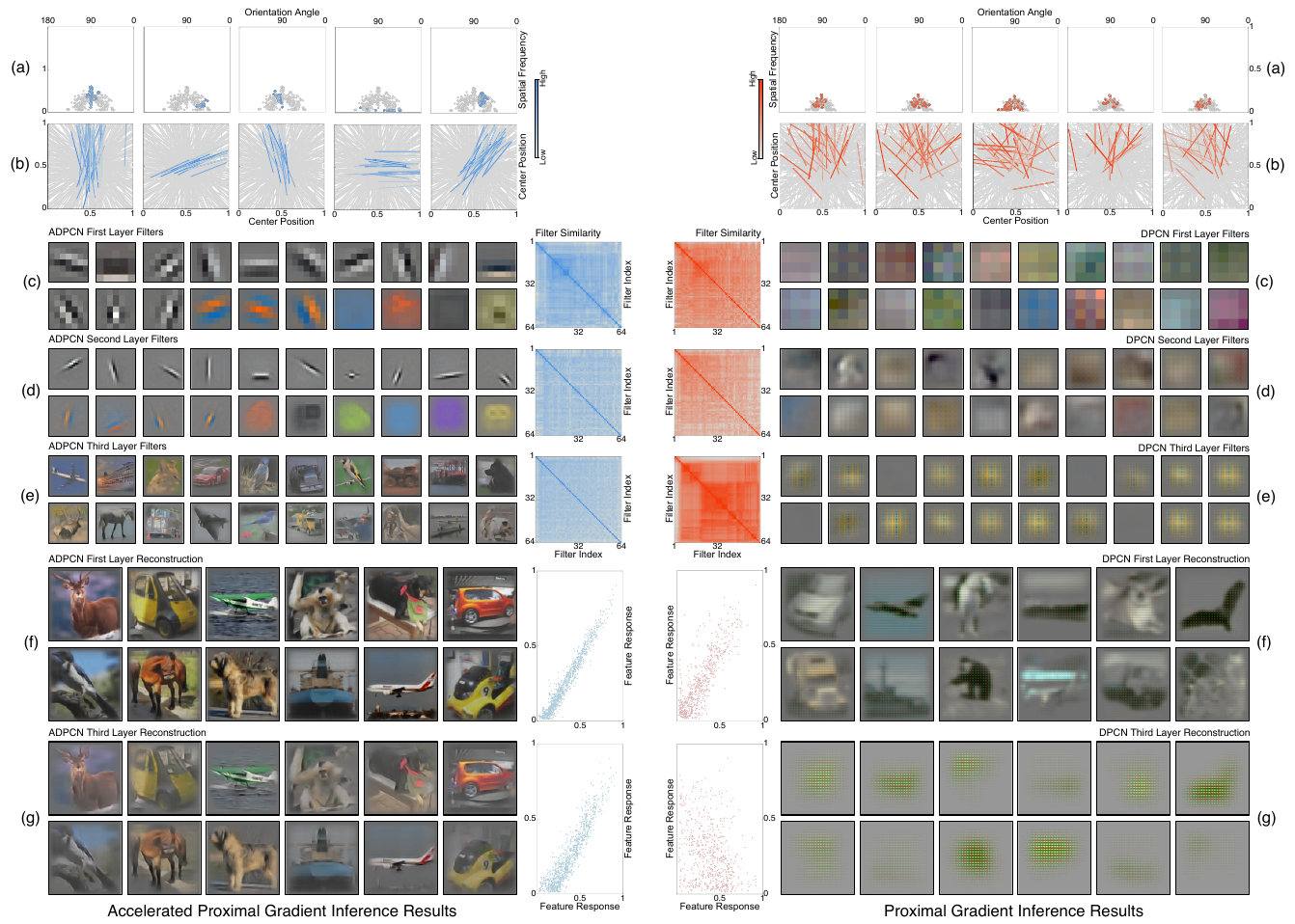}
   \caption[]{\fontdimen2\font=1.55pt\selectfont A comparison of accelerated proximal gradient inference and learning (left, blue) and proximal gradient inference and learning (right, red) the STL-10 dataset.  The presented results are shown after training with mini batches for two epochs.  See \cref{fig:mnist-results} for descriptions of the plots.\vspace{-0.4cm}}
   \label{fig:stl10-results}
\end{figure*}

{\small{\sf{\textbf{Simulation Results.}}}} Simulation findings are presented in \cref{fig:mnist-results} and \cref{fig:fmnist-results} for the single-channel MNIST and FMNIST datasets, respectively.  Findings for the multi-channel CIFAR-10/100 and STL-10 datasets are, respectively, shown in \cref{fig:cifar10-results} and \cref{fig:stl10-results}.  The results in these figures were for after two epochs.

For these datasets, the ADPCNs are successful in quickly uncovering invariant representations.  Most of the columns in the invariance matrix group dictionary elements that have very similar orientation and frequency while being insensitive to translation (see \cref{fig:mnist-results}(a) to \cref{fig:stl10-results}(a)).  Likewise, for each active invariance-matrix column, a subset of the dictionary elements are grouped by orientation and spatial position, which indicates invariance to other properties like spatial frequency and center position (see \cref{fig:mnist-results}(b) to \cref{fig:stl10-results}(b)).  The convolutional DPCNs, in comparison, have representations that are significantly altered by transformations other than translation.  This occurs because subsets of the dictionary elements are not grouped according to various characteristics.  Discrimination performance hence can suffer for stimuli samples that are slightly altered.

As well, the ADPCNs learn meaningful filters from the stimuli.  The first two stages of our ADPCNs have causal receptive fields that mimic the behavior of simple and complex cells in the primate vision system (see \cref{fig:mnist-results}(c)--(d) to \cref{fig:stl10-results}(c)--(d)).  The fields for the first stage are predominantly divided into two types: low-frequency and high-frequency, localized band-pass filters.  The former mainly encode regions of uniform intensity and color along with slowly varying texture.  The latter describe contours and hence sharp boundaries.  Such filters permit accurately reconstructing the input stimuli (see \cref{fig:mnist-results}(f)--\cref{fig:stl10-results}(f) and \cref{fig:recog-tables}(a)--(e)).  The second-stage receptive fields are non-linear combinations of those in the first that are activated by more complicated visual patterns, such as curves and junctions.  A similar division of receptive fields into two categories is often encountered in the second ADPCN stage.  More filters are activated by contours, however, than in the first stage.  For both stages, the filters are mostly unique, which is captured in the ordered similarity plots (see \cref{fig:mnist-results}(c)--(d) to \cref{fig:stl10-results}(c)--(d)).  Beyond two stages, the ADPCN receptive fields encompass entire objects (see \cref{fig:mnist-results}(e) to \cref{fig:stl10-results}(e)).  They are, however, average representations, not highly specific ones, due to the limited number of causes (see \cref{fig:mnist-results}(g) to \cref{fig:stl10-results}(g)).  The backgrounds in the visual stimuli are often suppressed at the third stage for CIFAR-10/100 and STL-10, which greatly enhance recognition performance (see \cref{fig:recog-tables}(c)--(e)).  There are no backgrounds for MNIST and FMNIST, so recognition is predominantly driven by the whole-object receptive fields in the third stage see \cref{fig:recog-tables}(a)--(b)).  The ordered similarity plots indicate that none of the third-stage filters appear to be duplicated for either dataset.  This trend also holds for the first- and second-stage receptive fields.  This implies that the ADPCNs emphasize the extraction of non-redundant features to form a complete visual stimuli basis at each stage.

The filters learned by ADPCNs are not only sensitive to whole objects.  They also are attuned to stylistic changes of objects.  As shown in \cref{fig:tsne-embedding-1}, the filter-derived cause features tend to segregate instances of objects that visually differ.  This occurs even within a given object class.


Convolutional DPCNs, in contrast, do not stabilize to viable receptive fields at the same rate as the ADPCNs (see \cref{fig:mnist-results}(c)--(d) to \cref{fig:stl10-results}(c)--(d)).  For MNIST, the first-stage DPCN receptive fields have some localized band-pass structure that is similar to Gabor filters.  The overall spread of the fields makes it difficult to accurately detect abrupt transitions and hence recreate the input stimuli, though.  The reconstructions thus are heavily distorted and blurred (see \cref{fig:mnist-results}(f) and \cref{fig:recog-tables}(a)).  For FMNIST, the first-stage receptive fields focus on low-frequency details, such as either constant grayscale values or slow-changing grayscale gradients.  They also focus onr higher-frequency details, such as periodic texture.  While some of the causes become specialized band-pass-like filters, there are not enough to adequately preserve sharp edges.  The stimuli reconstructions are thus also distorted, which removes much of the high-frequency content (see \cref{fig:fmnist-results}(f) and \cref{fig:recog-tables}(b)).  Similar results are encountered for CIFAR-10/100 and STL-10 (see \cref{fig:cifar10-results}(f)--\cref{fig:stl10-results}(f) and \cref{fig:recog-tables}(c)--(e)).  For all of the datasets, the second-stage receptive fields become even less organized than in the first stage.  They are mostly activated by blob-like visual patterns, which do not preserve enough visual content for recreating a close resemblance of the input stimuli beyond the first stage.  The convolutional DPCNs are unable to learn relevant representations in the third stage, as a consequence.  The receptive fields for this network stage are unique, by virtue of being essentially random.  They are, however, largely useless in extracting stimuli-specific details (see \cref{fig:mnist-results}(e) to \cref{fig:stl10-results}(e)).  This further degrades the reconstruction quality to where the inputs are unrecognizable (see \cref{fig:mnist-results}(g) to \cref{fig:stl10-results}(g)).  Discrimination is adversely impacted due to this severe lack of identifying characteristics (see \cref{fig:recog-tables}(a)--(e)).  The filter redundancy also impacts learning good model priors.  There is typically not enough unique information to be back-propagated to earlier stages to inform the choice of better receptive fields.

The causes formed by the ADPCNs and DPCNs specify invariant features that can be employed for discrimination.  In \cref{fig:class-results}, we show that the stage-aggregated ADPCN features yield high-performing unsupervised classifiers.  They achieve state-of-the-art unsupervised recognition rates for each dataset.  These recognition rates also are on par with deep networks trained in a supervised fashion, despite having orders of magnitude fewer parameters.  Although the causal features from all stages have a positive net contribution, those from the third stage contribute the most to recognition performance (see \cref{fig:recog-tables}(a)--(e)).  The DPCNs exhibit poor performance, in comparison.  Only the first-stage features aid classification (see \cref{fig:recog-tables}(a)--(e)).  The remainder largely worsen the recognition capabilities.  For both the DPCNs and ADPCNs, we rely on a seven-nearest neighbor classifier with an unsupervised-learned metric distance \cite{SenerO-coll2016a} to label the stimuli samples.  As we illustrate in \cref{fig:tsne-embedding-1}, the causal features require non-linear decision boundaries to distinguish between classes well.  This occurs even in high-dimensional spaces, as class overlap occurs.  Such a property motivates the use of nearest-neighbor classifiers.




{\small{\sf{\textbf{Simulation Discussions.}}}} Our simulations indicate that ADPCNs were more effective at uncovering highly discriminative feature representations of visual stimuli than the original DPCN inference strategy.  A trait that contributes greatly to the ADPCNs' success is its significantly improved search rate.  

As noted in the appendix, proximal-gradient-type schemes can undergo four separate search phases, some of which have different local convergence rates.  In one of the phases, the constant-step regime, both the states and causes undergo rapid improvements.  However, in two phases, the local convergence rate is slow whenever the largest eigenvalue of a certain recurrence matrix is less than the current inertial-sequence magnitude.  For linear inertial sequences, like those found in the proximal-gradient-based DPCNs, this condition occurs early during the optimization process.  That is, for such sequences, the growth is initially very rapid and follows a logarithmic rate.  Within just a few iterations, the sequence magnitude exceeds the eigenvalue, which preempts the fast constant-step regime.  The rate of convergence becomes worse than sublinear.  A large number of search steps is thus needed to move toward the global solution.  However, the search frequently terminates before this happens due to reaching the maximum number of proximal-gradient iterations.  The search, alternatively, stops early due to a lack of progress across consecutive iterations.  In either case, the states and causes do not adequately stabilize for a given mini batch.  The poor state and cause representations, naturally, are integrated into the filter dictionary matrices and invariant matrices during the learning updates.  This disrupts the priors in the early hierarchy for future stimuli.  The convergence is further stymied by cost rippling.  Proximal-gradient-based optimization does not behave like a pure descent method in two out of the four phases.  Such behavior is caused by the eigenvalues of another recurrence matrix being a pair of complex conjugates, which necessitates oscillating between the two.  All of these factors make it difficult to propagate meaningful bottom-up information \cite{MechelliA-jour2004a} beyond the first stage.  The top-down details from higher stages are hence ineffective at modifying the priors to disambiguate the stimuli \cite{BarM-jour2004a}.  

The ADPCNs largely avoid these issues.  After the constant-step regime, the search switches to one of two potentially slower phases.  However, the APDCNs' inertial-sequence growth rate is rather muted, as opposed to that of the DPCNs.  The chance of exceeding the largest eigenvalue of the augmented, auxiliary-variable recurrence matrix is low for the ADPCNs.  The searches thus can proceed unhindered toward the global solution.  Moreover, since the eigenvalue-magnitude threshold is often not reached, the largest eigenvalue of the augmented, mapping recurrence matrix is real-valued, not complex.  This means that the accelerated proximal gradients behave like a descent method.  It thus will not experience localized cost rippling due to alternating between conjugate pairs.  Both properties promote rapid stabilization of the states and causes, which expedite the formation of beneficial priors throughout much of the early-stage network inference.  These priors facilitate the construction of transformation-insensitive feature abstractions in deeper network stages, due to the bottom-up forwarding of stimuli-relevant signals.  The recurrent, top-down connections, in turn, suitably alter the representation sparsity.  Extraneous details that do not contribute greatly to the reconstruction quality are hence ignored.  This behavior biases in favor of valuable stimuli, which is analogous to observed functionality in the frontal and parietal cortices \cite{SerencesJT-jour2008a}.  The descending pathways also carry predictive responses, in the form of templates of expected stimuli, that are matched to the current and future mini batches to aid recognition \cite{UllmanS-jour1995a}.  That is, the templates convey contextual information for extracting task-specific information, which makes recognition more reliable \cite{PotterMC-jour1975a,PalmerSE-jour1975a}.  


There are other traits that contribute to the success of the ADPCNs.  For instance, the ADPCN's first-stage receptive fields are largely similar to that of simple cells in the primary visual cortex.  Simple cells often implement Gabor-like filters with generic preferred stimuli that correspond to oriented edges \cite{HubelDH-jour1965a,HubelDH-jour1968a}.  Such filters are useful within the ADCPN.  Relations between activations for specific spatial locations tend to be distinctive between objects in visual stimuli.  Activations are also obtained, in a Gabor space, that facilitate the construction of naturally sparse representations.  These representations can then be hierarchically extended, which is what we ultimately sought in the ADPCNs.  Changes in object location, scale, and orientation can be reliably detected, within this Gabor space, thereby aiding in the creation of transformation-invariant features.  These features permit near-perfect stimuli reconstructions at the first stage.  The uniqueness of the band-pass filters is beneficial, as it permits the ADPCNs to fixate on non-redundant stimuli characteristics.  How these filters arise and the rate at which they form are also important aspects that aid in the ADPCNs' success.  They emerge due to feed-back from high-level visual network areas, which quickly reduce activity in lower areas to simplify the description of the stimuli to some of its most basic elements \cite{MurraySO-jour2004a}.  In doing so, alternate explanations are suppressed, and only the most dominant, fundamental causes of the stimuli remain \cite{SpratlingMW-jour2011a}.  These causes are oriented luminance contours.  This rapid stabilization of filters is exactly functionality that we expect to encounter in an efficient predictive coding process.  It is hence well aligned with contemporary neurophysiological theory \cite{OlshausenBA-jour1997a}.

The plots of center position and orientation angle versus center frequency for the ADPCNs indicate that the learned Gabor filters form a meaningful reconstruction basis for the visual stimuli.  The entire center-position plane is thoroughly filled.  There hence are filter impulse responses everywhere in space when a sufficient number of first-stage filters are learned.  Likewise, the Gabor orientation range is well covered, implying that the filters can recognize edges at different angles.  Taken together, both properties indicate that these first-stage filters can reconstruct the stimuli well.  Given the similarities of the Gabor filters across the different stimuli datasets, it would appear that the inferred basis can recreate any natural-image stimuli well.  The low reconstruction errors for the first stage of the APDCN corroborate this claim.  For the DPCNs, we also have that the center-position plane is well covered.  However, the filters have poor spatial frequency bandwidths, so they are not activated much by contours.  Only the most dominant edges in the visual stimuli will be well preserved.  These usually are the object-background boundaries.  Inter-object details are not retained.  While the first-stage DPCN filters also form a basis, it is a poor one for reconstructing natural-image stimuli.  Without a meaningful, stable basis, the remainder of the network cannot build on it to reliably describe aspects of the stimuli.  The slow convergence of proximal gradients prevents this, as we note above.

Beyond the first stage, the ADPCNs exhibit an increase in specificity, abstraction, and invariance for deeper hierarchies \cite{RiesenhuberM-jour1999a}.  This behavior aligns well with the current understanding of the vision system \cite{TjanBS-jour2005a}.  It also aids in the ADPCNs' success.  The second-stage receptive fields, in the case of MNIST, become sensitive to curved sub-strokes for the hand-written digits, which is similar to prestriate cortex functionality \cite{HedgeJ-jour2000a}.  For FMNIST, they emphasize abrupt transitions found in regions of the apparel and fashion accessories.  This mimics aspects of shape-pattern-selectivity behaviors found in the extrastriate cortices \cite{PasupathyA-jour1999a,PasupathyA-jour2002a}.  For STL-10, the receptive fields are often elongated Gabor-like filters, which can be found in the prestriate cortex \cite{LiuL-jour2016b}.  All of these features systematically focus on object-relevant visual details, thereby aiding recognition.  At the deepest stage, the receptive fields are entirely object specific, which is functionality somewhat akin to the neurons in the primate infero-temporal cortex \cite{BruceC-jour1981a}.  It is also related to memory activity \cite{MiyashitaY-jour1988a,YakovlevV-jour1998a}.  The representations are additionally translation and rotation insensitive, similar to inferotemporal cortex neurons \cite{RollsET-jour1992a,ToveeMJ-jour1994a,SalinasE-jour1997a}, and change little with respect to scale and spatial-frequency changes, similar to neurons in the middle temporal area \cite{RollsET-jour1985a,LiuL-jour2007a}.  To our knowledge, such invariant, whole-object sensitivity within a single stage has not been witnessed in any existing predictive-coding model.  Based on contemporary theories \cite{LiangH-jour2017a}, we believe that the receptive-field feed-back from the final stage contribute to the effective connectivity in the earlier network stages \cite{RamalingamN-jour2013a}.  That is, the role of this final stage is to disambiguate local image components.  This is done by creating a template that is fed back, which then selectively enhances object components and suppresses interfering background components \cite{EpshteinB-jour2008a}.  This biologically plausible behavior is crucial for achieving high discrimination rates on CIFAR-10/100 and STL-10.  For these datasets, the objects of interest are scattered amongst distractors.  

As we show in our simulation results, the third stage of the ADPCN contributes more to recognition than the first two stages.  Such a finding is well-aligned with previous studies.  Although simple objects can be discriminated based on Gabor-like filters, complex objects require significant non-linearities \cite{ShiJV-jour2013a}.  The later stages provide this functionality due to the conversion from states to pooled, sparse causes.  The processes of pooling and sparsifying are highly non-linear.  They additionally lead to the emergence of complex-cell properties along with hence shift invariance \cite{HubelDH-jour1962a} and spatial phase invariance \cite{LianY-jour2021a}.



Including Toeplitz-based convolution with spatial pooling has a major impact on performance.  Spatial coherence behaviors that convolutional layers offer are essential for describing inter-object relationships in complex visual stimuli.  Pooling introduces invariance.  Our promising recognition results in Appendix B are a testament to this.  The corresponding cause embeddings are too.  They indicate that the ADPCNs self-organize the states and causes in an object-sensitive way, despite the lack of supervision.  DPCNs struggle to do this, since they rely on non-convolutional filters \cite{ChalasaniR-conf2013a,PrincipeJC-jour2014a}.  Their cause embeddings often group the stimuli in a worse way than embeddings of the input stimuli.  That is, their transformation of the stimuli often destroys much of the visual content.  Even the convolutional extension to DPCNs struggle due to a lack of a good reconstruction basis.  This shows that convolution alone is not sufficient to uncover good features.  The networks must be paired with an efficient inference process.  Inference for the convolutional and non-convolutional DPCNs is based on a slow, proximal-gradient-based approach.  The later-stage cause embeddings for these networks are thus quite poor, since the first- and second-stage bases do not stabilize quickly to meaningful filters.  It is likely that DPCNs would benefit from a greedy, stage-wise training to initialize the networks in a meaningful way.  This may overcome the poor convergence, to some extent.  Our ADPCNs would likely benefit from it too.  As well, the ADPCNs would probably benefit from fixing the first-stage states and causes.  As we have shown in our simulations, these features do not change much across color-image datasets.  Effort could be better spent on learning more advanced features in the later network stages.

Another benefit of including convolutional filters in the ADPCNs is that they help encode perceptual stimuli similarity and differences.  By this, we mean that well-trained ADPCNs group related stimuli together and force unrelated stimuli apart in the causal feature space.  Such behavior is stems from learning visual, appearance-based features at local and global object scales.  It also stems from iteratively sparsifying those features to selectively remove redundant details.  Inter-layer and intra-layer feed-back contribute too.   Other network architectures, like convolutional autoencoders, should also be capable of accounting for perceptual differences.  They, however, appear to currently lack both appropriate losses and regularizers, along with fundamental layer behaviors, to do this as effectively as ADPCNs.

\subsection*{\small{\sf{\textbf{5.$\;\;\;$Conclusions}}}}\addtocounter{section}{1}

Here, we have revisited the problem of unsupervised predictive coding.  We have considered a hierarchical, generative network, the ADPCN, for reconstructing observed stimuli.  This network is composed of temporal sparse-coding models with bi-directional connectivity.  The interaction of the information passed by top-down and bottom-up connections across the models permit the extraction of invariant, increasingly abstract feature representations.  Such representations preserve perceptual differences.

Our contribution in this paper is a new means of inferring the underlying components of the feature representations, which are the sparse states and causes.  Previously, a proximal-gradient-type approach has been used for this purpose.  Despite its promising theoretical guarantees, though, it exhibits poor empirical performance.  This practically limits the number of model stages that can be considered in DPCNs.  It also extensively curtails the quality of the stimuli features that can be extracted.  Here, we have considered a parallelizable, vastly accelerated proximal-gradient strategy that overcomes these issues.  It allows us to go beyond the existing two-stage limitation, facilitating the construction of arbitrary-staged DPCNs.  Each stage leads to increasingly enhanced performance for object analysis.  The information from higher stages is propagated to earlier ones to form more effective stimuli priors for bottom-up processing.  Most crucially, our optimization strategy immensely streamlined inference.  Often, only one or two presentations of the stimuli are necessary to reach a stable filter dictionary and hence a corresponding set of sparse states and sparse causes.  Good object analysis performance is hence observed.  The previous optimization approach requires many times more presentations before a stable set of filters is uncovered.  The resulting features are not nearly as discriminative as those from our proposed strategy.  We have mathematically proved that this stems from poor convergence-rate properties of the original inertial sequence used in DPCNs.

We have applied ADPCNs to static-image datasets.  For MNIST and FMNIST, the ADPCNs learn initial-stage receptive fields that mimic aspects of the early stages of the primate vision system.  The later network stages implement receptive fields that encompass entire objects.  In the case of MNIST, the back-projected filters become pseudo averages of the hand-written numerical digits.  Predominant writing styles are modeled well.  For FMNIST, the back-projected filters are the various types of clothing and personal articles.  General styles and some nuances are captured.  To our knowledge, this is the first time that such object-scale receptive fields have been learned for predictive coding.  This behavior helps yield unsupervised classifiers that achieve state-of-the-art performance.  Such classifiers also outperform supervised-trained deep networks, which lends credence to the complicated feature inference and invariance process that we employ.  It also supports the notion that the ADPCNs preserve perceptual differences.  Similar results are witnessed for natural-image datasets, such as CIFAR-10/100.  The later-stage receptive fields again encompass entire object categories.  This yields promising features that achieve generalization rates almost equivalent to that of supervised-trained deep networks.  This is despite our use of simple nearest-neighbor classifiers.  It is also despite the fact that our ADPCNs have many times fewer convolutional filters than these other deep networks.

ADPCNs are readily applicable to spatio-temporal stimuli processing, much like the original DPCNs.  This is because the ADPCNs implement a recurrent state-space model.  For such modalities, like video, the ADPCN recognition performance may be even better than for static stimuli, like images.  This is because top-down, feed-back connections impose temporal constraints on the learning process.  We will investigate this in our future work from the context of solving time-varying problems.  We will also demonstrate their superiority compared to purely feed-forward convolutional-recurrent autoencoders.  We will show that the lack of a top-down pathway impedes learning meaningful representations.

ADPCNs can also form the backbone of a general, unsupervised framework for stimuli learning and high-level understanding.  In our future work, we will illustrate that architectural extensions of it are well suited for making extrapolations about environment dynamics.  This has implications for many applications, including self-driving cars.  In particular, we will show that these modified ADPCNs learn temporal and perceptual cues that permit predicting egocentric and allocentric events in the short-term future.  For the egocentric events, the ADPCNs understand well that a moving camera will cause non-linear transformations of visual characteristics across video frames.  The ADPCNs thus can thus reliably predict, from a short history of video frames, the appearance of non-mobile objects and their locations in subsequent frames.  This will be useful for estimating vehicle position and angle relative to objects of interest, which can help refine steering control inputs for close-quarters maneuvers.  For the allocentric events, the ADPCNs understand well the dynamics of mobile objects for either a fixed or moving camera.  The ADPCNs can hence similarly predict the appearance and location of those objects.  This will prove invaluable for gauging how pedestrians may react and hence avoid collisions.  We will show that both types of event predictions emerge due to the ADPCNs learning at multiple time scales across different stages.  Multi-time-scale prediction is something that existing recurrent models struggle to do well.

\setstretch{0.95}\fontsize{9.75}{10}\selectfont
\putbib
\end{bibunit}



\clearpage\newpage
\begin{bibunit}
\bstctlcite{IEEEexample:BSTcontrol}

\RaggedRight\parindent=1.5em
\fontdimen2\font=2.1pt\selectfont
\singlespacing
\allowdisplaybreaks

\subsection*{\small{\sf{\textbf{Appendix A}}}}

Below, we outline the ADPCN training and inference process.

\begin{figure*}[h!]
\vspace{-0.6cm}
{\singlespacing\begin{algorithm}[H]
\DontPrintSemicolon
\SetAlFnt{\normalsize} \SetAlCapFnt{\small}
\caption{Accelerated Deep Prediction Network (ADPCN) Training and Inference}
\AlFnt{\small} \fontdimen2\font=1.6pt\selectfont {\bf Inputs}: Initial dictionary matrix $D_i^\top \!\in\! \mathbb{R}^{k_i \times k_{i-1}}_+$, state-transition matrix $C_i \!\in\! \mathbb{R}^{k_i \times k_i}$, and invariant matrix $G_i \!\in\! \mathbb{R}^{d_i \times k_i}$ for each network layer.  A set of time-varying, static stimuli $Y_t$, where $y_t \!\in\! Y_t$, $y_t \!\in\! \mathbb{R}^p$.  A set of initial states $\gamma_{i,0} \!\in\! \mathbb{R}^{k_i}$ and causes $\kappa_{i,0} \!\in\! \mathbb{R}^{d_i}$.\;

\AlFnt{\small} \For{$t \!=\! 0,1,2,\ldots$}{
\AlFnt{\small}    \fontdimen2\font=1.6pt\selectfont Initialize the bottom-up cause for the first layer as $\kappa_{0,t} \!=\! y_t$.\;
\AlFnt{\small}    \For{$i \!=\! 0,1,2,\ldots$}{
\AlFnt{\small}       \fontdimen2\font=1.6pt\selectfont For all layers but the last, the most likely top-down causes, $\kappa'_{i-1,t} \!\in\! \mathbb{R}^{d_{i-1}}$, are initialized at each iteration using the next stage's states $\gamma_{i,t} \!\in\! \mathbb{R}^{k_{i+1}}$ and the causes $\kappa_{i,t} \!\in\! \mathbb{R}^{d_{i+1}}$,\vspace{-0.2cm}
$$\kappa_{i-1,t}' \!=\! D_i^\top \gamma_{i,t}',\;\; \gamma_{i,t}' \!=\! \textnormal{arg min}_{\gamma_{i,t}} \Bigg(\!\lambda_i'\|\gamma_{i,t+1} \!-\! C_i\gamma_{i,t}\|_1 \!+\! \alpha_i'\|\gamma_{i,t}\|_1\Bigg(\!1 \!+\! \textnormal{exp}(-\textnormal{\sc unpool}(G_i \kappa_{i,t}))\!\Bigg)\!\Bigg).\vspace{-0.15cm}$$
This minimization problem has an algebraic expression for the global solution: $[\gamma_{i,t}']_k \!=\! [C_{i,t}\gamma_{i,t-1}]_k$, whenever $\alpha_i'\lambda_{i,t,k} \!<\! \alpha_i$, and zero otherwise.  For the last layer, $\kappa'_{i,t+1} \!=\! \kappa_{i,t}$.
\AlFnt{\small}    }
\AlFnt{\small}    \For{$i \!=\! 0,1,2,\ldots$}{
\AlFnt{\small}       \fontdimen2\font=1.6pt\selectfont Let $\beta_m \!\in\! \mathbb{R}_+$ be an inertial sequence, $\beta_m \!=\! (k_m \!-\! 1)/k_{m+1}$, where $k_m \!=\! 1 \!+\! (m^r \!-\! 1)/d\,$ with $r \!\in\! \mathbb{R}_{1,+}$ and $d \!\in\! \mathbb{R}_{+}$.  Given an adjustable step size $\tau_{i,t}^m \!\in\! \mathbb{R}_+$, update the states using proximal-gradient steps, indexed by $m$, until either convergence or a pre-set number of iterations has been reached\vspace{-0.2cm}
$$\gamma_{i,t+1}^{m} \!=\! \textnormal{\sc prox}_{\!\lambda_{i,t}}\!\Bigg(\!\pi_{i,t}^{m} \!-\! \lambda_{i,t}\tau^m_{i,t}(D_i^\top(\kappa_{i-1,t} \!-\! D_i \pi^{m}_{i,t}) \!+\! \alpha_i\Omega_i(\pi^{m}_{i,t}))\!\Bigg),\vspace{-0.2cm}$$
where $\pi_{i,t}^{m+1} \!=\! \gamma^{m}_{i,t+1} \!+\! \beta^m (\gamma^{m}_{i,t+1} \!-\! \gamma_{i,t+1}^{m-1})$.  The term $\Omega_i(\pi_{i,t}^{m})$ quantifies the contribution of the non-smooth state transition.  Use Nesterov smoothing, with $\mu_i \!\in\! \mathbb{R}_+$, to approximate them, $\Omega_i(\pi_{i,t}) \!=\! \textnormal{arg max}_{\|\Omega_{i,t}\|_\infty \leq 1} \Omega_{i,t}^\top (\pi_{i,t}^m \!-\! C_i \gamma_{i,t-1}) \!-\! \mu_i\|\Omega_{i,t}\|^2_2/2$.  Max-pool the states using non-overlapping windows $\gamma_{i,t+1} \!=\! \textnormal{\sc pool}(\gamma_{i,t+1})$.\;


\AlFnt{\small}       \fontdimen2\font=1.6pt\selectfont Let $\beta'_{m} \!\in\! \mathbb{R}_+$ be an inertial sequence, $\beta'_m \!=\! (k_m \!-\! 1)/k_{m+1}$, where $k_m \!=\! 1 \!+\! (m^r \!-\! 1)/d\,$ with $r \!\in\! \mathbb{R}_{1,+}$, $d \!\in\! \mathbb{R}_{+}$.  Given an adjustable step size $\tau'_{i,t}{}^{\!\hspace{-0.125cm}m} \!\in\! \mathbb{R}_+$, update the causes using proximal-gradient steps until either convergence or a pre-set number of iterations has been reached\vspace{-0.2cm}$$\kappa_{i,t+1}^m \!=\! \textnormal{\sc prox}_{\!\lambda_i'}\!\Bigg(\!\pi'_{i,t}{}^{\!\!\!\!m} \!-\! \lambda_i'\tau'_{i,t}{}^{\!\hspace{-0.1cm}m}(2\eta_i'(\kappa_{i,t+1}^m \!-\! \kappa'_{i,t}) \!-\! \alpha_i' G_i^\top\textnormal{exp}(-G_i\pi'_{i,t}{}^{\!\!\!\!m})|\gamma_{i,t+1}^j|)\!\Bigg),\vspace{-0.2cm}$$
where $\pi'_{i,t}{}^{\!\hspace{-0.1475cm}m+1} \!=\! \kappa_{i,t+1}^m \!+\! \beta'_m (\kappa_{i,t+1}^m \!-\! \kappa_{i,t}^{m-1})$.  Update the sparsity parameter using spatial max unpooling after the causes update has concluded, $\lambda_{i,t+1} \!=\! \alpha_i'(1 \!+\! \textnormal{exp}(-\textnormal{\sc unpool}(G_i \kappa_{i,t+1})))$.\;  

\AlFnt{\small} }

\AlFnt{\small}    \For{$i \!=\! 0,1,2,\ldots$}{
\AlFnt{\small}       Update the filter dictionary matrix $D_i^\top \!\in\! \mathbb{R}^{k_i \times k_{i-1}}_+$ and the state-transition matrix $C_{i,t} \!\in\! \mathbb{R}^{k_i \times k_i}$ independently, until either convergence or a pre-set number of iterations has been reached, via dual-estimation filtering, with steps indexed by $m$,\vspace{-0.2cm} $$D_{i}^{m+1}{}^\top \!=\! D_{i}^m{}^\top \!+\! \sigma_t \!+\! \psi_i^m \Bigg(\!(\gamma_{i-1,t+1} \!-\! D_{i}^m{}^\top \gamma_{i,t+1})\gamma_{i,t+1} \!+\! \theta_i^m (D_{i}^m \!-\! D_{i}^{m-1})\!\Bigg),\vspace{-0.175cm}$$
$$C_{i}^{m+1} \!=\! C_{i}^m \!+\! \sigma_t' \!+\! \psi_{i}'{}^{m} \Bigg(\!\textnormal{\sc sign}(\gamma_{i,t+1} \!-\! C_{i}^m\gamma_{i,t})\gamma_{i,t+1}^\top \!+\! \theta_i^m (C_{i}^m \!-\! C_{i}^{m-1})\!\Bigg),\vspace{-0.2cm}$$
where $\psi_i^m,\psi_{i}'{}^{m} \!\in\! \mathbb{R}_+$ are step sizes, $\theta_i^m \!\in\! \mathbb{R}_+$ is a momentum coefficient, and $\sigma_t,\sigma_t' \!\in\! \mathbb{R}$ is Gaussian transition noise over the parameters.  Normalize $D_{i}^{m+1}{}^\top$ to avoid returning a trivial solution.\;
\AlFnt{\small}       Update the causal invariance matrix $G_i \!\in\! \mathbb{R}^{d_i \times k_i}$ via dual-estimation filtering, with steps indexed by $m$,\vspace{-0.2cm} $$G_{i}^{m+1} \!=\! G_{i}^m \!+\! \sigma_t'' \!+\! \psi_t''{}^{\vspace{-0.01cm}m} \Bigg(\!(\textnormal{exp}(-G_i^m\kappa_{i+1,t})|\gamma_{i,t+1}|)\kappa_{i,t+1}^\top \!+\! \theta_i^m (G_{i}^m \!-\! G_{i}^{m-1})\!\Bigg),\vspace{-0.2cm}$$ where $\psi_t''{}^{\vspace{-0.01cm}m} \!\in\! \mathbb{R}_+$ is a step size, $\theta_i^m \!\in\! \mathbb{R}_+$ is a momentum coefficient, and $\sigma_t'' \!\in\! \mathbb{R}$ is Gaussian-distributed transition noise over the parameters.  Normalize $G_{i}^{m+1}$ to avoid returning a trivial solution.
\AlFnt{\small} }

\AlFnt{\small} }
\end{algorithm}}\vspace{-0.3cm}
\end{figure*}

The convergence of dual-estimation filtering is straightforward to demonstrate.  For the proximal-gradient inference process, it is much more involved.  We build up to it in what follows.  We focus on the case where no inference restarts are performed.  This is done to theoretically show that the speed-ups we obtain are due solely to the improved convergence rate offered by the new extra-gradient update.  The inclusion of multiple restarts serves to delay the eventual inference slowdowns toward the end of inference.  Restarts should not improve the maximally achievable convergence rate.

We first prove a weak convergence result that facilitates demonstrating a much stronger one when relying on properties of Cauchy sequences.  We then quantify the global convergence rate for our chosen inertial sequence.  We then compare the global convergence rate to a Nesterov-style inertial sequence that was used in the original DPCN to illustrate the advantages of the former for ADPCNs.  Lastly, we outline local convergence properties of both inertial sequences to explain the results presented for DPCNs and ADPCNs in the main part of the paper.  A high-level overview of both networks is presented in \cref{fig:dpcn-comparison}.

\renewcommand{\thefigure}{A.\arabic{figure}}
\setcounter{figure}{0}

\begin{figure}[t!]
\hspace{2cm}\begin{minipage}[]{1\textwidth}
{\small \RaggedRight\parindent=1.5em
\fontdimen2\font=2.2pt\selectfont

\begin{tabular}{L{3cm}C{3cm}c@{\hspace{8pt}}C{3cm}}
\toprule
& DPCN & & ADPCN\\
\cmidrule[0.4pt](){1-4}%
Unsupervised & Yes & & Yes\\
\cmidrule[0.4pt](){1-4}%
Small models & Yes & & Yes\\
\cmidrule[0.4pt](){1-4}%
Convolutional & No \cite{ChalasaniR-conf2013a,PrincipeJC-jour2014a} / Yes \cite{ChalasaniR-jour2015a} & & Yes\\
\cmidrule[0.4pt](){1-4}%
Inference Restarts & No & & Yes\\
\cmidrule[0.4pt](){1-4}%
Feed-Forward Inference Conv. Rate & {\fontdimen2\font=1.7pt\selectfont $O(\zeta_i/m^2)$} & & {\fontdimen2\font=1.7pt\selectfont $O(r\zeta_i/(r\!-\!1)m^r)$}\\
\cmidrule[0.4pt](){1-4}%
Feed-Back Inference Converg. Rate & $O(1)$ & & $O(1)$\\
\cmidrule[0.4pt](){1-4}%
Feed-Forward Infer. Comput. Complexity & $O(k_i^2) \!+\! O(d_i^2)$ & & $O(k_i^2) \!+\! O(d_i^2)$\\
\cmidrule[0.4pt](){1-4}%
Parameter Learning Converg. Rate & $O(1/m)$ & & $O(1/m)$\\
\cmidrule[0.4pt](){1-4}%
Parameter Learning Comput. Complexity & $O(k_i^2) \!+\! O(d_i^2)$ & & $O(k_i^2) \!+\! O(d_i^2)$\\
\bottomrule
\end{tabular}}
\end{minipage}\vspace{-0.1cm}
   \caption[]{\fontdimen2\font=1.55pt\selectfont A quantitative comparison of DPCNs and ADPCNs.  The main distinction is that ADPCNs enjoy a much faster convergence rate for feed-forward inference.  This enables them to better describe stimuli and hence extract useful details from them.  Better feed-forward inference naturally impacts feed-back parameter updating.\vspace{-0.4cm}}
   \label{fig:dpcn-comparison}
\end{figure}

Toward this end, it is important to show that the distance between a given iterate and the solution set for either inference cost can be bounded by the norm of the residual.  This is a local Lipschitzian error bound.  The Lipschitzian bound is satisfied whenever the norm of the residual is small.  The iterate must also be sufficiently close to the solution set for this bound to be satisfied.

\begin{itemize}
\item[] \-\hspace{0.5cm}{\small{\sf{\textbf{Proposition A.1.}}}} Let $\gamma_{i,t} \!\in\! \mathbb{R}^{k_i}$ be the hidden states and $\kappa_{i,t} \!\in\! \mathbb{R}^{d_i}$ be the hidden causes.  Let $\pi_{i,t} \!\in\! \mathbb{R}^{k_i}$ be the auxiliary states and let $\pi'_{i,t} \!\in\! \mathbb{R}^{d_i}$ be the auxiliary causes at layer $i$ and time $t$.  Let $\omega_{1,i} \!\in\! \mathbb{R}$, that is assumed to\\ \noindent satisfy $\omega_{1,i} \geq \mathcal{L}_1(\gamma_{i}^*,\kappa_{i,t},C_i,D_i^\top;\alpha_i,\lambda_i)$.  As well, let $\omega_{2,i} \!\in\! \mathbb{R}$, $\omega_{2,i} \geq \mathcal{L}_2(\gamma_{i,t},\kappa_i^*,G_i;\alpha_i',\lambda_i',\eta_i',\lambda_{i,t})$.  For\\ \noindent some $\omega_{1,i}$, there are $\epsilon_{1,i},\epsilon_{1,i}' \!\in\! \mathbb{R}_+$, such that, for step size $\tau_{i,t}^m \!\in\! \mathbb{R}_+$
\begin{equation*}
\textnormal{dist}(\gamma_{i,t+1}^m,\Gamma_i^*) \leq \epsilon_{1,i}'\Bigg\|\textnormal{\sc prox}_{\!1/\ell_{i,j}}\!\Bigg(\!\pi_{i,t}^m \!-\! \lambda_{i,t}\tau_{i,t}^m\nabla_{\pi_{i,t}^m}\mathcal{L}_1(\pi_{i,t}^m,\kappa_{i,t},C_i,D_i^\top;\alpha_i,\lambda_i)\!\Bigg) - \pi_{i,t}^m\Bigg\|
\end{equation*}
whenever the following conditions, $\|\textnormal{\sc prox}_{\!\lambda_{i,t}}(\pi_{i,t}^m \!-\! \lambda_{i,t}\tau_{i,t}^m\nabla_{\pi_{i,t}}\mathcal{L}_1(\pi_{i,t}^m,\kappa_{i,t},C_i,D_i^\top;\alpha_i,\lambda_{i,t})) \!-\! \gamma_{i,t}^m\| < \epsilon_{1,i}$\\ \noindent and $\omega_{1,i} \leq \mathcal{L}_1(\gamma_{i,t}^m,\kappa_{i,t},C_i,D_i^\top;\alpha_i,\lambda_i)$, are satisfied.  Here, $\textnormal{dist}(\gamma_{i,t+1}^m,\Gamma_i^*)$ denotes the distance of a given iterate with the solution set.  Likewise, for some $\omega_{2,i}$, there are $\epsilon_{2,i},\epsilon_{2,i}' \!\in\! \mathbb{R}_+$, such that, for step sizes $\tau'_{i,t}{}^{\!\hspace{-0.125cm}m} \!\in\! \mathbb{R}_+$, 
\begin{equation*}
\textnormal{dist}(\kappa_{i,t}^m,K_i^*) \leq \epsilon_{2,i}'\Bigg\|\textnormal{\sc prox}_{\!1/\ell'_{i,j}}\!\Bigg(\!\pi'_{i,t}{}^{\!\hspace{-0.1475cm}m} \!-\! \lambda_i'\tau'_{i,t}{}^{\!\hspace{-0.125cm}m}\nabla_{\pi'_{i,t}{}^{\!\hspace{-0.15cm}m}}\mathcal{L}_2(\gamma_{i,t+1},\pi'_{i,t}{}^{\!\hspace{-0.1475cm}m},G_i;\alpha_i',\lambda_i',\eta_i',\lambda_{i,t})\!\Bigg) - \pi'_{i,t}{}^{\!\hspace{-0.1475cm}m}\Bigg\|
\end{equation*}
whenever the following conditions $\|\textnormal{\sc prox}_{\!\lambda_{i}}(\pi'_{i,t}{}^{\!\hspace{-0.1475cm}m} \!-\! \lambda_i'\tau'_{i,t}{}^{\!\hspace{-0.125cm}m}\nabla_{\pi'_{i,t}{}^{\!\hspace{-0.15cm}m}}\mathcal{L}_2(\gamma_{i,t+1},\pi'_{i,t}{}^{\!\hspace{-0.1475cm}m},G_i;\alpha_i',\lambda_i',\eta_i',\lambda_{i,t})) \!-\! \kappa_{i,t}^m\| < \epsilon_{2,i}$\\ \noindent and $\omega_{2,i} \leq \mathcal{L}_2(\gamma_{i,t+1},\pi'_{i,t}{}^{\!\hspace{-0.1475cm}m},G_i;\alpha_i',\lambda_i',\eta_i',\lambda_{i,t})$ are satisfied.  Here, $\Gamma_i^*$, with $\gamma_i^* \!\in\! \Gamma_i^*$, $\gamma_i^* \!\in\! \mathbb{R}^{k_i}$, denotes the sol-\\ \noindent ution set for the state-inference cost function.  As well, $K_i^*$, with $\kappa_i^* \!\in\! K_i^*$, $\kappa_i^* \!\in\! \mathbb{R}^{d_i}$, denotes the solution set for\\ \noindent the cause-inference cost function.  Both $\ell_{i,j},\ell_{i,j}' \!\in\! \mathbb{R}_+$ denote Lipschitz constants of the state and cause costs.
\end{itemize}
\begin{itemize}
\item[] \-\hspace{0.5cm}{\small{\sf{\textbf{Proof:}}}} In what follows, for ease of presentation, we ignore the variables of the inference cost that remain fixed across inference iterations.  Let the $L_1$-sparsity term in the state-inference cost be re-written in an equivalent manner as $\sum_{k'=1}^{k_i} [\lambda_{i,t}]_{k'}\xi_{i,t}$, with the constraint $|[\gamma_{i,t}]_{k'}| \!-\! \xi_{i,t} \!\leq\! 0$, $\xi_{i,t} \!\in\! \mathbb{R}^{k_i}$ \cite{TsengP-jour2009a}.  There exists some $\omega_i \!\in\! [0,\infty)^{k_i}$ such that
\begin{equation*}
(\textnormal{\sc prox}_{\!1/\ell_{i,j}}(\pi_{i,t}^m \!-\! \lambda_{i,t}\tau_{i,t}^m\nabla_{\pi_{i,t}^m}\mathcal{L}_1(\pi_{i,t}^m)) - \gamma_{i,t}^m) + (\nabla_{\pi_{i,t}^m}\mathcal{L}_1(\pi_{i,t}^m) - \omega_i) = 0,
\end{equation*}
where $(\textnormal{\sc prox}_{\!1/\ell_{i,j}}(\pi_{i,t}^m \!-\! \lambda_{i,t}\tau_{i,t}^m\nabla_{\pi_{i,t}^m}\mathcal{L}_1(\pi_{i,t}^m)) \!=\! \xi_{i,t}$.  Here, we assume that the choice of $\pi_{i,t}^m$ is such that the\\ \noindent inequality conditions are satisfied for $\epsilon_{1,i},\epsilon_{1,i}' \!\in\! \mathbb{R}_+$.  As well, there exists some optimal $\pi_i^* \!\in\! \Gamma_i^*$, $\pi_i^* \!\in\! \mathbb{R}_+^{k_i}$, and a corresponding $\omega^*_i \!\in\! [0,\infty)^{k_i}$ such that $\nabla_{\pi_{i}^*}\mathcal{L}_1(\pi_{i}^*) - \omega^*_i \!=\! 0$, where $\pi_i^* \!=\! \xi_{i,t}$.  We also have that, for $\sigma \!\in\! \mathbb{R}_+$,\\ \noindent $2\sigma\|\pi_{i,t}^m \!-\! \pi_i^*\|^2 \!\leq\! \langle \pi_{i,t}^m \!-\! \pi_i^*,\,\nabla_{\pi_{i,t}^m}\mathcal{L}_1(\pi_{i,t}^m) \!-\! \nabla_{\pi_{i}^*}\mathcal{L}_1(\pi_{i}^*)\rangle$.  Hence, for some $\omega_{1,i}' \!\in\! \mathbb{R}$ that depends on $\omega_{1,i} \!\in\! \mathbb{R}$,\\ \noindent we have
\begin{equation*}
2\sigma\|\pi_{i,t}^m \!-\! \pi_i^*\| \leq (\omega_{1,i}' \!+\! ((\omega_{1,i}')^2 \!+\! 4\omega_{1,i}')^{1/2})\|\pi_{i,t}^m \!-\! \textnormal{\sc prox}_{\!1/\ell_{i,j}}(\pi_{i,t}^m \!-\! \lambda_{i,t}\tau_{i,t}^m\nabla_{\pi_{i,t}^m}\mathcal{L}_1(\pi_{i,t}^m))\|/2.
\end{equation*}
When combined with $\|(\pi_{i,t}^m,\omega_{i,t}) \!-\! (\pi_i^*,\omega^*_i)\| \!\leq\! \delta_{i}(\|\pi_{i,t}^m \!-\! \pi_i^*\| \!+\! \|\pi_{i,t}^m \!-\! \textnormal{\sc prox}_{\!\lambda_{i,t}}(\pi_{i,t}^m \!-\! \lambda_{i,t}\tau_{i,t}^m\nabla_{\pi_{i,t}}\mathcal{L}_1(\pi_{i,t}^m))\|)$ for $\delta_i \!\in\! \mathbb{R}_+$, we get that $\textnormal{min}_{\pi^* \in \Gamma_i^*}\|\pi_{i,t}^m \!-\! \pi^*\| \!\leq\! \epsilon_{1,i}'\|\pi_{i,t}^m \!-\! \pi_i^*\|$.  

A similar argument to what is above can be used for the causes.$\;\;\footnotesize\selectfont\blacksquare$


\end{itemize}


\noindent As Luo and Tseng \cite{LuoZQ-jour1992a} have shown, such locally held bounds are useful for analyzing the rate of convergence of iterative algorithms for constrained, smooth optimization.  The bounds can be used to very weakly demonstrate that the sequence of functional iterates will eventually reach the global functional solution, albeit independent of strong inertial-sequence properties.  Analogous bounds have been derived by Pang \cite{PangJS-jour1987a} for strongly convex problems.  Here, we do not impose the condition of strong convexity for the inference costs, which makes these results applicable to a broader set of functionals, much like the ones that we employ.  

Tseng and Yun \cite{TsengP-jour2009a} later extended this bound for constrained, non-smooth optimization.  However, this was done in the context of gradient descent, not proximal gradients.  

To provide a more general convergence result, we need to consider specific inertial sequences and incorporate them into the analysis.  Toward this end, we first bound the squared residual between the primary iterates, the states $\gamma_{i,t}$ and the causes $\kappa_{i,t}$, and their corresponding auxiliary iterates $\pi_{i,t}$ and $\pi_{i,t}'$.  

\begin{itemize}
\item[] \-\hspace{0.5cm}{\small{\sf{\textbf{Proposition A.2.}}}} Let $\gamma_{i,t} \!\in\! \mathbb{R}^{k_i}$ be the hidden states and $\kappa_{i,t} \!\in\! \mathbb{R}^{d_i}$ be the hidden causes.  Let $\pi_{i,t} \!\in\! \mathbb{R}^{k_i}$ be the auxiliary states at layer $i$ and time $t$.  Assume that the state update, for a positive step size $\tau_{i,t}^m \!\in\! \mathbb{R}_+$, is given by\\ \noindent the relation $\gamma_{i,t+1}^m \!=\! \textnormal{\sc prox}_{\!\lambda_{i,t}}(\pi_{i,t}^m \!-\! \lambda_{i,t}\tau_{i,t}^m\nabla_{\pi_{i,t}^m}\mathcal{L}_1(\pi_{i,t}^m,\kappa_{i,t},C_i,D_i^\top;\alpha_i,\lambda_{i,t}))$, with the auxiliary state update\\ \noindent $\pi_{i,t}^{m+1} \!=\! \gamma_{i,t+1}^m \!+\! \beta_m (\gamma_{i,t+1}^m \!-\! \gamma_{i,t+1}^{m-1})$.  Likewise, assume that the cause update, for a positive step size $\tau'_{i,t}{}^{\!\hspace{-0.125cm}m} \!\in\! \mathbb{R}_+$, is\\ \noindent given by the relation $\kappa_{i,t+1}^m \!=\! \textnormal{\sc prox}_{\!\lambda_i}(\pi'_{i,t}{}^{\!\hspace{-0.1475cm}m} \!-\! \lambda_i'\tau'_{i,t}{}^{\!\hspace{-0.125cm}m}\nabla_{\pi'_{i,t}{}^{\!\hspace{-0.15cm}m}}\mathcal{L}_2(\gamma_{i,t+1},\pi'_{i,t}{}^{\!\hspace{-0.1475cm}m},G_i;\alpha_i',\lambda_i',\eta_i',\lambda_{i,t}))$, with the auxiliary\\ \noindent cause relation $\pi'_{i,t}{}^{\!\hspace{-0.1475cm}m+1} \!=\! \kappa_{i,t+1}^m \!+\! \beta'_m (\kappa_{i,t+1}^m \!-\! \kappa_{i,t+1}^m)$.  In both cases, let $\beta_m,\beta'_m \!=\! (k_m \!-\! 1)/k_{m+1}$, with elements\\ \noindent $k_m \!=\! 1 \!+\! (m^r \!-\! 1)/d$ with $r \!\in\! \mathbb{R}_{1,+}$, $d \!\in\! \mathbb{R}_{+}$.  There are some $\epsilon_{1,i},\epsilon_{2,i} \!\in\! \mathbb{R}_+$ such that
\begin{equation*}
\|\pi_{i,t}^m \!-\! \gamma_{i,t+1}^m\|^2 \geq \frac{\tau_{i,t}^m}{\epsilon_{1,i}}\Bigg(\!\mathcal{L}_1(\gamma_{i,t+1}^m,\kappa_{i,t},C_i,D_i^\top;\alpha_i,\lambda_{i,t}) \!-\! \mathcal{L}_1(\gamma_{i}^*,\kappa_{i,t},C_i,D_i^\top;\alpha_i,\lambda_{i,t})\!\Bigg)
\end{equation*}
\begin{equation*}
\|\pi'_{i,t}{}^{\!\hspace{-0.1475cm}m} \!-\! \kappa_{i,t+1}^m\|^2 \geq \frac{\tau'_{i,t}{}^{\!\hspace{-0.125cm}m}}{\epsilon_{2,i}}\Bigg(\!\mathcal{L}_2(\gamma_{i,t+1},\kappa_{i,t+1}^m,G_i;\alpha_i',\lambda_i',\eta_i',\lambda_{i,t}) \!-\! \mathcal{L}_2(\gamma_{i,t+1},\kappa_{i}^*,G_i;\alpha_i',\lambda_i',\eta_i',\lambda_{i,t})\!\Bigg)
\end{equation*}
where $\gamma_i^* \!\in\! \Gamma_i^*$, $\gamma_i^* \!\in\! \mathbb{R}_+^{k_i}$, is a solution of the state-inference cost and $\kappa_i^* \!\in\! K_i^*$, $\kappa_i^* \!\in\! \mathbb{R}_+^{d_i}$, is a solution of the\\ \noindent cause-inference cost.
\end{itemize}
\begin{itemize}
\item[] \-\hspace{0.5cm}{\small{\sf{\textbf{Proof:}}}} We focus on the case of the hidden states; that for the hidden causes has only slight differences.  In what follows, for ease of presentation, we ignore the variables of the inference cost that remain fixed across inference iterations.  We have, for $m$ being sufficiently large, that
\begin{align*}
\mathcal{L}_1(\gamma_{i,t+1}^m) \!-\! \mathcal{L}_1(\gamma_{i}^*) &\leq 2\|\gamma_{i,t+1}^{m+1} \!-\! \pi_{i,t}^{m+1}\|^2/\tau_{i,t}^m \!+\! \textnormal{dist}(\gamma_{i,t+1}^{m+1},\Gamma_i^*)\|\gamma_{i,t+1}^{m+1} \!-\! \pi_{i,t}^{m+1}\|/\tau_{i,t}^m\vspace{0.05cm}\\
&\leq \xi_{1,i}^{-1}(4\epsilon_{1,i}'' \!+\! \xi_{1,i})\|\gamma_{i,t+1}^{m+1} \!-\! \pi_{i,t}^{m+1}\|^2/2\tau_{i,t}^m
\end{align*}
where $\epsilon_{1,i}'' \!\in\! \mathbb{R}_+$ and $\xi_{1,i} \!\in\! \mathbb{R}_+$.  There exists some $\epsilon_{1,i} \!\geq\! \xi_{1,i}^{-1}(4\epsilon_{1,i}'' \!+\! \xi_{1,i})/2$, which must naturally be positive, such that the proposition is true.  Here, the second inequality follows from proposition A.1,
\begin{equation*}
\textnormal{dist}(\gamma_{i,t+1}^m,\Gamma_i^*) \leq \epsilon_{1,i}''\|\textnormal{\sc prox}_{\!\tau_{i,t}^m}(\gamma_{i,t+1}^m \!-\! \lambda_{i,t}\tau_{i,t}^m\nabla_{\gamma_{i,t+1}^m}\mathcal{L}_1(\gamma_{i,t+1}^m) \!-\! \gamma_{i,t+1}^m\|/\ell_{i,t}\tau_{i,t}^m,
\end{equation*}
This arises from the non-decreasing nature of the proximal-norm function and the Cauchy-Schwarz inequality, which implies that dividing by $\ell_{i,t}\tau_{i,t}^m$ leads to a non-increasing function.  The iterate-solution distance can thus be further bounded from above as $\textnormal{dist}(\gamma_{i,t+1}^m,\Gamma_i^*) \!\leq\! 2\epsilon_{1,i}''\xi_{1,i}\|\gamma_{i,t+1}^m \!-\! \pi_{i,t}^m\|$.  Since the relationship holds for arbitrary $m$ sufficiently large, it can be increased by one iteration.

A similar argument to what is above can be used for the causes.$\;\;\footnotesize\selectfont\blacksquare$
\end{itemize}


\noindent We now can bound the functional-value difference between an arbitrary iterates, $\gamma_{i,t}$ and $\kappa_{i,t}$, and optimal solutions, $\gamma_i^* \!\in\! \Gamma_i^*$ and $\kappa_i^* \!\in\! K_i^*$.  Note that, due to convexity of the two inference costs, every solution is guaranteed to be a\\ \noindent globally optimal one.  From this result, we will be able to obtain convergence of the function values and iterates.

\begin{itemize}
\item[] \-\hspace{0.5cm}{\small{\sf{\textbf{Proposition A.3.}}}} Let $\gamma_{i,t} \!\in\! \mathbb{R}^{k_i}$ be the hidden states and $\kappa_{i,t} \!\in\! \mathbb{R}^{d_i}$ be the hidden causes.  Let the inertial\\ \noindent sequences used, respectively, for the state-inference and cause-inference costs be $\beta_m,\beta'_m \!=\! (k_m \!-\! 1)/k_{m+1}$, where $k_m \!=\! 1 \!+\! (m^r \!-\! 1)/d$ with $r \!\in\! \mathbb{R}_{1,+}$, $d \!\in\! \mathbb{R}_{+}$.  We have that 
\begin{equation*}
\sum_{m=1}^\infty k_{m+1}^2\Bigg(\!\mathcal{L}_1(\gamma_{i,t+1}^m,\kappa_{i,t},C_i,D_i^\top;\alpha_i,\lambda_{i,t}) \!-\! \mathcal{L}_1(\gamma_{i}^*,\kappa_{i,t},C_i,D_i^\top;\alpha_i,\lambda_{i,t})\!\Bigg)
\end{equation*}
\begin{equation*}
\sum_{m=1}^\infty k_{m+1}^2\Bigg(\!\mathcal{L}_2(\gamma_{i,t+1},\kappa_{i,t+1}^m,G_i;\alpha_i',\lambda_i',\eta_i',\lambda_{i,t}) \!-\! \mathcal{L}_2(\gamma_{i,t+1},\kappa_{i}^*,G_i;\alpha_i',\lambda_i',\eta_i',\lambda_{i,t})\!\Bigg)
\end{equation*}
are convergent.  Here, $\gamma_i^* \!\in\! \Gamma_i^*$, $\gamma_i^* \!\in\! \mathbb{R}{k_i}$, is a solution of the state-inference cost and $\kappa_i^* \!\in\! K_i^*$, $\kappa_i^* \!\in\! \mathbb{R}^{d_i}$, is a\\ \noindent solution of the cause-inference cost.  We have assumed, here, that the state update, for a positive step size\\ \noindent $\tau_{i,t}^m \!\in\! \mathbb{R}_+$, was given by the relation $\gamma_{i,t+1}^m \!=\! \textnormal{\sc prox}_{\!\lambda_{i,t}}(\pi_{i,t}^m \!-\! \lambda_{i,t}\tau_{i,t}^m\nabla_{\pi_{i,t}^m}\mathcal{L}_1(\pi_{i,t}^m,\kappa_{i,t},C_i,D_i^\top;\alpha_i,\lambda_{i,t}))$, with the\vspace{-0.025cm}\\ \noindent auxiliary update $\pi_{i,t}^{m+1} \!=\! \gamma_{i,t+1}^m \!+\! \beta_m (\gamma_{i,t+1}^m \!-\! \gamma_{i,t+1}^{m-1})$.  As well, the cause update, for a positive step size $\tau_{i,t}' \!\in\! \mathbb{R}_+$,\\ \noindent was $\kappa_{i,t+1}^m \!=\! \textnormal{\sc prox}_{\!\lambda_i}(\pi'_{i,t}{}^{\!\hspace{-0.1475cm}m} \!-\! \lambda_i'\tau'_{i,t}{}^{\!\hspace{-0.125cm}m}\nabla_{\pi'_{i,t}{}^{\!\hspace{-0.15cm}m}}\mathcal{L}_2(\gamma_{i,t+1},\pi'_{i,t}{}^{\!\hspace{-0.1475cm}m},G_i;\alpha_i',\lambda_i',\eta_i',\lambda_{i,t}))$, with the auxiliary cause update\\ \noindent $\pi'_{i,t}{}^{\!\hspace{-0.1475cm}m+1} \!=\! \kappa_{i,t+1}^m \!+\! \beta'_m (\kappa_{i,t+1}^m \!-\! \kappa_{i,t+1}^m)$. 
\end{itemize}
\begin{itemize}
\item[] \-\hspace{0.5cm}{\small{\sf{\textbf{Proof:}}}} For ease of presentation, we ignore the variables of the inference cost that remain fixed across inference iterations.  It can be shown that, for some $\xi_{1,i} \!\in\! \mathbb{R}_+$,
\begin{align*}
\mathcal{L}_1(\gamma_{i,t+1}^{m+1}) &\leq \mathcal{L}_1(\gamma_{i,t+1}'{\!\!\!\!\!\!\hspace{-0.18cm}}^{m}\hspace{0.265cm}) \!-\! \|\gamma_{i,t+1}'{\!\!\!\!\!\!\hspace{-0.18cm}}^{m}\hspace{0.265cm} - \gamma_{i,t+1}^{m+1}\|^2/2\tau_{i,t}^m + \|\gamma_{i,t+1}'{\!\!\!\!\!\!\hspace{-0.18cm}}^{m}\hspace{0.265cm} \!- \pi_{i,t}^{m+1}\|^2/2\tau_{i,t}^m - (1 \!-\! \xi_{i,t})\|\gamma_{i,t+1}^{m+1} \!-\! \pi_{i,t}^{m+1}\|^2/\tau_{i,t}^m\vspace{0.05cm}\\
&\leq (1 \!-\! \beta_{m+1}^{-1})\mathcal{L}_1(\gamma_{i,t}^m) + \beta_{m+1}\mathcal{L}_1(\gamma_{i}^*) + \beta_{m+1}^{-2}\|\beta_{m}\gamma_{i,t+1}^{m+1} \!-\! (\beta_{m+1} \!-\! 1)\gamma_{i,t+1}^{m} \!-\! \gamma_i^*\|^2/2\tau_{i,t}^m\vspace{0.05cm}\\
&\;\;\;\;\;\;\;\;\;\;\;\; + \beta_{m+1}^{-2}\|\beta_m\gamma_{i,t+1}^m \!-\! (\beta_m \!-\! 1)\gamma_{i,t+1}^{m-1} \!-\! \gamma_i^*\|^2/2\tau_{i,t}^m - (1 \!-\! \xi_{i,t})\|\gamma_{i,t+1}^{m+1} \!-\! \pi_{i,t}^{m+1}\|^2/\tau_{i,t}^m
\end{align*}
where $\gamma_{i,t+1}'{\!\!\!\!\!\!\hspace{-0.18cm}}^{m}\hspace{0.265cm} \!=\! \beta_{m+1}^{-1}\gamma_i^* \!+\! (1\!-\!\beta_{m+!}^{-1})\gamma_{i,t}^m$.  Multiplying both sides by $k_{m+1}^2$ and re-arranging terms yields
\begin{align*}
& k_{m+1}^2(\mathcal{L}_1(\gamma_{i,t+1}^k) \!-\! \mathcal{L}_1(\gamma_i^*)) \!-\! k_{m+1}^2(\mathcal{L}_1(\gamma_{i,t+1}^{m+1}) \!-\! \mathcal{L}_1(\gamma_i^*))\vspace{0.05cm}\\
&\;\;\;\;\;\;\;\;\;\;\;\;\geq (k_m^2 \!-\! k_{m+1}^2 \!-\! k_{m+1})(\mathcal{L}_1(\gamma_{i,t+1}^m) \!-\! \mathcal{L}_1(\gamma_i^*)) + k_{m+1}^2(1 \!-\! \xi_{i,t})\|\gamma_{i,t+1}^m \!-\! \pi_{i,t}^m\|^2/2\tau_{i,t}^m \vspace{0.05cm}\\
&\;\;\;\;\;\;\;\;\;\;\;\;\;\;\;\;\;\;\;\;\;\;\;\; - \|k_{m+1}\gamma_{i,t+1}^{m+1} \!-\! (k_{m+1} \!-\! 1)\gamma_{i,t+1}^m \!-\! \gamma_i^*\|^2/\tau_{i,t}^m - \|k_{m}\gamma_{i,t+1}^{m} \!-\! (k_{m} \!-\! 1)\gamma_{i,t+1}^{m-1} \!-\! \gamma_i^*\|^2/\tau_{i,t}^m.
\end{align*}
The result derived in proposition A.2 can be applied to show that $k_{m+1}^2(1 \!-\! \xi_{i,t})\|\gamma_{i,t+1}^m \!-\! \pi_{i,t}^m\|^2/2\tau_{i,t}^m$ is bounded above by $k_{m+1}^2(1 \!-\! \xi_{i,t})(\mathcal{L}_1(\gamma_{i,t+1}^{m+1}) \!-\! \mathcal{L}_1(\gamma_i^*))/4\epsilon_{1,i}$.  Continuing from above, we have that
\begin{align*}
&(2k_m^2 \!-\! k_{m+1}^2 \!-\! k_{m+1})(\mathcal{L}_1(\gamma_{i,t+1}^m) \!-\! \mathcal{L}_1(\gamma_i^*)) - (k_m^2 \!-\! k_{m+1})(\gamma_{i,t+1}^{m+1}) \!-\! \mathcal{L}_1(\gamma_i^*))\vspace{0.05cm}\\
&\;\;\;\;\;\;\;\;\;\;\;\;\geq (k_m^2 \!-\! k_{m+1}^2 \!-\! k_{m+1})(\mathcal{L}_1(\gamma_{i,t+1}^m) \!-\! \mathcal{L}_1(\gamma_i^*)) + k_{m+1}^2(1 \!-\! \xi_{i,t})\|\gamma_{i,t+1}^m \!-\! \pi_{i,t}^m\|^2/2\tau_{i,t}^m \vspace{0.05cm}\\
&\;\;\;\;\;\;\;\;\;\;\;\;\;\;\;\;\;\;\;\;\;\;\;\; + k_{m+1}^2(1\!-\!\xi_{i,t})\|\gamma_{i,t+1}^{m+1} \!-\! \pi_{i,t}^{m+1}\|^2/4\tau_{i,t}^m.\hspace{6.7cm}\phantom{.}
\end{align*}
From this, we can see that $(2k_m^2 \!-\! k_{m+1}^2 \!-\! k_{m+1})(\mathcal{L}_1(\gamma_{i,t+1}^m) \!-\! \mathcal{L}_1(\gamma_i^*)) \!+\! k_{m+1}^2(1 \!-\! \xi_{i,t})\|\gamma_{i,t+1}^m \!-\! \pi_{i,t}^m\|^2/2\tau_{i,t}^m$ is\\ \noindent a non-increasing sequence for $m$.  It is bounded below.  This implies convergence of the sequence in $m$ and hence for $m \!+\! 1$.  This takes care of the two terms on the left and the first two terms on the right-hand side.  This leaves the final term on the right-hand side, $k_{m+1}^2(1 \!-\! \xi_{i,t})\|\gamma_{i,t+1}^{m+1} \!-\! \pi_{i,t}^{m+1}\|^2/4\tau_{i,t}^m$, which is also is convergent in $m$.  Applying proposition A.2 to this final term on the right-hand side proves the proposition for the hidden states.

A similar argument to what is above can be used for the causes.$\;\;\footnotesize\selectfont\blacksquare$
\end{itemize}

\noindent Based on properties of the inertial series $\{\beta_m\}_{m=1}^\infty$ and $\{\beta_m'\}_{m=1}^\infty$, particularly that the inverse of the inertial subcomponents $\sum_{m=1}^\infty k_m^{-1}$ is convergent, we immediately obtain that the state $\{\gamma_{i,t}^m\}_{m=1}^\infty$ and cause $\{\kappa_{i,t}^m\}_{m=1}^\infty$ iterates are Cauchy.  The state and cause iterates are thus bounded.  The Bolzano-Weierstrass theorem implies convergence of iterate subsequences for complete spaces, which applies to our case.  The iterates themselves are also strongly convergent to global solutions.  Convergence of proximal-gradient-type schemes is not new.  It, however, needed to be verified for our accelerated case.

We are now able to prove the main convergence result of the paper.

\begin{itemize}
\item[] \-\hspace{0.5cm}{\small{\sf{\textbf{Proposition 4.}}}} Let $\gamma_{i,t} \!\in\! \mathbb{R}^{k_i}$ be the hidden states and $\kappa_{i,t} \!\in\! \mathbb{R}^{d_i}$ be the hidden causes.  The state iterates\\ \noindent $\{\gamma_{i,t+1}^m\}_{m=1}^\infty$ strongly converge to the global solution of $\mathcal{L}_1(\gamma_{i,t},\kappa_{i,t},C_i,D_i^\top;\alpha_i,\lambda_{i,t})$ for the accelerated proximal gradient scheme.  Likewise, the cause iterates $\{\kappa_{i,t+1}^m\}_{m=1}^\infty$ for the accelerated proximal gradient scheme strongly converge to the global solution of $\mathcal{L}_2(\gamma_{i,t+1},\kappa_{i,t},G_i;\alpha_i',\lambda_i',\eta_i',\lambda_{i,t})$ at a sub-polynomial rate.  This occurs when using the inertial sequences $\beta_m,\beta'_m \!=\! (k_m \!-\! 1)/k_{m+1}$, where $k_m$ depends polynomially on $m$. 
\end{itemize}
\begin{itemize}
\item[] \-\hspace{0.5cm}{\small{\sf{\textbf{Proof:}}}} Strong convergence of the states $\{\gamma_{i,t+1}^m\}_{m=1}^\infty$ and causes $\{\kappa_{i,t+1}^m\}_{m=1}^\infty$ to the optimal solutions\\ \noindent $\gamma_i^* \!\in\! \Gamma_i^*$ and $\kappa_i^* \!\in\! K_i^*$ can be obtained from an extension of proposition A.3.  For the convergence rate, we note that there is some $\zeta_i \!\in\! \mathbb{R}_+$ such that $\zeta_i m^{-r} \!\geq\! \|\gamma_{i,t+1}^{m-1} \!-\! \gamma_{i,t}^m\|$.  For $m' \!>\! 1$, we have that $\|\gamma_{i,t+1}^{m+m'} \!-\! \gamma_{i,t}^m\|$ is bounded\vspace{-0.025cm}\\ \noindent above by $\sum_{j=m+1}^{m+m'} \|\gamma_{i,t}^j \!-\! \gamma_{i,t}^{j-1}\| \!\leq\! \zeta_i \sum_{j=m+1}^{m+m'} m^{-r}$.  As $m' \!\to\! \infty$, $\|\gamma_{i,t+1}^{m} \!-\! \gamma_i^*\| \!\leq\! \zeta_i r/m^r (r\!-\!1)$, which\\ \noindent implies a sub-$r$-polynomial rate of convergence for the state iterate sequence.  

A similar result holds for the cause iterates.$\;\;\footnotesize\selectfont\blacksquare$
\end{itemize}

The choice of the inertial sequence greatly affects convergence properties.  The classical sequence proposed by Nesterov, for instance, yields iterates $\{\gamma_{i,t+1}^m\}_{m=1}^\infty$ and $\{\kappa_{i,t+1}^m\}_{m=1}^\infty$ that only weakly converge to global solutions $\gamma_i^* \!\in\! \Gamma_i^*$ and $\kappa_i^* \!\in\! K_i^*$, which stems from the fact that $\sum_{m=1}^\infty k_m^{-1}$, with $k_{m+1} \!=\! (1 \!+\! (1 \!+\! 4k_m^2)^{1/2})/2$, is divergent.  In\\ \noindent finite-dimensional Euclidean spaces, this is not a shortcoming, since it implies convergence component-wise and thus is equivalent to strong convergence.

The original DPCN relied on a Nesterov-style sequence, so we analyze its convergence.  
\begin{itemize}
\item[] \-\hspace{0.5cm}{\small{\sf{\textbf{Proposition A.4.}}}} Let the inertial sequences used, respectively, for the state-inference and cause-inference costs be $\beta_m,\beta'_m \!=\! (k_m \!-\! 1)/k_{m+1}$, where $k_{m+1} \!=\! (1 \!+\! (1 \!+\! 4k_m^2)^{1/2})/2$.  We have that 
\begin{equation*}
\sum_{m=1}^\infty k_{m}^2\Bigg(\!\mathcal{L}_1(\gamma_{i,t+1}^m,\kappa_{i,t},C_i,D_i^\top;\alpha_i,\lambda_{i,t}) - \mathcal{L}_1(\gamma_{i}^*,\kappa_{i,t},C_i,D_i^\top;\alpha_i,\lambda_{i,t}) + \frac{1}{2\tau_{i,t}^m}\|\gamma_{i,t+1}^m \!-\! \gamma_{i,t+1}^{m-1}\|^2\!\Bigg)
\end{equation*}
\begin{equation*}
\sum_{m=1}^\infty k_{m}^2\Bigg(\!\mathcal{L}_2(\gamma_{i,t+1},\kappa_{i,t+1}^m,G_i;\alpha_i',\lambda_i',\eta_i',\lambda_{i,t}) \,-\, \mathcal{L}_2(\gamma_{i,t+1},\kappa_{i}^*,G_i;\alpha_i',\lambda_i',\eta_i',\lambda_{i,t}) \,+\, \frac{1}{2\tau'_{i,t}{}^{\!\hspace{-0.125cm}m}}\|\kappa_{i,t+1}^m - \kappa_{i,t+1}^{m-1}\|^2\!\Bigg)
\end{equation*}
are convergent.  Here, $\gamma_i^* \!\in\! \Gamma_i^*$, $\gamma_i^* \!\in\! \mathbb{R}^{k_i}$, is a solution of the state-inference cost and $\kappa_i^* \!\in\! K_i^*$, $\kappa_i^* \!\in\! \mathbb{R}^{d_i}$, is a\\ \noindent solution of the cause-inference cost.  We have assumed, here, that the state update, for a positive step size\\ \noindent $\tau_{i,t}^m \!\in\! \mathbb{R}_+$, was given by the relation $\gamma_{i,t+1}^m \!=\! \textnormal{\sc prox}_{\!\lambda_{i,t}}(\pi_{i,t}^m \!-\! \lambda_{i,t}\tau_{i,t}^m\nabla_{\pi_{i,t}^m}\mathcal{L}_1(\pi_{i,t}^m,\kappa_{i,t},C_i,D_i^\top;\alpha_i,\lambda_{i,t}))$, with the\\ \noindent auxiliary update $\pi_{i,t}^{m+1} \!=\! \gamma_{i,t+1}^m \!+\! \beta_m (\gamma_{i,t+1}^m \!-\! \gamma_{i,t+1}^{m-1})$.  As well, the cause update, for a positive step size $\tau_{i,t}' \!\in\! \mathbb{R}_+$,\\ \noindent was $\kappa_{i,t+1}^m \!=\! \textnormal{\sc prox}_{\!\lambda_i}(\pi'_{i,t}{}^{\!\hspace{-0.1475cm}m} \!-\! \lambda_i'\tau'_{i,t}{}^{\!\hspace{-0.125cm}m}\nabla_{\pi'_{i,t}{}^{\!\hspace{-0.15cm}m}}\mathcal{L}_2(\gamma_{i,t+1},\pi'_{i,t}{}^{\!\hspace{-0.1475cm}m},G_i;\alpha_i',\lambda_i',\eta_i',\lambda_{i,t}))$, with the auxiliary cause update\\ \noindent $\pi'_{i,t}{}^{\!\hspace{-0.1475cm}m+1} \!=\! \kappa_{i,t+1}^m \!+\! \beta'_m (\kappa_{i,t+1}^m \!-\! \kappa_{i,t+1}^m)$. 
\end{itemize}
\begin{itemize}
\item[] \-\hspace{0.5cm}{\small{\sf{\textbf{Proof:}}}} For ease of presentation, we ignore the variables of the inference cost that remain fixed across inference iterations.  It can be shown that
\begin{equation*}
\begin{array}{l}
\!\!\!\!\!\!\!\!\!\mathcal{L}_1(\gamma_{i,t+1}^{m}) - \mathcal{L}_1(\gamma_{i}^*) + \beta_m^2\|\gamma_{i,t+1}^m \!-\! \gamma_{i,t+1}^{m-1}\|^2/2\tau_{i,t}^m\vspace{0.05cm}\\
\;\;\;\;\;\;\;\;\;\;\;\;\;\;\;\;\;\;\;\;\;\;\;\;\;\;\;\;\;\;\;\;\;\;\;\;\;\;\;\;\;\;\;\;\;\;\;\;\;\;\;\;\;\;\;\;\;\;\;\;\;\;\;\geq \mathcal{L}_1(\gamma_{i,t+1}^{m+1}) - \mathcal{L}_1(\gamma_{i}^*) + \|\gamma_{i,t+1}^m \!-\! \gamma_{i,t+1}^{m+1}\|^2/2\tau_{i,t}^{m+1}.
\end{array}
\end{equation*}
Multiplying both sides by $k_{m+1}^2$, performing an addition by zero, and re-arranging terms yields
\begin{align*}
& k_{m+1}^3(\mathcal{L}_1(\gamma_{i,t+1}^{m}) \!-\! \mathcal{L}_1(\gamma_{i}^*)) + k_{m+1}^3\|\gamma_{i,t+1}^m \!-\! \gamma_{i,t+1}^{m-1}\|^2/2\tau_{i,t}^m\vspace{0.05cm}\\
&\;\;\;\;\;\;\;\;\;\;\;\;\leq (k_{m+1}^3 \!+\! k_m^3 \!-\! k_m^3)(\mathcal{L}_1(\gamma_{i,t+1}^{m}) \!-\! \mathcal{L}_1(\gamma_{i}^*)) - k_{m+1}(k_m \!-\! 1)^2\|\gamma_{i,t+1}^m \!-\! \gamma_{i,t+1}^{m-1}\|^2/2\tau_{i,t}^m\vspace{0.05cm}\\
&\;\;\;\;\;\;\;\;\;\;\;\;\leq k_m^3(\mathcal{L}_1(\gamma_{i,t+1}^{m}) \!-\! \mathcal{L}_1(\gamma_{i}^*)) + k'(\mathcal{L}_1(\gamma_{i,t+1}^{m}) \!-\! \mathcal{L}_1(\gamma_{i}^*)) - (k_m^2(2 \!-\! k') \!+\! k_m(2k' \!-\! 1) \!-\! k')/2\tau_{i,j}^m\vspace{0.05cm}\\
&\;\;\;\;\;\;\;\;\;\;\;\;\;\;\;\;\;\;\;\;\;\;\;\; + k_m^3\|\gamma_{i,t+1}^m \!-\! \gamma_{i,t+1}^{m-1}\|^2/2\tau_{i,t}^m.
\end{align*}
The last inequality follows because $\sum_{m=1}^\infty k_m^{-1}$ is divergent.  In this case, there exists some $0 \!<\! k' \!<\! 2$ such that $k_{m+1} \!-\! k_m \!\leq\! k'$ for all $m \!>\! m'$, $m' \!>\! 0$.  We therefore have that
\begin{align*}
& k'(k_m^2 \!+\! k_m \!+\! 1)(\mathcal{L}_1(\gamma_{i,t+1}^{m}) \!-\! \mathcal{L}_1(\gamma_{i}^*))\vspace{0.05cm}\\
&\;\;\;\;\;\;\;\;\;\;\;\;\geq k_{m+1}^3(\mathcal{L}_1(\gamma_{i,t+1}^{m+1}) \!-\! \mathcal{L}_1(\gamma_{i}^*) + \|\gamma_{i,t+1}^{m+1} \!-\! \gamma_{i,t+1}^{m}\|^2/2\tau_{i,t}^{m+1}) + k_m^3(\mathcal{L}_1(\gamma_{i,t+1}^{m}) \!-\! \mathcal{L}_1(\gamma_{i}^*)\vspace{0.05cm}\\
&\;\;\;\;\;\;\;\;\;\;\;\;\;\;\;\;\;\;\;\;\;\;\;\;  + \|\gamma_{i,t+1}^{m} \!-\! \gamma_{i,t+1}^{m-1}\|^2/2\tau_{i,t}^{m}).
\end{align*}
Re-organizing terms allows us to demonstrate that $\sum_{m=1}^\infty \|\gamma_{i,t+1}^{m} \!-\! \gamma_{i,t+1}^{m-1}\|^2/2\tau_{i,t}^{m}$ is convergent via a Cauchy test.  This implies that $\sum_{m=1}^\infty k_m^2(\mathcal{L}_1(\gamma_{i,t+1}^{m}) \!-\! \mathcal{L}_1(\gamma_i^*) \!+\! \|\gamma_{i,t+1}^m \!-\! \gamma_{i,t+1}^{m-1}|^2/2\tau_{i,t}^m)$ is also convergent.  

A similar argument to what is above can be used for the causes.$\;\;\footnotesize\selectfont\blacksquare$

\end{itemize}

\noindent The rate of convergence, though, is limited when choosing a Nesterov-style inertial sequence, both locally and globally.  In the global case, we have that
\begin{itemize}
\item[] \-\hspace{0.5cm}{\small{\sf{\textbf{Proposition A.5.}}}} Let $\gamma_{i,t} \!\in\! \mathbb{R}^{k_i}$ be the hidden states and $\kappa_{i,t} \!\in\! \mathbb{R}^{d_i}$ be the hidden causes.  The state iterates\\ \noindent $\{\gamma_{i,t+1}^m\}_{m=1}^\infty$ strongly converge to the global solution of $\mathcal{L}_1(\gamma_{i,t},\kappa_{i,t},C_i,D_i^\top;\alpha_i,\lambda_{i,t})$ for the accelerated proximal gradient scheme.  Likewise, the cause iterates $\{\kappa_{i,t+1}^m\}_{m=1}^\infty$ for the accelerated proximal gradient scheme strongly converge to the global solution of $\mathcal{L}_2(\gamma_{i,t+1},\kappa_{i,t},G_i;\alpha_i',\lambda_i',\eta_i',\lambda_{i,t})$ at a sub-quadratic rate.  This occurs when using the inertial sequences $\beta_m,\beta'_m \!=\! (k_m \!-\! 1)/k_{m+1}$, where $k_{m+1} \!=\! (1 \!+\! (1 \!+\! 4k_m^2)^{1/2})/2$.
\end{itemize}
\begin{itemize}
\item[] \-\hspace{0.5cm}{\small{\sf{\textbf{Proof:}}}} Strong convergence of the states $\{\gamma_{i,t+1}^m\}_{m=1}^\infty$ and causes $\{\kappa_{i,t+1}^m\}_{m=1}^\infty$ to the optimal solutions\\ \noindent $\gamma_i^* \!\in\! \Gamma_i^*$ and $\kappa_i^* \!\in\! K_i^*$ can be obtained from an extension of proposition A.4.  For the convergence rate, we note that there is some $\zeta_i \!\in\! \mathbb{R}_+$ such that $\zeta_i m^{-3/2} \!\geq\! \|\gamma_{i,t+1}^{m-1} \!-\! \gamma_{i,t}^m\|$.  For $m' \!>\! 1$, we have that $\|\gamma_{i,t+1}^{m+m'} \!-\! \gamma_{i,t}^m\|$ is bounded\vspace{-0.05cm}\\ \noindent above by $\sum_{j=m+1}^{m+m'} \|\gamma_{i,t}^j \!-\! \gamma_{i,t}^{j-1}\| \!\leq\! \zeta_i \sum_{j=m+1}^{m+m'} m^{-3}$.  As $m' \!\to\! \infty$, $\|\gamma_{i,t+1}^{m} \!-\! \gamma_i^*\| \!\leq\! \zeta_i/2m^{2}$, which implies a\\ \noindent sub-quadratic rate of convergence for the state iterate sequences.

A similar argument to what is above demonstrates the convergence rate for the cause iterate sequences.$\;\;\footnotesize\selectfont\blacksquare$
\end{itemize}

From this proposition, we see that a polynomial extra-gradient step permits making either equal or larger adjustments, on average, per iteration than the Nesterov step.  It is, in essence, artificially adapting the step size without causing divergence.  The ADPCN iterates thus often reach a given loss magnitude more quickly than those for the DPCN.  

This potentially improved step size offered by a polynomial inertial sequence has a major impact on ADPCN performance.  A feed-forward, bottom-up inference problem is solved to propagate a batch of stimuli through the network and convert it into states and causes.  This problem has to be solved at every stage, which we do using the method of proximal gradients.  If a subpar solution is returned for the bottom-up inference, then it impacts the quality of the parameter updates.  This, in turn, affects the network response for the next batch of stimuli.  A great many presentations of the stimuli may be needed to counteract this issue.  There is a chance that the corresponding network parameters learned using a linear extra-gradient step may never approach the same performance as those for our polynomial-based step.

Note that a feed-back, top-down inference problem must also be solved at every stage to back-propagate the causes to earlier layers.  However, top-down inference has a global best solution that can be obtained directly without relying on non-iterative processes.  This is why we listed it as having a constant-time convergence rate in \cref{fig:dpcn-comparison}.

To better understand why the convergence is better with a polynomial inertial sequence, it is helpful to re-cast the proximal gradient updates in a way that permits understanding local convergence behaviors using spectral analysis.  We do this first for the state-inference process.

\begin{itemize}
\item[] \-\hspace{0.5cm}{\small{\sf{\textbf{Proposition A.6.}}}} The state update $\gamma_{i,t+1}^{m} \!=\! \textnormal{\sc prox}_{\!\lambda_{i,t}}(\pi_{i,t}^m \!-\! \lambda_{i,t}\tau_{i,t}^m\nabla_{\pi_{i,t}}\mathcal{L}_1(\pi_{i,t},\kappa_{i,t},C_i,D_i^\top;\alpha_i,\lambda_{i,t}))$,\\ \noindent with $\pi_{i,t}^{m+1} \!=\! \gamma_{i,t}^{m} \!+\! \beta_m (\gamma_{i,t}^m \!-\! \gamma_{i,t}^{m-1})$, is equivalent to $\gamma_{i,t+1}^{m} \!=\! \textnormal{\sc shrink}(w_{i,t}^m;\lambda_{i,t}/\ell_{i,t})$, for the auxiliary variable
\begin{equation*}
w_{i,t}^m \!=\! \Bigg(\!I_{k_i \times k_i} \!-\! \ell_{i,t}^{-1}D_i^\top D_i\!\Bigg)\pi_{i,t}^m \!+\! \ell_{i,t}^{-1}D_i^\top \kappa_{i-1,t} \!+\! \alpha_i \ell_{i,t}^{-1}\textnormal{\sc proj}_\infty\Bigg(\!(\pi_{i,t}^m \!-\! C_i\gamma_{i,t-1})/\mu_i\!\Bigg),
\end{equation*}
where $I_{k_i \times k_i} \!\in\! \mathbb{R}_+^{k_i \times k_i}$ is the identity matrix, $0_{1 \times 2k_i} \!\in\! \mathbb{R}^{2k_i}$ is a row vector of zeros and $1_{k_i+1 \times 1} \!\in\! \mathbb{R}^{k_i+1}_+$ is a\\ \noindent column vector of ones.  This is equivalent to the matrix recurrence $(w^{m+1}_{i,t},w^{m}_{i,t},1)^\top \!\!=\! S_{i,t}^m(w^{m}_{i,t},w^{m-1}_{i,t},1)^\top$.\\ \noindent  More specifically,
\begin{equation*}
\begin{pmatrix}
w^{m+1}_{i,t}\vspace{0.05cm}\\
w^{m}_{i,t}\vspace{0.05cm}\\
1
\end{pmatrix} = 
\underbrace{\begin{pmatrix}
W_{i,t}^{m} & \ell_{i,t}^{-1} D_i^\top \kappa_{i-1,t} \!+\! (I_{k_i \times k_i} \!-\! \ell_{i,t}^{-1} D_i^\top D_i) z_{i,t}^{m} \!+\! \alpha_i\ell_{i,t}^{-1}f_{i,t}^m\vspace{0.05cm}\\
0_{1 \times 2k_i} & 1_{k_i+1 \times 1}\\
\end{pmatrix}}_{S_{i,t}^m}\! \begin{pmatrix}
w^{m}_{i,t}\vspace{0.05cm}\\
w^{m-1}_{i,t}\vspace{0.05cm}\\
1
\end{pmatrix}\vspace{-0.1cm}
\end{equation*}
where $z_{i,t}^{m} \!=\! -(1 \!+\! \beta_m)\lambda_{i,t} s_{i,t}^{m}/\ell_{i,t} \!+\! \beta_m\lambda_{i,t}s_{i,t}^{m-1}/\ell_{i,t} $, with $s_{i,t}^{m} \!=\! \textnormal{\sc sign}(\textnormal{\sc shrink}(w^{m}_{i,t};\lambda_{i,t}/\ell_{i,t}))$.  The term $f_{i,t}^m$ accounts for the Nesterov-smoothed component, $f_{i,t}^m \!=\! \textnormal{\sc proj}_\infty(((H_{i,t}^m)^2 w_{i,t}^m \!-\! \lambda_{i,t}^m\ell_{i,t}^{-1} s_{i,t}^m \!-\! C_i\gamma_{i,t-1})/\mu_i)$, which is given by projecting the $L_1$-sparse state-transition component onto an $L_\infty$ ball.  The matrix $W_{i,t}^{m} \!\in\! \mathbb{R}^{2k_i \times 2k_i}$ is
\begin{equation*}
W_{i,t}^{m} = \begin{pmatrix}
(1 \!+\! \beta_m)(I_{k_i \times k_i} \!-\! \ell_{i,t}^{-1} D_i^\top D_i)(H_{i,t}^{m})^2 & -\beta_{m+1}(I_{k_i \times k_i} \!-\! \ell_{i,t}^{-1} D_i^\top D_i)(H_{i,t}^{m})^2\vspace{0.05cm}\\
I_{k_i \times k_i} & 0_{k_i \times k_i}
\end{pmatrix}.
\end{equation*}
Here, $\ell_{i,t} \!\in\! \mathbb{R}_{0,+}$ is the Lipschitz constant of the state-inference cost at a given layer $i$ and for the current batch $t$.  The flag matrix $H_{i,t}^{m} \!=\! \textnormal{\sc diag}(\textnormal{\sc sign}(\textnormal{\sc shrink}(w^{m}_{i,t};\lambda_{i,t}/\ell_{i,t})))$ is diagonal and relies on a sparse shrinkage process\\ \noindent for the auxiliary variable.
\end{itemize}
\begin{itemize}
\item[] \-\hspace{0.5cm}{\small{\sf{\textbf{Proof:}}}} The underlying update for accelerated proximal gradients can be re-written as
\begin{align*}
\gamma_{i,t+1}^{m} &= \textnormal{arg min}_\pi (\|\pi \!-\! (\pi_{i,t}^m \!-\! \ell_{i,t}^{-1}\nabla_{\pi_{i,t}^m}\mathcal{L}_1'(\pi_{i,t}^m))\|^2/2\ell_{i,t} \!+\! \lambda_{i,t}\|\pi\|_1)\vspace{0.05cm}\\
&= \textnormal{\sc shrink}((I_{k_i \times k_i} \!-\! \ell_{i,t}^{-1}D_i^\top D_i)\pi_{i,t}^m \!+\! \ell_{i,t}^{-1}D_i^\top \kappa_{i-1,t} \!+\! \alpha_i \ell_{i,t}^{-1}\textnormal{\sc proj}_\infty((\pi_{i,t}^m \!-\! C_i\gamma_{i,t-1})/\mu_i);\lambda_{i,t}/\ell_{i,t})
\end{align*}
where $\mathcal{L}_1'(\pi_{i,t}^m)$ represents the state-inference cost but without the $L_1$-sparsity constraint on the states.  Here, we have used a Nesterov smoothing approach, with $\mu_i \!\in\! \mathbb{R}_+$, to deal with the $L_1$-sparse state-transition update,\\ \noindent $\textnormal{arg max}_{\|\Omega_{i,t}\|_\infty \leq 1} \Omega_{i,t}^\top (\pi_{i,t}^m \!-\! C_i \gamma_{i,t-1}) \!-\! \mu_i\|\Omega_{i,t}\|^2_2/2 \!=\! \textnormal{\sc proj}_\infty((\pi_{i,t}^m \!-\! C_i\gamma_{i,t-1})/\mu_i)$.  This projection onto an $L_\infty$-ball has the closed-form solution
\begin{equation*}
\textnormal{\sc proj}_\infty((\pi_{i,t}^m \!-\! C_i\gamma_{i,t-1})/\mu_i = \begin{cases}1, & (\pi_{i,t}^m \!-\! C_i\gamma_{i,t-1})/\mu_i \!>\! 1\vspace{0.05cm}\\
(\pi_{i,t}^m \!-\! C_i\gamma_{i,t-1})/\mu_i, & -1 \!\leq\! (\pi_{i,t}^m \!-\! C_i\gamma_{i,t-1})/\mu_i \!\leq\! 1\vspace{0.05cm}\\
-1, & (\pi_{i,t}^m \!-\! C_i\gamma_{i,t-1})/\mu_i \!<\! -1
\end{cases}.
\end{equation*}
We replace the states by the auxiliary variable $w_{i,t}^m$ and note that $\gamma_{i,t+1}^{m} = \textnormal{\sc shrink}(w_{i,t}^m;\lambda_{i,t}/\ell_{i,t})$, where
\begin{equation*}
\textnormal{\sc shrink}(w_{i,t}^m;\lambda_{i,t}/\ell_{i,t}) = \textnormal{\sc diag}(\textnormal{\sc sign}(\textnormal{\sc shrink}(w_{i,t}^m;\lambda_{i,t}/\ell_{i,t})))^2w_{i,t}^m \!-\! \lambda_{i,t}\textnormal{\sc sign}(\textnormal{\sc shrink}(w_{i,t}^m;\lambda_{i,t}/\ell_{i,t}))/\ell_{i,t},
\end{equation*}
\begin{equation*}
\textnormal{\sc sign}(\textnormal{\sc shrink}(w_{i,t}^m;\lambda_{i,t}/\ell_{i,t})) = \begin{cases}
1, & w_{i,t}^m \!>\! \lambda_{i,t}/\ell_{i,t}\vspace{0.05cm}\\
0, & -\lambda_{i,t}/\ell_{i,t} \!\leq\! w_{i,t}^m \!\leq\! \lambda_{i,t}/\ell_{i,t}\vspace{0.05cm}\\
-1, &  w_{i,t}^m \!<\! -\lambda_{i,t}/\ell_{i,t}
\end{cases}.
\end{equation*}
We can systematically re-write the auxiliary-variable update as
\begin{align*}
w_{i,t}^{m+1} &= (I_{k_i \times k_i} \!-\! \ell_{i,t}^{-1}D_i^\top D_i)(\gamma_{i,t+1}^m \!+\! \beta_m(\gamma_{i,t+1}^m \!-\! \gamma_{i,t+1}^{m-1})) + \ell_{i,t}^{-1}D_i^\top \kappa_{i-1,t} + \alpha_i \ell_{i,t}^{-1}\textnormal{\sc proj}_\infty((\gamma_{i,t+1}^m \!-\! C_i\gamma_{i,t})/\mu_i)\vspace{0.05cm}\\
&= (I_{k_i \times k_i} \!-\! \ell_{i,t}^{-1}D_i^\top D_i)((1 \!+\! \beta_{m})(H_{i,t}^{m-1})^2w_{i,t}^m \!-\! \lambda_{i,t}\textnormal{\sc sign}(\textnormal{\sc shrink}(w_{i,t}^{m-1};\lambda_{i,t}/\ell_{i,t}))/\ell_{i,t})\vspace{0.05cm}\\
&\;\;\;\;\;\;\;\;\;\;\;\; - (I_{k_i \times k_i} \!-\! \ell_{i,t}^{-1}D_i^\top D_i)(\beta_{m-1}(H_{i,t}^m)^2w_{i,t}^m \!+\! \lambda_{i,t}\textnormal{\sc sign}(\textnormal{\sc shrink}(w_{i,t}^{m-1};\lambda_{i,t}/\ell_{i,t}))/\ell_{i,t})\vspace{0.05cm}\\
&\;\;\;\;\;\;\;\;\;\;\;\; - \ell_{i,t}^{-1}D_i^\top \kappa_{i-1,t} + \alpha_i \ell_{i,t}^{-1}\textnormal{\sc proj}_\infty((\gamma_{i,t+1}^m \!-\! C_i\gamma_{i,t})/\mu_i)\vspace{0.05cm}\\
&= (1 \!+\! \beta_m)(I_{k_i \times k_i} \!-\! \ell_{i,t}^{-1}D_i^\top D_i)(H_{i,t}^m)^2w_{i,t}^m -\beta_{m}(I_{k_i \times k_i} \!-\! \ell_{i,t}^{-1}D_i^\top D_i)(H_{i,t}^{m-1})^2w_{i,t}^{m-1} - \ell_{i,t}^{-1}D_i^\top \kappa_{i-1,t}\vspace{0.05cm}\\
&\;\;\;\;\;\;\;\;\;\;\;\; + (I_{k_i \times k_i} \!-\! \ell_{i,t}^{-1}D_i^\top D_i)((-\lambda_{i,t} \!-\! \lambda_{i,t}\beta_m)\textnormal{\sc sign}(\textnormal{\sc shrink}(w_{i,t}^{m-1};\lambda_{i,t}/\ell_{i,t}))/\ell_{i,t}\vspace{0.05cm}\\
&\;\;\;\;\;\;\;\;\;\;\;\; + \lambda_{i,t}\beta_m)\textnormal{\sc sign}(\textnormal{\sc shrink}(w_{i,t}^{m-1};\lambda_{i,t}/\ell_{i,t}))/\ell_{i,t}) + \alpha_i \ell_{i,t}^{-1}\textnormal{\sc proj}_\infty((\gamma_{i,t+1}^m \!-\! C_i\gamma_{i,t})/\mu_i).
\end{align*}
The matrix recurrence follows from this update.$\;\;\footnotesize\selectfont\blacksquare$
\end{itemize}

\noindent We now characterize the cause inference in a similar manner.

\begin{itemize}
\item[] \-\hspace{0.5cm}{\small{\sf{\textbf{Proposition A.7.}}}} The cause update $\kappa_{i,t+1}^m \!=\! \textnormal{\sc prox}_{\!\lambda_i}(\pi'_{i,t}{}^{\!\hspace{-0.1475cm}m} \!-\! \lambda_i'\tau'_{i,t}{}^{\!\hspace{-0.125cm}m}\nabla_{\pi'_{i,t}{}^{\!\hspace{-0.15cm}m}}\mathcal{L}_2(\gamma_{i,t+1},\pi'_{i,t}{}^{\!\hspace{-0.1475cm}m},G_i;\alpha_i',\lambda_i',\eta_i',\lambda_{i,t}))$,\\ \noindent with $\pi'_{i,t}{}^{\!\hspace{-0.1475cm}m+1} \!=\! \kappa_{i,t+1}^m \!+\! \beta'_m (\kappa_{i,t+1}^m \!-\! \kappa_{i,t+1}^m)$, is equivalent to $\kappa_{i,t+1}^{m} \!=\! \textnormal{\sc shrink}(v_{i,t}^m;\lambda_{i}'/\ell'_{i,t})$, for the auxiliary\\ \noindent variable
\begin{equation*}
v_{i,t}^m \!=\! \Bigg(\!I_{d_i \times d_i} \!-\! 1/\ell'_{i,t}\!\Bigg)\pi_{i,t}'{}^{\!\!\!\!m} \!+\! 2\eta_i'I_{d_i \times d_i} \kappa_{i,t}'/\ell_{i,t}' \!-\! \alpha_i'/\ell_{i,t}'\Bigg(\!G_i^\top \textnormal{exp}(-G_i \pi_{i,t}'{}^{\!\!\!\!m})|\gamma_{i,t+1}^j| \!\Bigg),
\end{equation*}
where $I_{d_i \times d_i} \!\in\! \mathbb{R}_+^{d_i \times d_i}$ is the identity matrix, $0_{1 \times 2d_i} \!\in\! \mathbb{R}^{2d_i}$ is a row vector of zeros and $1_{d_i+1 \times 1} \!\in\! \mathbb{R}^{d_i+1}_+$ is a\\ \noindent column vector of ones.  This is equivalent to the matrix recurrence $(v^{m+1}_{i,t},v^{m}_{i,t},1)^\top \!\!=\! T_{i,t}^m(v^{m}_{i,t},v^{m-1}_{i,t},1)^\top$.  More\\ \noindent  specifically,
\begin{equation*}
\begin{pmatrix}
v^{m+1}_{i,t}\vspace{0.05cm}\\
v^{m}_{i,t}\vspace{0.05cm}\\
1
\end{pmatrix} = 
\underbrace{\begin{pmatrix}
V_{i,t}^{m} & 2\eta_i'I_{d_i \times d_i} \kappa_{i,t}'/\ell_{i,t}' \!+\! (I_{d_i \times d_i} \!-\! 1/\ell_{i,t}') z_{i,t}'{}^{\!\!\!\!m} \!-\! \alpha_i'g_{i,t}^m/\ell_{i,t}'\vspace{0.05cm}\\
0_{1 \times 2d_i} & 1_{d_i+1 \times 1}\\
\end{pmatrix}}_{T_{i,t}^m}\! \begin{pmatrix}
v^{m}_{i,t}\vspace{0.05cm}\\
v^{m-1}_{i,t}\vspace{0.05cm}\\
1
\end{pmatrix}\vspace{-0.1cm}
\end{equation*}
where $z_{i,t}^{m} \!=\! -(1 \!+\! \beta_m')\lambda_i' q_{i,t}^m/\ell_{i,t}' \!+\! \beta_m'\lambda_i'q_{i,t}^{m-1}/\ell_{i,t}' $, with $q_{i,t}^m \!=\! \textnormal{\sc sign}(\textnormal{\sc shrink}(v^{m}_{i,t};\lambda_i'/\ell_{i,t}'))$.  The term $g_{i,t}^m$\\ \noindent accounts for the invariant-matrix component, $g_{i,t}^m \!=\! G_i^\top \textnormal{exp}(-G_i v_{i,t}'{}^{\!\!\!\!m})|\gamma_{i,t+1}^j|$.  The matrix $V_{i,t}^{m} \!\in\! \mathbb{R}^{2d_i \times 2d_i}$ is
\begin{equation*}
V_{i,t}^{m} = \begin{pmatrix}
(1 \!+\! \beta_m')(I_{d_i \times d_i} \!-\! 1/\ell_{i,t}')(M_{i,t}^{m})^2 & -\beta_{m+1}'(I_{d_i \times d_i} \!-\! 1/\ell_{i,t}')(M_{i,t}^{m})^2\vspace{0.05cm}\\
I_{d_i \times d_i} & 0_{d_i \times d_i}
\end{pmatrix}.
\end{equation*}
Here, $\ell_{i,t}' \!\in\! \mathbb{R}_{0,+}$ is the Lipschitz constant of the state-inference cost at a given layer $i$ and for the current batch $t$.  The flag matrix $M_{i,t}^{m} \!=\! \textnormal{\sc diag}(\textnormal{\sc sign}(\textnormal{\sc shrink}(v^{m}_{i,t};\lambda_i'/\ell_{i,t}')))$ is diagonal and relies on a sparse shrinkage process\\ \noindent for the auxiliary variable.
\end{itemize}

We now list spectral properties of the iteration sub-matrices $W_{i,t}^m \!\in\! \mathbb{R}^{k_i+1 \times k_i+1}$ and $V_{i,t}^m \!\in\! \mathbb{R}^{d_i+1 \times d_i+1}$.  The\\ \noindent validity of these claims follows from extensions of work on the alternating direction method of multipliers \cite{BoleyD-jour2013a}.
\begin{itemize}
\item[] \-\hspace{0.5cm}{\small{\sf{\textbf{Lemma A.1.}}}} Suppose that the flag matrices across consecutive iterations $m$ of the hidden state and cause updates respectively satisfy $H_{i,t}^{m-1} \!=\! H_{i,t}^{m} \!=\! H_{i,t}^{m+1}$ and $M_{i,t}^{m-1} \!=\! M_{i,t}^{m} \!=\! M_{i,t}^{m+1}$.  The iteration matrices $W_{i,t}^m$ and $V_{i,t}^m$ are different at each step and satisfy:
\begin{itemize}
\item[] \-\hspace{0.5cm}(i) $\|W_{i,t}^m\|_2 \!\leq\! 1$ and $\|V_{i,t}^m\|_2 \!\leq\! 1$.  Also, $\|(I_{k_i \times k_i} \!-\! \ell_{i,t}^{-1}D_i^\top D_i)H_{i,t}^{m}\|_2 \!\leq\! 1$ and $\|(I_{d_i \times d_i} \!-\! 1/\ell_{i,t}')M_{i,t}^{m}\|_2 \!\leq\! 1$.
\item[] \-\hspace{0.5cm}(ii) For any $0 \!<\! \beta_m,\beta_m' \!\leq\! 1$, the eigenvalues of $W_{i,t}^m$ and $V_{i,t}^m$ lie in a closed circle in the real-complex plane that is centered at $(0,\frac{1}{2})$ and that has a radius of $\frac{1}{2}$.  If either of these iteration matrices has eigenvalues with absolute values of $\rho(W_{i,t}^m) \!=\! 1$ and $\rho(V_{i,t}^m) \!=\! 1$, then there must be no imaginary component.  If the step\\ \noindent sizes are such that $\beta_m,\beta_m' \!<\! 1$ and if $W_{i,t}^m$ and $V_{i,t}^m$ have eigenvalues of one, then these eigenvalues must have a complete set of eigenvectors.
\end{itemize}
The full iteration matrices have spectral decompositions $S_{i,t}^m \!=\! P_{i,t}^mJ^m_{i,t}(P^m_{i,t})^{-1}$ and $T_{i,t}^m \!=\! Q^m_{i,t} R^m_{i,t} (Q^{m}_{i,t})^{-1}$\\ \noindent where the block-diagonal eigenvalue matrices $J^m_{i,t}$ and $R^m_{i,t}$ have the form
\begin{equation*}
J^m_{i,t} \!=\! \begin{pmatrix}\!\begin{pmatrix}1 & 1\vspace{0.05cm}\\ 0 & 1\end{pmatrix}\!\!\! & 0 & 0\vspace{0.05cm}\\ 0 & I^m_{i,t}\!\! & 0\vspace{0.05cm}\\ 0 & 0 & J'_{i,t}{}^{\!\!\!\!m}\end{pmatrix}\!,\;\;    R^m_{i,t} \!=\! \begin{pmatrix}\!\begin{pmatrix}1 & 1\vspace{0.05cm}\\ 0 & 1\end{pmatrix}\!\!\! & 0 & 0\vspace{0.05cm}\\ 0 & I^m_{i,t}\!\! & 0\vspace{0.05cm}\\ 0 & 0 & R'_{i,t}{}^{\!\!\!\!m}\end{pmatrix}\!,
\end{equation*}
where $I^m_{i,t}$ is an appropriately sized identity matrix that will depend on the flag matrix at iteration $m$ for layer $i$ and batch $t$.  Here, we have the condition that the spectral radii $\rho(J'_{i,t}{}^{\!\!\!\!m}) \!<\! 1$ and $\rho(R'_{i,t}{}^{\!\!\!\!m}) \!<\! 1$.  
\end{itemize}

This lemma suggests that there are multiple local phases that can arise from matrix-based proximal-gradient recurrence.  These phases depend on properties of the flag matrix on the spectral characteristics of the full iteration matrices that define the matrix recurrence.  If the flag matrices remain the same across consecutive iterations, then the total-iteration-matrix operator remains invariant.  The structure of the spectrum for that operator controls the convergence behavior of the process.  If the flag matrix changes, then the set of active constraints at the current pass in the process has changed across consecutive iterations.  The current iteration is thus a transition to a different operator with a different eigenstructure.  The algorithm then searches for a correct set of active constraints.  The specific phases are distinguished by the eigenstructure of the total-iteration-matrix operator.

In what follows, we make explicit the phases that can emerge.  This is a slight extension of the work on the alternating direction method of multipliers \cite{BoleyD-jour2013a}.  It is readily applicable to the proximal-gradient case, however, due to the similar treatment of the sparsity terms via a splitting process.

\begin{itemize}
\item[] \-\hspace{0.5cm}{\small{\sf{\textbf{Proposition A.8.}}}} Suppose that the flag matrices across consecutive iterations $m$ of the hidden state and cause updates respectively satisfy $H_{i,t}^{m-1} \!=\! H_{i,t}^{m}$ and $M_{i,t}^{m-1} \!=\! M_{i,t}^{m}$.  Let the eigendecompositions of the total-iteration matrices be such that $S_{i,t}^m \!=\! P_{i,t}^mJ^m_{i,t}(P^m_{i,t})^{-1}$ and $T_{i,t}^m \!=\! Q^m_{i,t} R^m_{i,t} (Q^{m}_{i,t})^{-1}$, where $J^m_{i,t}$ and $R^m_{i,t}$ have block-diag-\\ \noindent onal forms.  Then, the iterates can belong to one of the following phases:
\begin{itemize}
\item[] \-\hspace{0.5cm}(i) Let the spectral radii $\rho(W_{i,t}^m) \!<\! 1$ and $\rho(V_{i,t}^m) \!<\! 1$.  In this case, the $\mathbb{R}_{0,+}^{2 \times 2}$ Jordan blocks in the upper-left\\ \noindent corners of the eigenvalue matrices $J_{i,t}^m$ and $R_{i,t}^m$ are not present.  The identity-matrix blocks $I_{i,t}^m \!\in\! \mathbb{R}^{1 \times 1}_+$ are\\ \noindent degenerate.  As long as the flag matrices $H_{i,t}^m$ and $M_{i,t}^m$ do not change across $m$ and if the iterates are close enough to the optimal solution, then linear convergence is achieved to that solution.  Such solutions are unique fixed points which are eigenvectors $[P_{i,t}^m]_{1:2k_i+1,2k_i+1}$ and $[Q_{i,t}^m]_{1:2d_i+1,2d_i+1}$ of $S_{i,t}^m$ and $T_{i,t}^m$ with unit eigenvalues.  If the eigenvectors are non-negative, then they satisfy the Karush-Kuhn-Tucker conditions for the state and cause inference costs.
\item[] \-\hspace{0.5cm}(ii) If $\rho(W_{i,t}^m) \!=\! 1$ and $\rho(V_{i,t}^m) \!=\! 1$, then $S_{i,t}^m$ and $T_{i,j}^m$ both have non-trivial $\mathbb{R}_{0,+}^{2 \times 2}$ Jordan blocks in the upper-left corners of $J_{i,t}^m$ and $R_{i,t}^m$.  There are no other eigenvalues on the unit circle.  The theory of the power method implies that the vector iterates will converge to an invariant subspace corresponding to the unit eigenvalue.  The presence of the non-trivial Jordan block implies the existence of a Jordan chain.  That is, there are non-zero vectors $\varphi,\varphi' \!\in\! \mathbb{R}^{2k_i+1}_{0,+}$ and $\phi,\phi' \!\in\! \mathbb{R}^{2d_i+1}_{0,+}$ such that the equivalence relations $(S_{i,t}^m \!-\! I_{2k_i+1,2k_i+1})\varphi \!=\! \varphi'$ and $(S_{i,t}^m \!-\! I_{2k_i+1,2k_i+1})\varphi' \!=\! 0$ along with $(T_{i,t}^m \!-\! I_{2d_i+1,2d_i+1})\phi \!=\! \phi'$ and $(T_{i,t}^m \!-\! I_{2d_i+1,2d_i+1})\phi' \!=\! 0$ are\\ \noindent satisfied.  Any vector that includes a component of the form $a\varphi \!+\! b\varphi'$ and $a\phi \!+\! b\phi'$, for $a,b \!\in\! \mathbb{R}$ would add\\ \noindent a constant factor $a\varphi$ and $a\phi$, respectively, to $S_{i,t}^m$ and $T_{i,t}^m$, plus descending lower-order terms from the other lesser eigenvalues.  If $H_{i,t}^m$ and $M_{i,t}^m$ do not change across $m$, then the state $w_{i,t}^m$ and cause $v_{i,t}^m$ iterates take constant-sized steps and either diverge or drive some component negative, which results in a change in the iteration matrices $W_{i,t}^m$ and $V_{i,t}^m$.
\item[] \-\hspace{0.5cm}(iii) Suppose that $\rho(W_{i,t}^m) \!=\! 1$ and $\rho(V_{i,t}^m) \!=\! 1$, but $S_{i,t}^m$ and $T_{i,j}^m$ have no non-diagonal Jordan block for that\\ \noindent eigenvalue.  If we assume that the solution is unique, then the unit eigenvalues of $S_{i,t}^m$ and $T_{i,j}^m$ must be simple.  There are no other eigenvalues on the unit circle.  If the iterates are close enough to the optimal solution, then they linearly converge to it, as the inference process behaves similarly to a Von Mises iteration.  These unique, fixed-point solutions are, by definition, the eigenvectors $[P_{i,t}^m]_{1:2k_i+1,2k_i+1}$ and $[Q_{i,t}^m]_{1:2d_i+1,2d_i+1}$ of $S_{i,t}^m$ and $T_{i,t}^m$ for the unit eigenvalues.  The convergence rate is determined by the next-largest eigenvalues in the absolute value, that is, the largest eigenvalues of $J'_{i,t}{}^{\!\!\!\!m}$ and $R'_{i,t}{}^{\!\!\!\!m}$, as long as the flag matrices $H_{i,t}^m$ and $M_{i,t}^m$ do not change across $m$.  This phase cannot be last in the inference process, as the search will eventually jump to a different one due to the eigenvalue properties.
\end{itemize}
Now, suppose that $H_{i,t}^{m-1} \!\neq\! H_{i,t}^{m}$ and $M_{i,t}^{m-1} \!\neq\! M_{i,t}^{m}$ for iterations $m$.  In this case, the iteration operator does not remain invariant over more than one pass.  The iteration matrices $W_{i,t}^m$ and $V_{i,t}^m$ could match one of the conditions in the above phases.  They could also have the following eigenstructure associated with a fourth phase:
\begin{itemize}
\item[] \-\hspace{0.5cm}(iv) $W_{i,t}^m$ and $V_{i,t}^m$ have eigenvalues with absolute value one, but not equal to one.  This occurs when the iterates transition to a new set of active constraints.  The next pass will result in a different operator with a different flag matrix.
\end{itemize}
\end{itemize}
\begin{itemize}
\item[] \-\hspace{0.5cm}{\small{\sf{\textbf{Proof:}}}} For $S_{i,t}^m$, the upper-left sub-matrix, which is defined by $W_{i,t}^m$, contributes to the eigenvalue blocks $I_{i,t}^m$ and $J'_{i,t}{}^{\!\!\!\!m}$ of $J_{i,t}^m$.  Here, we assume that the spectral decomposition is written as $S_{i,t}^m \!=\! P_{i,t}^m J_{i,t}^m (P_{i,t}^m)^{-1}$, with $J_{i,t}^m$ having the form defined in lemma A.1.  Both $W_{i,t}^m$ and $S_{i,t}^m$ have the same set of eigenvalues with equivalent geometric and algebraic multiplicities, except when an eigenvalue has an absolute value of one.  No eigenvalue with an absolute value of one can have a non-diagonal Jordan block.  Hence, the blocks, $I_{i,t}^m$ and $J'_{i,t}{}^{\!\!\!\!m}$, corresponding to those eigenvalues must be diagonal.  

If $W_{i,t}^m$ has no eigenvalue equal to one, then $S_{i,t}^m$ has a simple eigenvalue that is one.  In this case, the number\\ \noindent of times an eigenvalue appears as a root of the characteristic polynomial increases by one.  The eigenspace dimensionality either increases by one or stays the same.  Alternatively, the algebraic and geometric multiplicities for the unit eigenvalue for $W_{i,t}^m$ differ.  This implies that there is a Jordan block, which is given by the $\mathbb{R}_{0,+}^{2 \times 2}$ sub-matrix in the top-left corner of $J_{i,t}^m$.

The eigenvalue properties of $S_{i,t}^m$ give rise to the three phases listed above.  A similar argument can be applied to the cause iteration matrices $T_{i,t}^m$.  The fourth phase is trivial to show.$\;\;\footnotesize\selectfont\blacksquare$


\end{itemize}

Such results indicate that there are local search regimes in the inference process where the convergence is quicker than the global convergence rate.  We can specify the conditions when this will occur.
\begin{itemize}
\item[] \-\hspace{0.5cm}{\small{\sf{\textbf{Proposition A.9.}}}} Let the state and cause inference recurrences be defined as in propositions A.6 and A.7, respectively, for the accelerated proximal gradient search.  For a non-accelerated search, they become
\begin{equation*}
\begin{pmatrix}
w'_{i,t}{}^{\!\!\!\!m+1}\vspace{0.05cm}\\
1
\end{pmatrix} =
\begin{pmatrix}
W_{i,t}'{}^{\!\!\hspace{-0.075cm}m} & \ell_{i,t}D_i^\top\kappa_{i-1,t} \!-\! \lambda_{i,t}\ell_{i,t}^{-1}(I_{k_i \times k_i} \!-\! \ell_{i,t}^{-1}D_i^\top D_i)s_{i,t}' \!+\! \alpha_i\ell_{i,t}f_{i,t}^m\vspace{0.05cm}\\
0 & 1
\end{pmatrix}\!
\begin{pmatrix}
w'_{i,t}{}^{\!\!\!\!m+1}\vspace{0.05cm}\\
1
\end{pmatrix}
\end{equation*}
where $W_{i,t}'{}^{\!\!\hspace{-0.075cm}m} \!=\! (I_{k_i \times k_i} \!-\! \ell_{i,t}^{-1}D_i^\top D_i)(H_{i,t}'{}^{\!\!\!\!m})^2$ and $s_{i,t}'{}^{\!\!\!\!m} \!=\! \textnormal{\sc sign}(\textnormal{\sc shrink}(w'_{i,t}{}^{\!\!\!\!m};\lambda_{i,t}/\ell_{i,t}))$.  The flag matrix is such that $H_{i,t}'{}^{\!\!\!\!m} \!=\! \textnormal{\sc diag}(s_{i,t}'{}^{\!\!\!\!m})$.  For the non-accelerated cause inference, we have the recurrence relation
\begin{equation*}
\begin{pmatrix}
v'_{i,t}{}^{\!\!\!\!m+1}\vspace{0.05cm}\\
1
\end{pmatrix} =
\begin{pmatrix}
V_{i,t}'{}^{\!\!\!m} & 2\eta_i'I_{d_i \times d_i} \kappa_{i,t}'/\ell_{i,t}' \!+\! (I_{d_i \times d_i} \!-\! 1/\ell_{i,t}') q_{i,t}'{}^{\!\!\!\!m} \!-\! \alpha_i'g_{i,t}^m/\ell_{i,t}'\vspace{0.05cm}\\
0 & 1
\end{pmatrix}\!
\begin{pmatrix}
v'_{i,t}{}^{\!\!\!\!m+1}\vspace{0.05cm}\\
1
\end{pmatrix}
\end{equation*}
where $V_{i,t}'{}^{\!\!\!m} \!=\! (I_{d_i \times d_i} \!-\! 1/\ell_{i,t}')(M_{i,t}'{}^{\!\!\!\!m})^2$ and $q_{i,t}'{}^{\!\!\!\!m} \!=\! \textnormal{\sc sign}(\textnormal{\sc shrink}(v'_{i,t}{}^{\!\!\!\!m};\lambda_{i}/\ell'_{i,t}))$.  The flag matrix is such that\\ \noindent $M_{i,t}'{}^{\!\!\!\!m} \!=\! \textnormal{\sc diag}(q_{i,t}'{}^{\!\!\!\!m})$.  We have that:
\begin{itemize}
\item[] \-\hspace{0.5cm}(i) For the second phase, the constant step-size vector has the form $(1\!-\! \beta_m)^{-1}(\varphi,\varphi,0)^\top$, $W_{i,t}'{}^{\!\!\hspace{-0.075cm}m}\varphi \!=\! \varphi$,\\ \noindent with $\varphi \!\in\! \mathbb{R}^{k_i}$ being a scaled eigenvector of the state-inference total-iteration matrix.  Likewise, for the\\ \noindent cause inference, the constant step-size vector has the form $(1\!-\! \beta_m')^{-1}(\phi,\phi,0)^\top$, where $V_{i,t}'{}^{\!\!\hspace{-0.05cm}m}\phi \!=\! \phi$, with\\ \noindent $\phi \!\in\! \mathbb{R}^{d_i}$ being a scaled eigenvector.  Since $\beta_m,\beta_m'$ have a limit of one, the constant-step-size vector is\\ \noindent larger than the one for the states, $(\varphi',0)$, $W_{i,t}'{}^{\!\!\hspace{-0.075cm}m}\varphi' \!=\! \varphi'$, in the non-accelerated proximal-gradient case.   The constant-step-size vector in the accelerated case is also larger than the one for the causes, $(\phi',0)$, $V_{i,t}'{}^{\!\!\hspace{-0.05cm}m}\phi' \!=\! \phi'$, in the case of non-accelerated proximal gradients.
\item[] \-\hspace{0.5cm}(ii) In the first and third phases, if $1 \!>\! \rho(W_{i,t}'{}^{\!\!\hspace{-0.075cm}m}) \!>\! \beta_m$ and $1 \!>\! \rho(V_{i,t}'{}^{\!\!\!m}) \!>\! \beta_m'$, then the accelerated proximal gradient scheme will be faster than the non-accelerated case for the state and cause inference.  It will be slower, however, if $1 \!>\! \beta_m \!>\! \rho(W_{i,t}'{}^{\!\!\hspace{-0.075cm}m})$ and $1 \!>\! \beta_m' \!>\! \rho(V_{i,t}'{}^{\!\!\!m})$.  When $\beta_m \!>\! \rho(W_{i,t}'{}^{\!\!\hspace{-0.075cm}m})$ and $\beta_m' \!>\! \rho(V_{i,t}'{}^{\!\!\!m})$, then the\\ \noindent largest eigenvalues of $W_{i,t}^m$ and $V_{i,t}^m$ must be a pair of complex conjugates.  According to the theory of Von Mises iterations, the convergence will oscillate between the two complex numbers and the search will not be monotonically decreasing across iterations.  The accelerated case will hence be slower than the non-accelerated case, as $\rho(W_{i,t}'{}^{\!\!\!\hspace{-0.01cm}m})$ and $\rho(V_{i,t}'{}^{\!\!\!m})$ will remain fixed for a specific phase, while the steps $\beta_m$ and $\beta_m'$ will monotonically increase.
\end{itemize}
\end{itemize}
\begin{itemize}
\item[] \-\hspace{0.5cm}{\small{\sf{\textbf{Proof:}}}} For part (i), a single update has the form $(w_{i,t}^{m+1},w_{i,t}^m,1)^\top \!+\! (\varphi_1,\varphi_2,0)^\top$.  There exists a Jordan\\ \noindent block in $J_{i,t}^m$ and hence a Jordan chain.  Therefore, $(S_{i,t}^m - I_{2k_i + 1 \times 2k_i + 1})(w_{i,t}^{m+1},w_{i,t}^m,1)^\top \!=\! (\varphi_1,\varphi_2,0)^\top$ and\\ \noindent $S_{i,t}^m(\varphi_1,\varphi_2,0)^\top \!=\! (\varphi_1,\varphi_2,0)^\top$.  It can be seen that $\varphi_1 \!=\! \varphi_2$.  This implies $W_{i,t}'{}^{\!\!\hspace{-0.075cm}m}\varphi \!=\! \varphi$ for the accelerated case.\\ \noindent  Moreover,
\begin{equation*}
(S_{i,t}^m - I_{2k_i + 1 \times 2k_i + 1}) \!\begin{pmatrix}w_{i,t}^{m+1}\vspace{0.05cm}\\ w_{i,t}^{m}\vspace{0.05cm}\\ 1\end{pmatrix} \!=\! 
\begin{pmatrix} 
((1 \!+\! \beta_m)W_{i,t}'{}^{\!\!\hspace{-0.075cm}m} \!-\! I_{k_i \times k_i})w_{i,t}^m \!-\! \beta_m W_{i,t}'{}^{\!\!\hspace{-0.075cm}m} w_{i,t}^{m-1} \!+\! (I_{k_i \times k_i} \!-\! \ell_{i,t}^{-1} D_i^\top D_i)z_{i,t}^m\vspace{0.05cm}\\
w_{i,t}^{m} \!-\! w_{i,t}^{m-1}\vspace{0.05cm}\\
0
\end{pmatrix}
\end{equation*}
where $(S_{i,t}^m \!-\! I_{2k_i + 1 \times 2k_i + 1})(w_{i,t}^{m+1},w_{i,t}^{m},1)^\top \!=\! (1 \!-\! \beta_m)^{-1}(\varphi,\varphi,0)^\top$.  For this to occur, we must have that\\ \noindent $w_{i,t}^{m-1} \!=\! w_{i,t}^{m} \!+\! (\beta_m \!-\! 1)^{-1}\varphi$.  Similar arguments apply to the causes.  The analysis is similar for the non-accel-\\ \noindent erated proximal-gradient case.

For part (ii), we prove properties for the states and note that they extend to the causes with few changes.  Let $(v_1,v_2)^\top$ be an eigenvector of $W_{i,t}^m$.  We have that
\begin{equation*}
W_{i,t}^m (v_1,v_2)^\top = 
\begin{pmatrix}
(1 + \beta_m)\rho(W_{i,t}^m) v_2 + \beta_m (I_{k_i \times k_i} \!-\! \ell_{i,t}D_i^\top D_i)(H_{i,t}^m)^2 v_2\vspace{0.05cm}\\
v_2
\end{pmatrix} = \rho(W_{i,t}^m)\begin{pmatrix} \rho(W_{i,t}^m)v_2\vspace{0.05cm}\\ v_2 \end{pmatrix}
\end{equation*}
since $v_1 \!=\! \rho(W_{i,t}^m) v_2$.  Hence, $(I_{k_i \times k_i} \!-\! \ell_{i,t}D_i^\top D_i)(H_{i,t}^m)^2 v_2 \!=\! \rho(W_{i,t}^m)^2 v_2/((1 \!+\! \beta_m)\rho(W_{i,t}^m) \!-\! \beta_m)$.  This implies that $\rho(W_{i,t}^m)^2 \!+\! \beta_m\rho(K_{i,t}^m) \!-\! (1 \!+\! \beta_m)\rho(W_{i,t}^m)\rho(W_{i,t}'{}^{\!\!\hspace{-0.075cm}m}) \!=\! 0$, where $W_{i,t}'{}^{\!\!\hspace{-0.075cm}m} \!=\! (I_{k_i \times k_i} \!-\! \ell_{i,t}D_i^\top D_i)(H_{i,t}^m)^2$.  It can be\\ \noindent seen that $\rho(W_{i,t}^m)$ has real-valued roots if $4\beta_m(1 \!+\! \beta_m)^2 \!<\! \rho(W_{i,t}^m)$ and complex-valued roots otherwise.  When\\ \noindent there are real-valued roots, then $\rho(W_{i,t}'{}^{\!\!\hspace{-0.075cm}m}) \!>\! \beta_m$.  Moreover, $\rho(W_{i,t}^m) \!=\! (1 \!+\! \beta_m)\rho(W_{i,t}'{}^{\!\!\hspace{-0.075cm}m})/2 + ((1 \!+\! \beta_m)^2 \rho(W_{i,t}'{}^{\!\!\hspace{-0.075cm}m})^2/4$\vspace{-0.025cm}\\ \noindent $ - \beta_m \rho(W_{i,t}'{}^{\!\!\hspace{-0.075cm}m}))^{1/2} \!<\! \rho(W_{i,t}'{}^{\!\!\hspace{-0.075cm}m})$, which follows from the fact that $\beta_m \!<\! 1$ as per its definition.  When there are com-\\ \noindent plex-valued roots, then $|\rho(W_{i,t}^m)| \!=\! (\beta_m\rho(W_{i,t}'{}^{\!\!\hspace{-0.075cm}m}))^{1/2}$ and hence $|\rho(W_{i,t}^m)| \!<\! \rho(W_{i,t}'{}^{\!\!\hspace{-0.075cm}m})$ whenever $\rho(W_{i,t}'{}^{\!\!\hspace{-0.075cm}m}) \!>\! \beta_m$.  If, however, $\beta_m \!>\! \rho(W_{i,t}'{}^{\!\!\hspace{-0.075cm}m})$, then $|\rho(W_{i,t}^m)| \!>\! \rho(W_{i,t}'{}^{\!\!\hspace{-0.075cm}m})$.

Both the accelerated and non-accelerated proximal gradients reduce to a Von Mises iteration in the first and third phases.  The rate of convergence is, respectively, determined by $|\rho(W_{i,t}^m)|$ and $|\rho(W_{i,t}'{}^{\!\!\hspace{-0.075cm}m})|$.$\;\;\footnotesize\selectfont\blacksquare$
\end{itemize}

This result provides insight into why Nesterov inertial sequences are worse than the one that we employed.  It is trivial to demonstrate that the step sizes for a Nesterov inertial sequence will be strictly greater than the step sizes for our polynomial inertial sequence.  The former will reach the eigenvalue-versus-inertial-sequence inequality conditions more quickly, yielding an inference slowdown whenever the search is in either the first or third phase.  Such phases typically occur near the beginning of the inference process and occupy a majority of the overall process.  Moreover, severe cost rippling will be encountered due to the emergence of complex-conjugate eigenvalues in the iteration matrices.  The state and cause iterates thus alternate between locally minimizing and maximizing the LASSO costs.  Over a great many iterations, they generally lead to average LASSO cost decreases.  However, this does not always occur.  The average LASSO loss across iterations can remain unchanged, leading to learning stagnation.

In contrast to the Nesterov-based inertial sequence used by DCPNs, that used by ADPCNs will likely never reach either the first or third phases where oscillation occurs.  This is because the inertial sequence often grows too slowly compared to the iteration-matrix eigenvalues.  If either phase is reached, then it will typically occur far later during inference than in the Nesterov-sequence case.  This allows for our accelerated strategy to take advantage of the faster convergence rate for a greater number of iterations.  The iterates are usually strictly decreasing the LASSO costs throughout much of the inference process.  For the remainder, they are monotonically non-increasing.  This insinuates that the states and causes continuously improve their characterization of the stimuli.  Often, inference terminates before oscillation can occur.  In other cases, the inference process will restart, thereby delaying the onset of the first or third phases while still meaningfully decreasing the LASSO costs.


There has been a great deal of work on multiple restarts \cite{OdonoghueB-jour2013a,KimD-jour2018a,FercoqO-jour2019a} of iterative methods for convex and non-convex costs.  We have found that certain approaches are adept at suppressing cost oscillations.  It is likely that they suitably change the spectral properties of the iteration matrices to force the inference process back into either the first or the third phase where it exhibits an improved convergence rate.  

Here, we have focused on assessing convergence and the convergence rate for DPCN and ADPCN inference without multiple restarts.  This was done, as we mentioned above, to highlight that the learning improvement we empirically observed stemmed from the inertial-sequence choice.  In our future work, we will analyze the theoretical behaviors of integrating restart procedures into our inference process.  We will prove that they can they entirely preempt cost oscillation.  They also guarantee monotonic LASSO cost decreases.  This will demonstrate that we can run the feed-forward, bottom-up inference process over an arbitrarily large number of iterations without slowdown concerns.  Increasing the number of inference steps per forward pass will likely further reduce the number of stimuli presentations needed to uncover good network parameters.  We also will prove performing multiple restarts in a principled way will still guarantee convergence to solutions without adversely impacting the improved convergence rate that we have established.

\setstretch{0.95}\fontsize{9.75}{10}\selectfont
\putbib
\end{bibunit}

\clearpage\newpage

\begin{bibunit}
\bstctlcite{IEEEexample:BSTcontrol}

\subsection*{\small{\sf{\textbf{Appendix B}}}}

\renewcommand{\thefigure}{B.\arabic{figure}}
\setcounter{figure}{0}

\begin{figure*}[h!]
   \hspace{-0.1cm}\begin{tabular}{l}\includegraphics[]{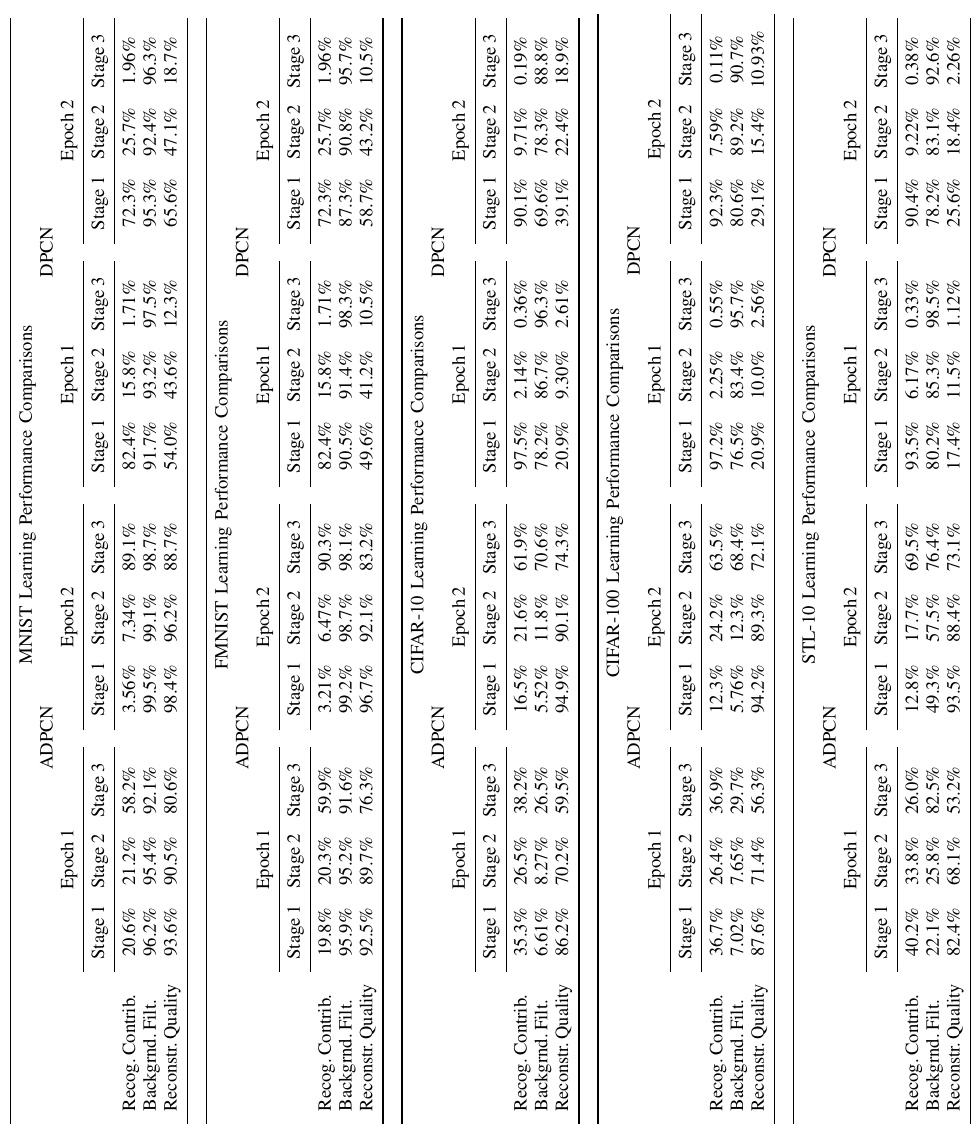}\vspace{-0.15cm}\\
   \hspace{1.5cm}{\footnotesize(a)}\hspace{3.025cm}{\footnotesize(b)}\hspace{3.0cm}{\footnotesize(c)}\hspace{3cm}{\footnotesize(d)}\hspace{3cm}{\footnotesize(e)}
   \end{tabular}
   \caption[]{\fontdimen2\font=1.55pt\selectfont Statistical summaries of ADPCN and convolutional DPCN results across two epochs.  For each stimuli dataset, we report the stage-wise contribution of the causes on the recognition rate.  We do this by iteratively appending the causes from each stage to assess the discrimination change.  Higher values indicate that the causes from a particular stage are more relevant for making classification decisions.  We also list the ability of each stage to either remove or smooth background details.  We do this by assessing the top-down and bottom-up image saliency to delineate object boundaries.  We then quantify the high-frequency content, relative to the original stimuli, for everything not within the object boundaries.  Higher values denote that superfluous, non-object visual details are removed.  Lastly, we provide the stage-wise reconstruction quality, in terms of the mean-squared-error percentage.  Higher values indicate that a stage's reconstruction response aligns well with the input stimuli.  All statistics were averaged across five Monte Carlo simulations with random parameter initializations.\vspace{-0.1cm}}
   \label{fig:recog-tables}
\end{figure*}

\clearpage\newpage


\renewcommand{\thefigure}{B.\arabic{figure}}
\setcounter{figure}{1}

\begin{figure}[h!]
\hspace{0.0cm}\begin{minipage}[]{1\textwidth}
\begin{multicols}{3}
{\small
\begin{tabular}{c@{\hspace{8pt}}c@{\hspace{8pt}}c@{\hspace{6pt}}@{\hskip 0.1cm} c@{\hspace{8pt}}c@{\hspace{8pt}}c@{\hspace{8pt}}@{\hskip 0.1cm} c@{\hspace{8pt}}c@{\hspace{8pt}}c@{\hspace{8pt}}@{\hskip 0.1cm} c@{\hspace{8pt}}c@{\hspace{8pt}}c@{\hspace{8pt}}}

\toprule
\multicolumn{3}{c}{{{MNIST Error\! Rate\! Comparison}}}\vspace{0.05cm}\\
Method & Learning\! Style & Error\\

\cmidrule[0.4pt](){1-3}%

\textcolor{myred}{DPCN} \cite{PrincipeJC-jour2014a} & \textcolor{myred}{Unsupervised} & \textcolor{myred}{19.2\%}\\
AE \cite{BengioY-coll2007a} & Unsupervised & 18.8\%\\
GAN \cite{RadfordA-conf2016a} & Unsupervised & 17.2\%\\
CRPN \cite{ChalasaniR-jour2015a} & Unsupervised & 5.65\%\\
IMSAT \cite{HuW-conf2017a} & Unsupervised & 1.60\%\\
IIC \cite{JiX-conf2019a} & Unsupervised & 1.60\%\\
SCAE \cite{KosiorekA-coll2019a} & Unsupervised & 1.00\% \\
EBSR \cite{RanzatoM-coll2007a} & Unsupervised & 0.39\%\\
\textcolor{mygreen}{ADPCN} & \textcolor{mygreen}{Unsupervised} & \textcolor{mygreen}{0.31\%}\\
\textcolor{myred}{MON} \cite{GoodfellowIJ-conf2013a} & \textcolor{myred}{Supervised} & \textcolor{myred}{0.45\%}\\
RCNN \cite{LiangM-conf2015a} & Supervised & 0.31\%\\
\textcolor{mygreen}{MCNN} \cite{CireganD-conf2012a} & \textcolor{mygreen}{Supervised} & \textcolor{mygreen}{0.23\%}\\
\bottomrule
\end{tabular}}

{\small
\begin{tabular}{c@{\hspace{8pt}}c@{\hspace{8pt}}c@{\hspace{6pt}}@{\hskip 0.1cm} c@{\hspace{8pt}}c@{\hspace{8pt}}c@{\hspace{8pt}}@{\hskip 0.1cm} c@{\hspace{8pt}}c@{\hspace{8pt}}c@{\hspace{8pt}}@{\hskip 0.1cm} c@{\hspace{8pt}}c@{\hspace{8pt}}c@{\hspace{8pt}}}

\toprule
\multicolumn{3}{c}{{{FMNIST Error\! Rate\! Comparison}}}\vspace{0.05cm}\\
Method & Learning\! Style & Error\\

\cmidrule[0.4pt](){1-3}%

\textcolor{myred}{DPCN} \cite{PrincipeJC-jour2014a} & \textcolor{myred}{Unsupervised} & \textcolor{myred}{42.7\%}\\
kSCN \cite{ZhangT-conf2018a} & Unsupervised & 39.9\%\\
CRPN \cite{ChalasaniR-jour2015a} & Unsupervised & 15.5\%\\
\textcolor{mygreen}{ADPCN} & \textcolor{mygreen}{Unsupervised} & \textcolor{mygreen}{3.51\%}\\
\textcolor{mygreen}{TLML} \cite{YuB-conf2018a} & \textcolor{mygreen}{Semi-super.} & \textcolor{mygreen}{11.2\%}\\
\textcolor{myred}{ZSDA} \cite{PengKC-conf2018a} & \textcolor{myred}{Supervised} & \textcolor{myred}{15.5\%}\\
DAGH \cite{ChenY-conf2019a} & Supervised & 6.30\%\\
DCAP \cite{RajasegaranJ-conf2019a} & Supervised & 5.54\%\\
DRBC \cite{SabourS-coll2017a} & Supervised & 4.30\%\\
DNET \cite{HuangG-conf2017a} & Supervised & 4.60\%\\
LES \cite{NoklandA-conf2019a} & Supervised & 4.14\%\\
\textcolor{mygreen}{JOUT} \cite{WangS-conf2019a} & \textcolor{mygreen}{Supervised} & \textcolor{mygreen}{2.87\%}\\
\bottomrule
\end{tabular}}

{\small \begin{tabular}{c@{\hspace{8pt}}c@{\hspace{8pt}}c@{\hspace{6pt}}@{\hskip 0.1cm} c@{\hspace{8pt}}c@{\hspace{8pt}}c@{\hspace{8pt}}@{\hskip 0.1cm} c@{\hspace{8pt}}c@{\hspace{8pt}}c@{\hspace{8pt}}@{\hskip 0.1cm} c@{\hspace{8pt}}c@{\hspace{8pt}}c@{\hspace{8pt}}}

\toprule
\multicolumn{3}{c}{{{CIFAR-10 Error\! Rate\! Comparison}}}\vspace{0.05cm}\\
Method & Learning\! Style & Error\\
\cmidrule[0.4pt](){1-3}%

\textcolor{myred}{DPCN} \cite{PrincipeJC-jour2014a} & \textcolor{myred}{Unsupervised} & \textcolor{myred}{68.5\%}\\
NOMP \cite{LinTH-conf2014a} & Unsupervised & 39.2\%\\
CRPN \cite{ChalasaniR-jour2015a} & Unsupervised & 32.7\%\\
RFL \cite{JiaY-conf2012a} & Unsupervised & 16.9\%\\
\textcolor{mygreen}{ADPCN} & \textcolor{mygreen}{Unsupervised} & \textcolor{mygreen}{12.7\%}\\
\textcolor{myred}{MON} \cite{GoodfellowIJ-conf2013a} & \textcolor{myred}{Supervised} & \textcolor{myred}{9.38\%}\\
DASN \cite{StollengaMF-coll2014a} & Supervised & 9.22\%\\
ACN \cite{SpringenbergJT-conf2015a} & Supervised & 9.08\%\\
SPNET \cite{RippelO-coll2015a} & Supervised & 8.60\%\\
HNET \cite{SrivastavaRK-coll2015a} & Supervised & 7.69\%\\
LAF \cite{AgostinelliF-conf2015a} & Supervised & 7.51\%\\
\textcolor{mygreen}{RCNN} \cite{LiangM-conf2015a} & \textcolor{mygreen}{Supervised} & \textcolor{mygreen}{7.09\%}\\
\bottomrule
\end{tabular}}

\end{multicols}

\begin{multicols}{3}
{\small \begin{tabular}{c@{\hspace{8pt}}c@{\hspace{8pt}}c@{\hspace{6pt}}@{\hskip 0.1cm} c@{\hspace{8pt}}c@{\hspace{8pt}}c@{\hspace{8pt}}@{\hskip 0.1cm} c@{\hspace{8pt}}c@{\hspace{8pt}}c@{\hspace{8pt}}@{\hskip 0.1cm} c@{\hspace{8pt}}c@{\hspace{8pt}}c@{\hspace{8pt}}}

\toprule
\multicolumn{3}{c}{{{CIFAR-100 Error\! Rate\! Comparison}}}\vspace{0.05cm}\\
Method & Learning\! Style & Error\\
\cmidrule[0.4pt](){1-3}%

\textcolor{myred}{DPCN} \cite{PrincipeJC-jour2014a} & \textcolor{myred}{Unsupervised} & \textcolor{myred}{95.6\%}\\
AEVB \cite{KingmaDP-conf2013a} & Unsupervised & 84.8\%\\
DEC \cite{XieJ-conf2016a} & Unsupervised & 81.5\%\\
DAIC \cite{ChangJ-conf2017a} & Unsupervised & 76.2\%\\
DCCM \cite{WuJ-conf2019a}\! & Unsupervised & 67.3\%\\
CRPN \cite{ChalasaniR-jour2015a} & Unsupervised & 59.7\%\\
\textcolor{mygreen}{ADPCN} & \textcolor{mygreen}{Unsupervised} & \textcolor{mygreen}{39.7\%}\\
\textcolor{myred}{PMO} \cite{SpringenbergJT-conf2014a} & \textcolor{myred}{Supervised} & \textcolor{myred}{38.1\%}\\
TREE \cite{SrivastavaN-coll2013a} & Supervised & 36.8\%\\
SBO \cite{SnoekJ-conf2015a} & Supervised & 27.4\%\\
INIT \cite{MishkinD-conf2016a} & Supervised & 26.3\%\\
\textcolor{mygreen}{DNET} \cite{HuangG-conf2017a} & \textcolor{mygreen}{Supervised} & \textcolor{mygreen}{17.1\%}\\
\bottomrule
\end{tabular}}

{\small \begin{tabular}{c@{\hspace{8pt}}c@{\hspace{8pt}}c@{\hspace{6pt}}@{\hskip 0.1cm} c@{\hspace{8pt}}c@{\hspace{8pt}}c@{\hspace{8pt}}@{\hskip 0.1cm} c@{\hspace{8pt}}c@{\hspace{8pt}}c@{\hspace{8pt}}@{\hskip 0.1cm} c@{\hspace{8pt}}c@{\hspace{8pt}}c@{\hspace{8pt}}}

\toprule
\multicolumn{3}{c}{{{STL-10 Error\! Rate\! Comparison}}}\vspace{0.05cm}\\
Method & Learning\! Style & Error\\
\cmidrule[0.4pt](){1-3}%

\textcolor{myred}{DPCN} \cite{PrincipeJC-jour2014a} & \textcolor{myred}{Unsupervised} & \textcolor{myred}{87.1\%}\\
SWAE \cite{ZhaoJ-conf2016a} & Unsupervised & 72.9\%\\
JULE \cite{YangJ-conf2016a} & Unsupervised & 72.3\%\\
DCNN \cite{ZeilerMD-conf2010a}\! & Unsupervised & 70.1\%\\
DEC \cite{XieJ-conf2016a} & Unsupervised & 64.1\%\\
CRPN \cite{ChalasaniR-jour2015a} & Unsupervised & 55.5\%\\
SRF \cite{CoatesA-coll2011a} & Unsupervised & 39.9\%\\
\textcolor{mygreen}{ADPCN} & \textcolor{mygreen}{Unsupervised} & \textcolor{mygreen}{25.3\%}\\
\textcolor{myred}{CKN} \cite{MairalJ-coll2014a} & \textcolor{myred}{Supervised} & \textcolor{myred}{37.6\%}\\
MTBO \cite{SwerskyK-coll2013a} & Supervised & 29.9\%\\
SSTN \cite{OyallonE-conf2017a} & Supervised & 23.4\%\\
\textcolor{mygreen}{RESN} \cite{HeK-conf2016a} & \textcolor{mygreen}{Supervised} & \textcolor{mygreen}{14.2\%}\\
\bottomrule
\end{tabular}}

\end{multicols}
\end{minipage}\vspace{0.15cm}
   \caption[]{\fontdimen2\font=1.55pt\selectfont Performance errors for the MNIST, FMNIST, CIFAR-10, CIFAR-100, and STL-10 datasets.  For each of the referenced methods, we either report the test-set classification errors that the authors listed or report the best known test-set classification errors obtained using that approach.  We also specify whether the chosen approaches were predominantly unsupervised, semi-supervised, or supervised.  The best methods for each learning category are highlighted in green.  The worst are highlighted in red.\vspace{-0.4cm}}
   \label{fig:class-results}
\end{figure}

\setstretch{0.95}\fontsize{9.75}{10}\selectfont
\putbib
\end{bibunit}

\clearpage\newpage

\subsection*{\small{\sf{\textbf{Appendix C}}}}

\RaggedRight\parindent=1.5em
\fontdimen2\font=2.1pt\selectfont
\singlespacing

\renewcommand{\thefigure}{C.\arabic{figure}}
\setcounter{figure}{0}

\begin{figure*}[h!]\vspace{-0.75cm}
   \hspace{-0.3cm}\begin{tabular}{c}
      \includegraphics[width=6.65in]{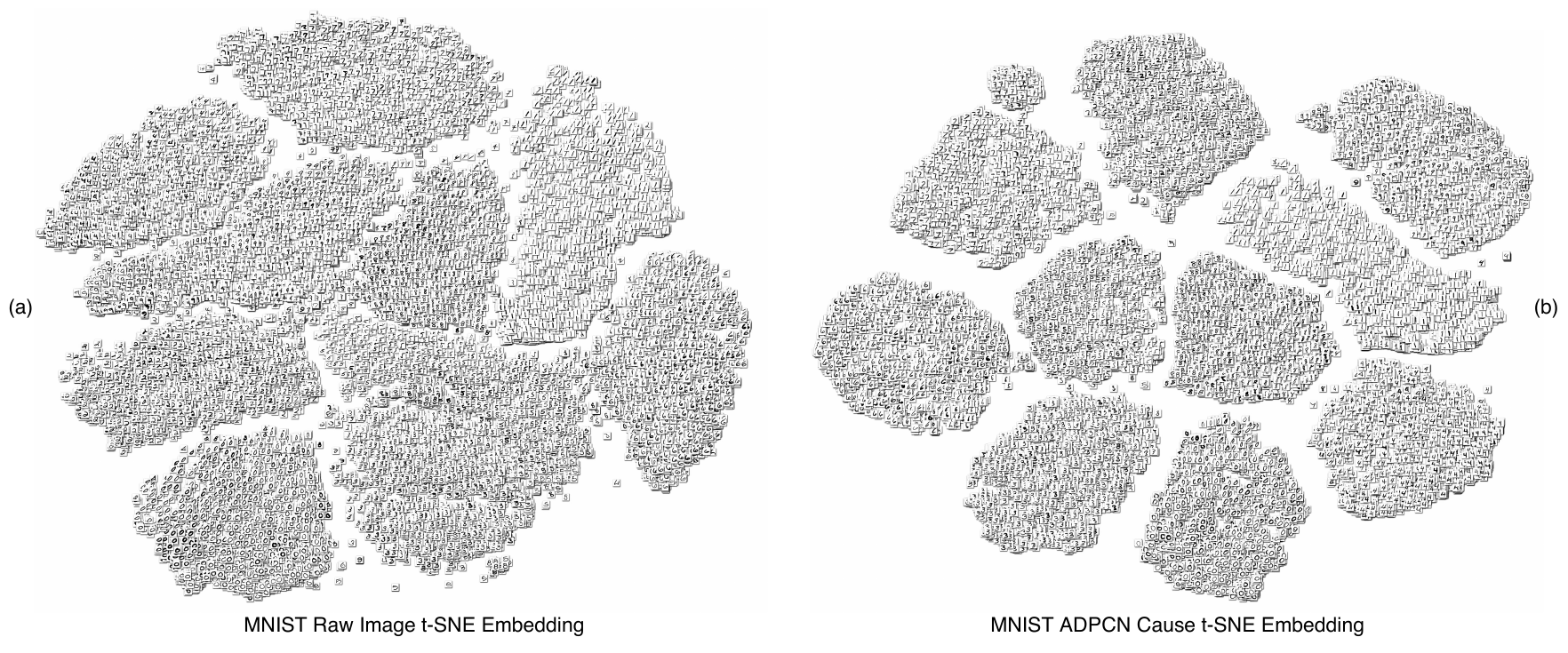}\vspace{-0.05cm}\\
      \includegraphics[width=6.65in]{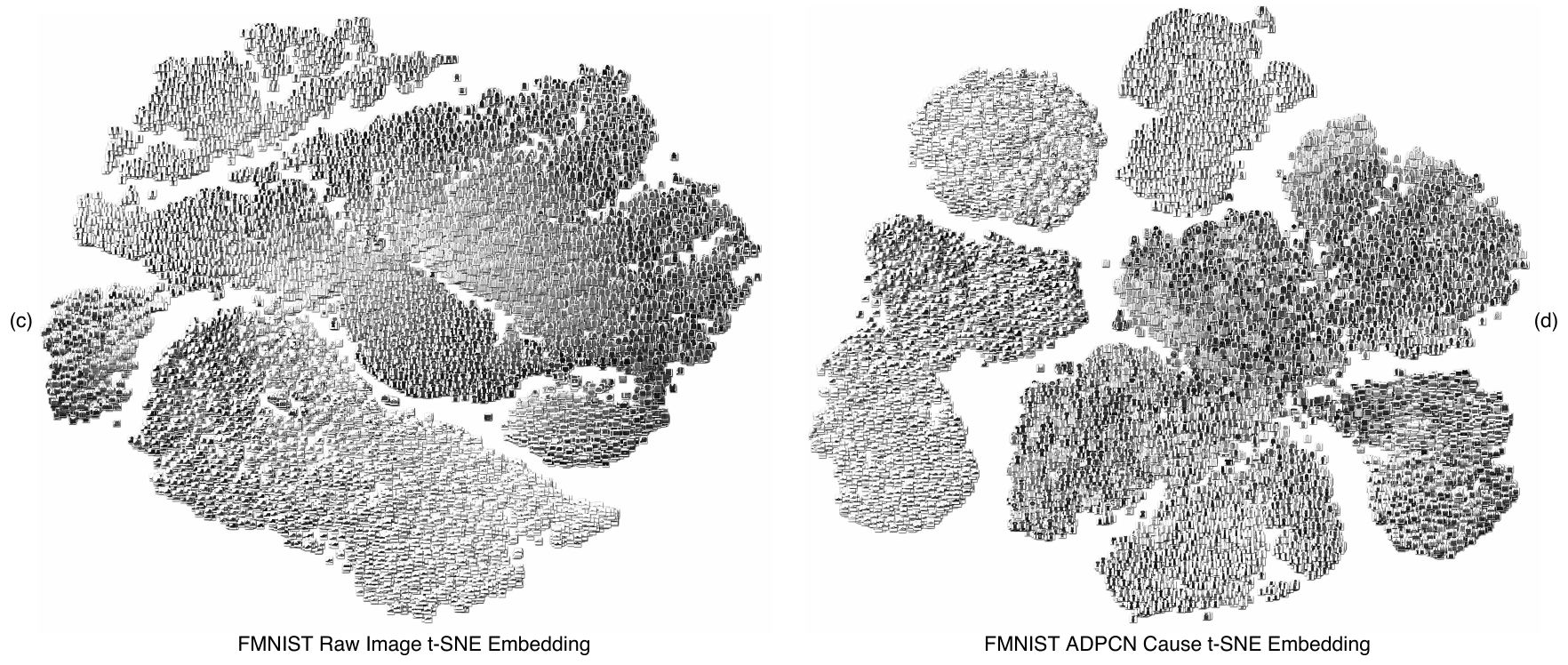}\vspace{-0.1cm}
   \end{tabular}
   \caption[]{\fontdimen2\font=1.55pt\selectfont A comparison of t-SNE embeddings for the raw stimuli and the ADPCN causes.  The projections of the ADPCN causes show that they preserve perceptual differences well.  For each row, the plot on the left-hand side shows the embedding for the visual stimuli.  The embedding tends to emphasize grouping stimuli of simliar brightness together.  This often does not correspond to grouping according to object classes, except in very limited circumstances.  The stimuli this need to be transformed in a meaningful way.  The plot on the right-hand side shows the embedding of the causes from multiple ADPCN stages.  The causes often organize the stimuli in a way that groups related sets of visual content.  This better aligns with the object classes than the raw pixel values.  We recommend zooming in to see the full image details.\vspace{-0.1cm}}
   \label{fig:tsne-embedding-1}
\end{figure*}

\addtocounter{figure}{-1}

\begin{figure*}[h!]\vspace{-0.3cm}
   \hspace{-0.3cm}\begin{tabular}{c}
      \includegraphics[width=6.65in]{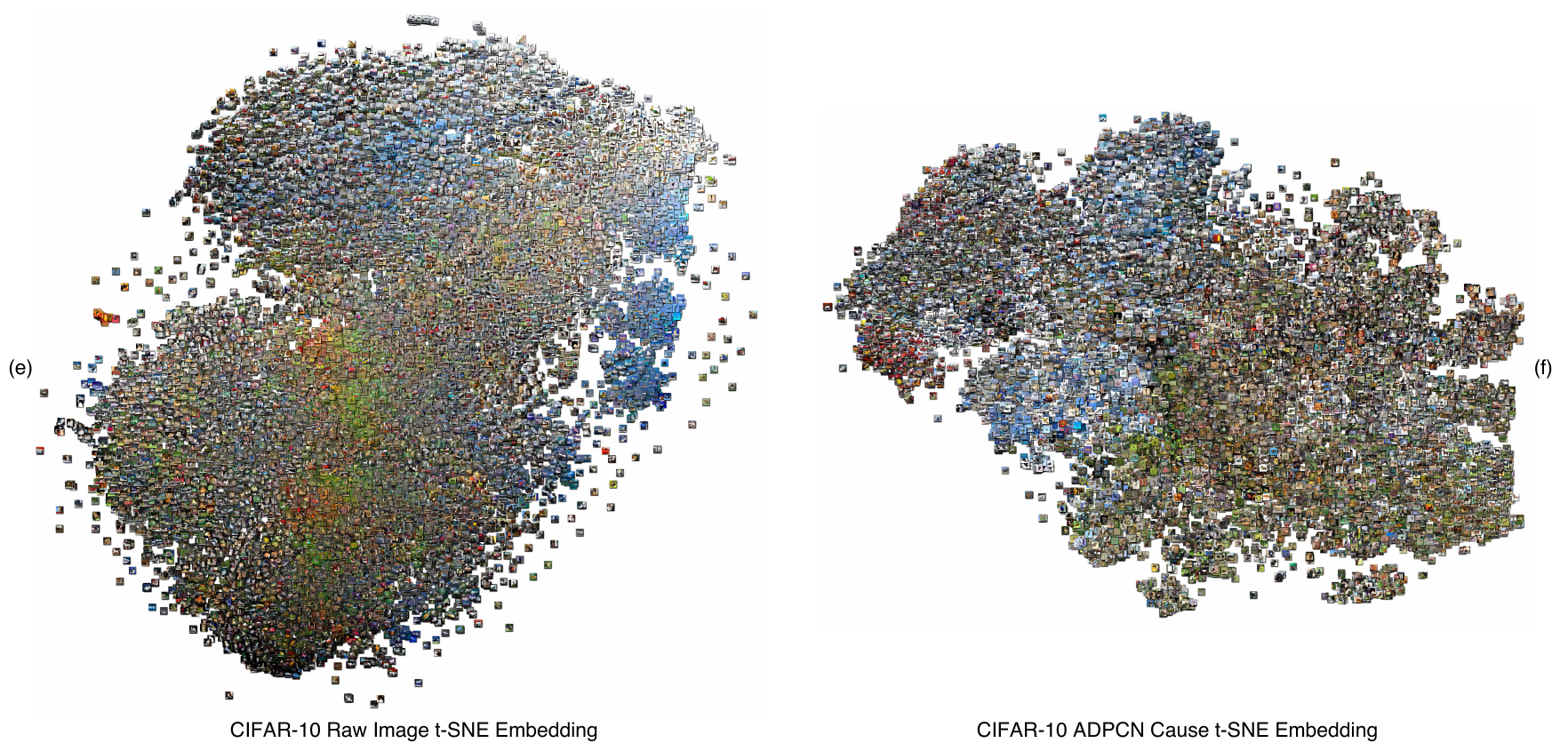}\vspace{-0.1cm}\\
      \includegraphics[width=6.65in]{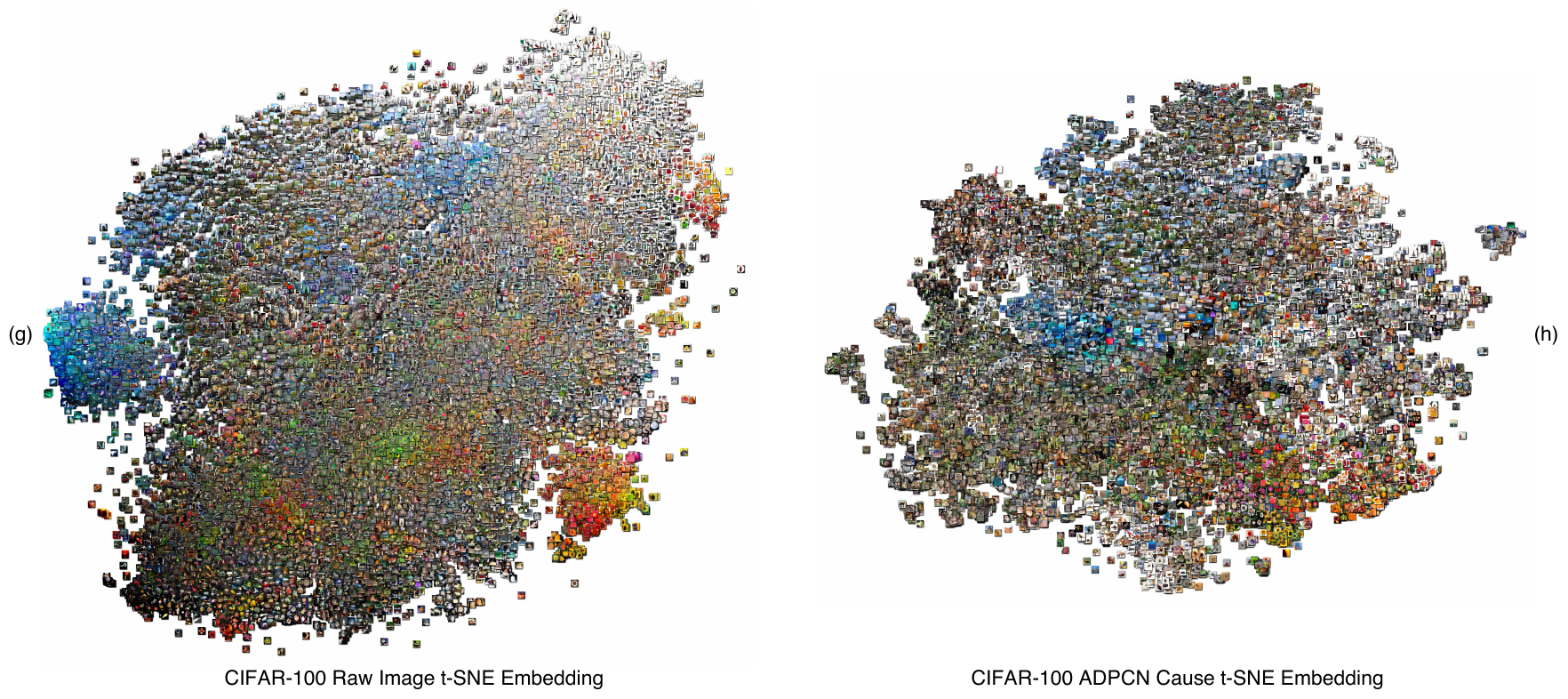}\vspace{-0.15cm}\\
      \includegraphics[width=6.65in]{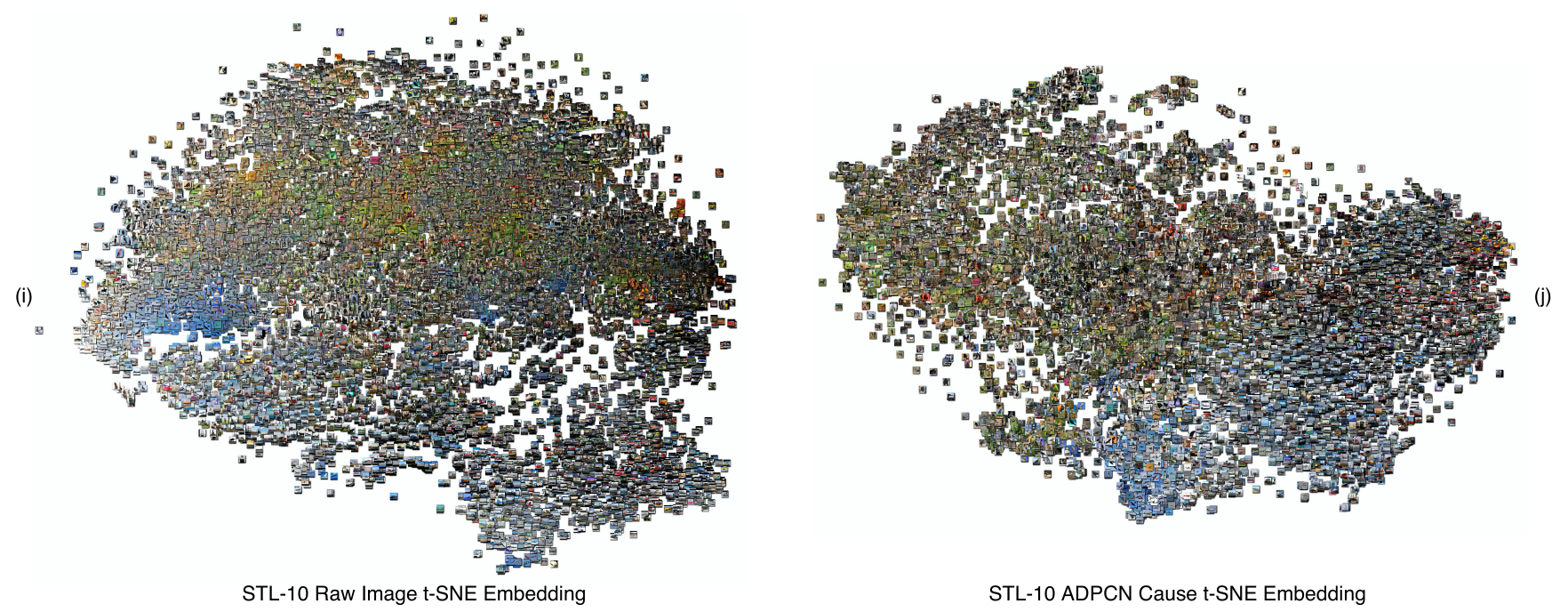}\vspace{-0.1cm}
   \end{tabular}
   \caption[]{\fontdimen2\font=1.55pt\selectfont (Continued from the previous page)\vspace{-0.4cm}}
   \label{fig:tsne-embedding-2}
\end{figure*}

In \cref{fig:tsne-embedding-1}, we provide plots of the t-SNE embeddings of the raw visual stimuli and the ADPCN features extracted from them.  There are multiple interesting findings that we observe, all of which indicate that the ADPCNs preserve perceptual differences due to being sensitive to visual stimuli appearance:

\begin{itemize}
\item[]\-\hspace{0.5cm}{\small{\sf{\textbf{MNIST}}}}: \Cref{fig:tsne-embedding-1}(a) indicates that MNIST is largely a separable dataset without any processing.  Most of the digit classes group into distinct distributions.  There is, however, some inherent ambiguity when distinguishing between certain classes.  For instance, digits eight and three may resemble digit five, depending on the writing style, when solely considering the raw pixels as features.  Similar issues are encountered for digits four and nine.  However, the ADPCNs are adept at separating the digits based on their local and global appearance.  This resolves much of the inherent ambiguity and leads to better separated distributions.  Interestingly, the ADPCNs often additionally cluster the samples based on the writing style.  As shown in \cref{fig:tsne-embedding-1}(b), it separates well instances of digit seven which do and do not have horizontal strokes.  The former are mostly located in a smaller, isolated distribution.  For digit one, the distribution smoothly varies from slanted strokes to mostly horizontal strokes.  For digit two, there is a progression from strokes which have loops to those that are flat.  This indicates that the ADPCN is additionally differentiating between handwriting styles.

\item[]\-\hspace{0.5cm}{\small{\sf{\textbf{FMNIST}}}}: Unlike MNIST, FMNIST is not trivially separable when considering raw pixel values.  \Cref{fig:tsne-embedding-1}(c) illustrates that sandals, sneakers, and ankle boots are located in a single, giant distribution.  Pullovers, coats, shirts, dresses, and t-shirts form another giant distribution.  There is some progression from one type of clothing to another, but, often, the classes are interspersed.  The remaining classes, trousers and bags, are often well separated.  \Cref{fig:tsne-embedding-1}(d) highlights that the ADPCNs better segregate the visual stimuli in an appearance-based manner.  The ADPCN features for sandals are rather distinct from those of other footwear.  The network thus organizes them in a separate distribution.  Features for ankle boots and tennis shoes form a heavily bi-modal distribution.  There is a clear dividing line between modes, though, which illustrates that the ADPCNs are sensitive to the visual appearance of the stimuli.  Likewise, the features for bags form a bi-modal distribution.  Those with straps are predominantly located in one mode and those without straps in another mode.  This, again, demonstrates that the networks are cognizant of stimuli appearance.  It also suggests that the ADPCNs understand that these stimuli belong to the same class, despite not providing any class labels to the network.  Dresses, shirts, coats, and t-shirts are all grouped in a quad-modal distribution, which we believes occurs due to their shared visual characteristics.  Instances of these classes all have, more or less, a similar global shape.  There are conspicuous local shape differences, though.  The ADPCNs thus could distinguish that dresses are distinct from t-shirts, coats, and long-sleeved shirts.  There is hence a rather clear division between them.  There is also an abrupt transition from t-shirts to shirts.  Coats and shirts are more difficult to separate.  Long-sleeved shirts are typically allocated to non-contiguous, bi-modal distributions.  Coats tend to be offshoots of these distributions.

\item[]\-\hspace{0.5cm}{\small{\sf{\textbf{CIFAR-10/100}}}}: Adding, essentially, random backgrounds to the stimuli preempt any easy separability and hence discrimination.  Including multiple object perspectives, appearances, and so forth compounds this issue.  The t-SNE embedding for CIFAR-10, given in \cref{fig:tsne-embedding-1}(e), highlight this.  That for CIFAR-100, in \cref{fig:tsne-embedding-1}(g), further corroborate it.  For both datasets, the stimuli are locally grouped according to global image hue, saturation, and lightness.  All of the classes are highly intermixed.  The ADPCNs re-organize the stimuli in a mostly class-sensitive and hue-sensitive way.  However, owing to the visual difficulty of the stimuli, the network features do not yield completely separable distributions.  Instead, the stimuli belong to a single, multi-modal distribution, with each mode usually corresponding to either a different class or some subset of a class.  For example, in the case of CIFAR-10, images for the automobile class are located on the left-hand side of the plot in \cref{fig:tsne-embedding-1}(f).  Red, orange, and yellow cars form a sub-distribution.  Large trucks, interestingly, form another sub-distribution.  The automobile distribution mode gives way to modes for airplanes, boats, and birds toward the middle-left of the embedding plot.  Birds are, visually, more similar to planes and boats, so stimuli from the former two categories are proximally located.  For the remaining classes, there is heavy inter-mixing of stimuli.  Deer and horses resemble each other greatly.  There are scant distinguishing characteristics, so the ADPCNs often group them together.  Such a finding, once again, shows that the ADPCNs are sensitive to visual content.  This property does not, however, always imply that the networks will be sensitive to classes.  We encounter similar visual-grouping behaviors for stimuli from CIFAR-100, as shown in \cref{fig:tsne-embedding-1}(h).  Many of the classes form sub-modes.  These sub-modes approximately correspond to around thirty coarsely-grained superclasses, not the hundred finely-grained classes.  For instance, there are sub-modes for flowers, fruits, insects, plates, cups, bottles, cellphones, keyboards, lawn mowers, automobiles and trucks, trees, whales and sharks, people, and various types of mammals. 

\item[]\-\hspace{0.5cm}{\small{\sf{\textbf{STL-10}}}}: STL-10 has the same difficulties as CIFAR-10/100, so it is not surprising that the raw pixel values do not divide the stimuli in a class-based manner.  As with \cref{fig:tsne-embedding-1}(e) and \cref{fig:tsne-embedding-1}(g), \cref{fig:tsne-embedding-1}(i) shows that the stimuli are mainly arranged by global hue, saturation, and lightness.  When transformed by the ADPCNs, the stimuli are amassed in a distributional manner with several modes.  In \cref{fig:tsne-embedding-1}(j), it can be seen that there are contiguous modes that individually correspond to planes, boats, cars, trucks, and birds.  There are multiple, non-contiguous modes for dogs, monkeys, and cats.  Some internal structure is observed within each mode, such as a grouping according to hue.  While not easily separable, there are non-linear transition boundaries between sets of modes.  Deer and horses share a similar distribution mode, just as in CIFAR-10/100.  We also encounter other trends in the organization of the ADPCN features for STL-10 that resemble those from CIFAR-10.
\end{itemize}

From various ablation studies that we performed, this appearance-based sensitivity appears to predominantly emerge from the interaction of the various convolutional network stages.  Each ADCPN stage learns features at a different spatial scale.  Intra-layer feed-back provided by the top-down connections helps to propagate contextual details between the layers.  This results in lower-stage features that are often more informative.  Such features permit uncovering higher-stage attributes that are more sensitive to object appearance.  In fact, as we show, whole-object sensitivity is often possible.  It is likely that including forward skip connections would further improve object sensitivity.

Visual sensitivity can occur without top-down feed-back.  However, it requires many times more epochs.  Often, the cause embeddings are not as good as those shown above.  The classes heavily overlap with no clear transition boundaries.  There are often multiple, non-contiguous modes corresponding to each class.


\end{document}